\documentclass{article}
\usepackage{microtype}
\usepackage{graphicx}
\usepackage{subcaption}
\usepackage{booktabs} 

\usepackage{hyperref}
\usepackage{algpseudocode}
\usepackage{amsmath,amssymb}
\PassOptionsToPackage{numbers, compress}{natbib}
\usepackage[main,final]{neurips_2026}
\usepackage{tabularx}
\usepackage{array}
\usepackage{booktabs}
\usepackage{booktabs}
\usepackage{placeins}
\usepackage{caption}

\usepackage[utf8]{inputenc} 
\usepackage[T1]{fontenc}    
\usepackage{hyperref}       
\usepackage{url}            
\usepackage{booktabs}       
\usepackage{amsfonts}       
\usepackage{nicefrac}       
\usepackage{microtype}      
\usepackage{xcolor}         
\usepackage{hyperref}
\usepackage{amsmath}
\usepackage{amssymb}
\usepackage{mathtools}
\usepackage{amsthm}
\usepackage{multirow}

\usepackage[capitalize,noabbrev]{cleveref}
\usepackage{amsthm}
\usepackage{enumitem}
\usepackage{algorithm}
\usepackage{algpseudocode}

\theoremstyle{plain}
\newtheorem{theorem}{Theorem}[section]

\newtheorem{lemma}[theorem]{Lemma}

\theoremstyle{definition}

\theoremstyle{remark}

\newcommand{\x}{\boldsymbol{x}}
\newcommand{\y}{\boldsymbol{y}}
\newcommand{\z}{\boldsymbol{z}}
\newcommand{\bu}{\boldsymbol{u}}
\newcommand{\boldv}{\boldsymbol{v}}

\newcommand{\w}{\boldsymbol{w}}

\newcommand{\R}{\mathbb{R}}
\newcommand{\E}{\mathbb{E}}

\usepackage[textsize=tiny]{todonotes}
\title{CAB: Accelerating Flow and Diffusion Sampling via Rectification and Corrected Adams-Bashforth}

\author{
  Anuska Roy \quad Pravin Nair \\
  Department of Electrical Engineering \\
  Indian Institute of Technology Madras, Chennai, India \\
  \texttt{ee25s070@smail.iitm.ac.in, pravinnair@ee.iitm.ac.in}
}

\begin{document}

\maketitle

\begin{abstract}
Flow and diffusion models achieve high-fidelity, high-resolution image synthesis, but often require many function evaluations (NFEs) at sampling time. Existing acceleration methods either require additional training through distillation or rely on training-free high-order solvers, and both can degrade sample quality at low NFE budgets. We propose CAB (Corrected Adams-Bashforth), a training-free sampler that accelerates both flow and diffusion models. CAB first transforms the sampling dynamics to a common rectified coordinate system, and then applies a multistep Adams--Bashforth predictor augmented with a simple correction term based on past velocity evaluations and therefore incurs no additional NFEs. The resulting method is simple, has the same algorithmic form across model classes, and has at least third-order local truncation error and second-order global error. 
Experiments on pretrained flow and diffusion models, including class-conditional and large-scale text-to-image benchmarks, show that CAB improves quality--NFE trade-offs in the low-step regime of \(6\)--\(20\) NFEs. It also remains competitive with strong training-free samplers at higher step counts across most tested models. The official implementation is available at \url{https://github.com/Anuska-Roy/CAB}.
\end{abstract}

\section{Introduction}
\label{sec:Intro}

Flow and diffusion models have become the dominant choice for high-fidelity image and video generation~\citep{ho2020ddpm,lipman2023flowmatching,
ho2022videodiffusion}, and they increasingly serve as backbones for editing
and restoration pipelines ~\cite{huang2025diffusioneditingsurvey,saharia2021palette,
lugmayr2022repaint,li2023restorationenhancement,
wang2025videodiffusionsurvey}.
Despite their impressive quality, sampling from these models remains computationally expensive because generation often involves numerically integrating a learned reverse-time ordinary differential equation over many neural network evaluations~\cite{lipman2023flowmatching,song2021score_sde,lu2022dpmsolver}. This cost is particularly limiting in practical deployment, where latency and throughput matter, motivating continued interest in training-free few-step samplers that improve the quality-compute trade-off without any additional training~\cite{lu2022dpmsolver,song2021ddim,zhao2023unipc}. The most important regime is often the low-NFE setting, where only $6$ to $20$ network evaluations are affordable, but it is also the most challenging, since discretization errors accumulate rapidly and visibly degrade sample fidelity~\cite{lu2022dpmsolver,zhao2023unipc}.

\begin{figure*}[t]
\vspace{-0.8em}
\centering
\setlength{\abovecaptionskip}{3pt}
\setlength{\belowcaptionskip}{0pt}

\begin{minipage}[b]{0.24\textwidth}
    \centering \small \textbf{Euler / DDIM}
\end{minipage}
\hfill
\begin{minipage}[b]{0.24\textwidth}
    \centering \small \textbf{DPM-Solver++}
\end{minipage}
\hfill
\begin{minipage}[b]{0.24\textwidth}
    \centering \small \textbf{STORK}
\end{minipage}
\hfill
\begin{minipage}[b]{0.24\textwidth}
    \centering \small \textbf{CAB-2}
\end{minipage}

\vspace{0.2em}

\includegraphics[width=0.24\textwidth]{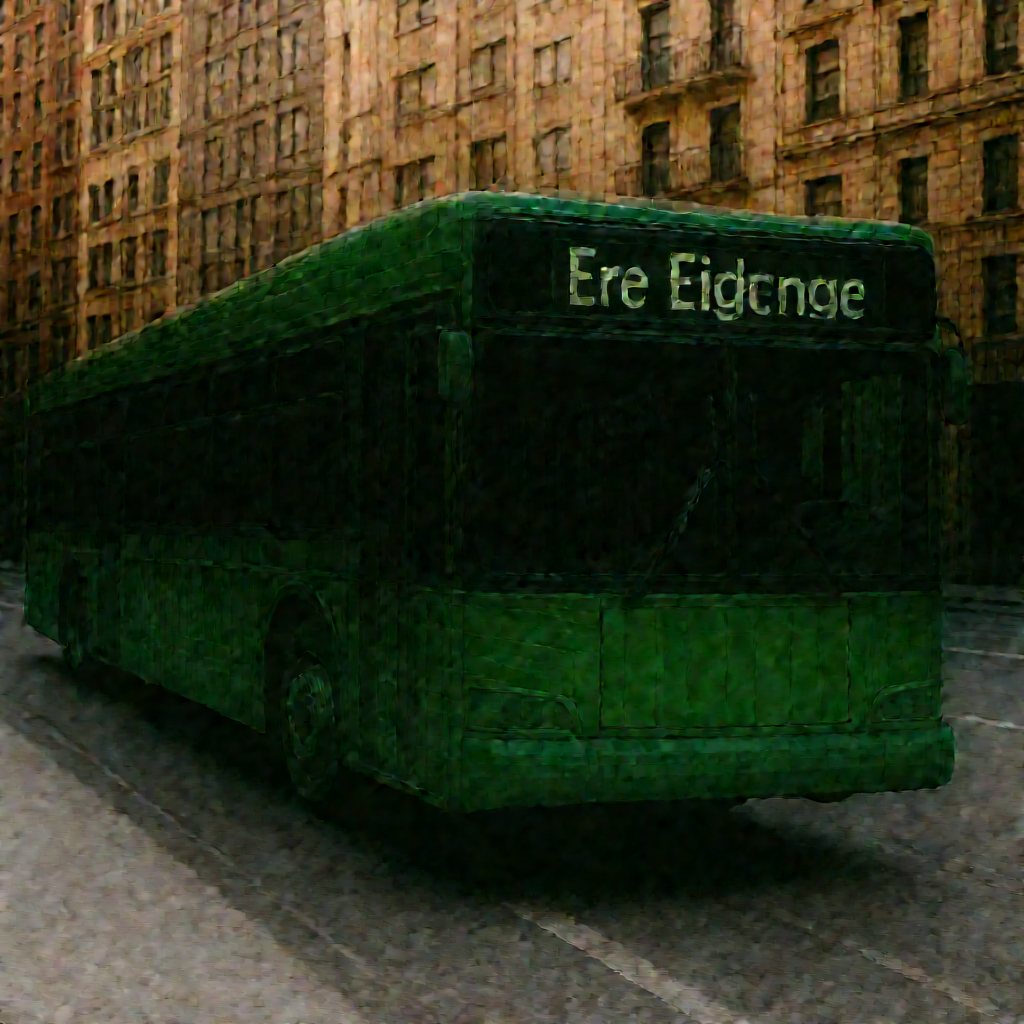}\hfill
\includegraphics[width=0.24\textwidth]{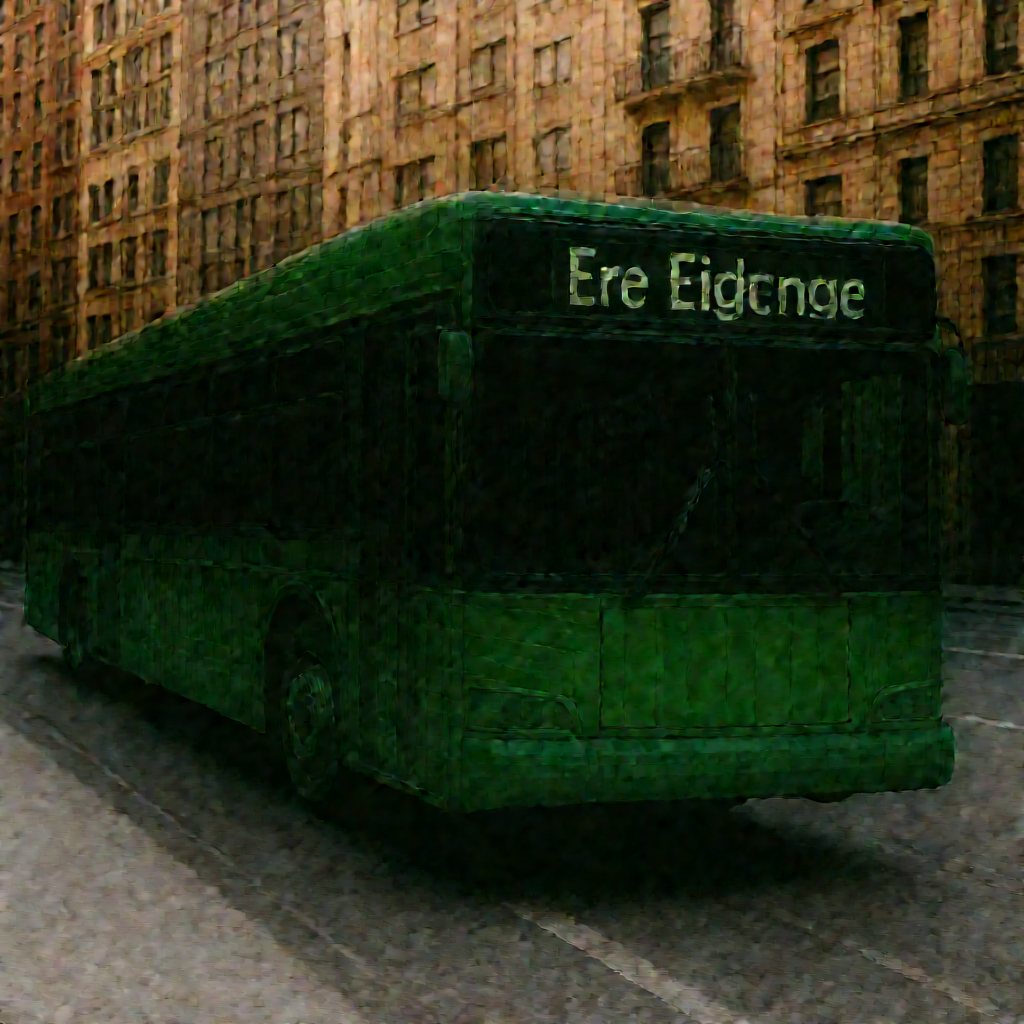}\hfill
\includegraphics[width=0.24\textwidth]{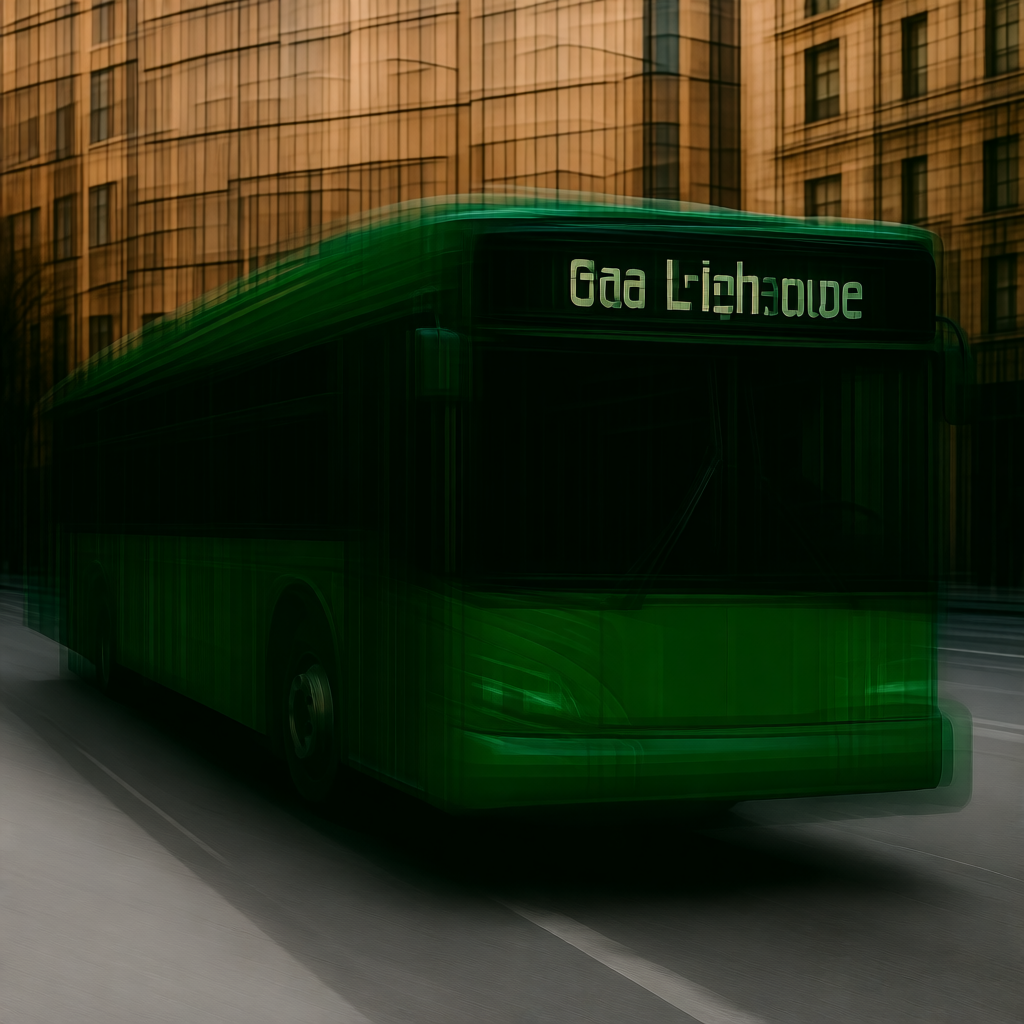}\hfill
\includegraphics[width=0.24\textwidth]{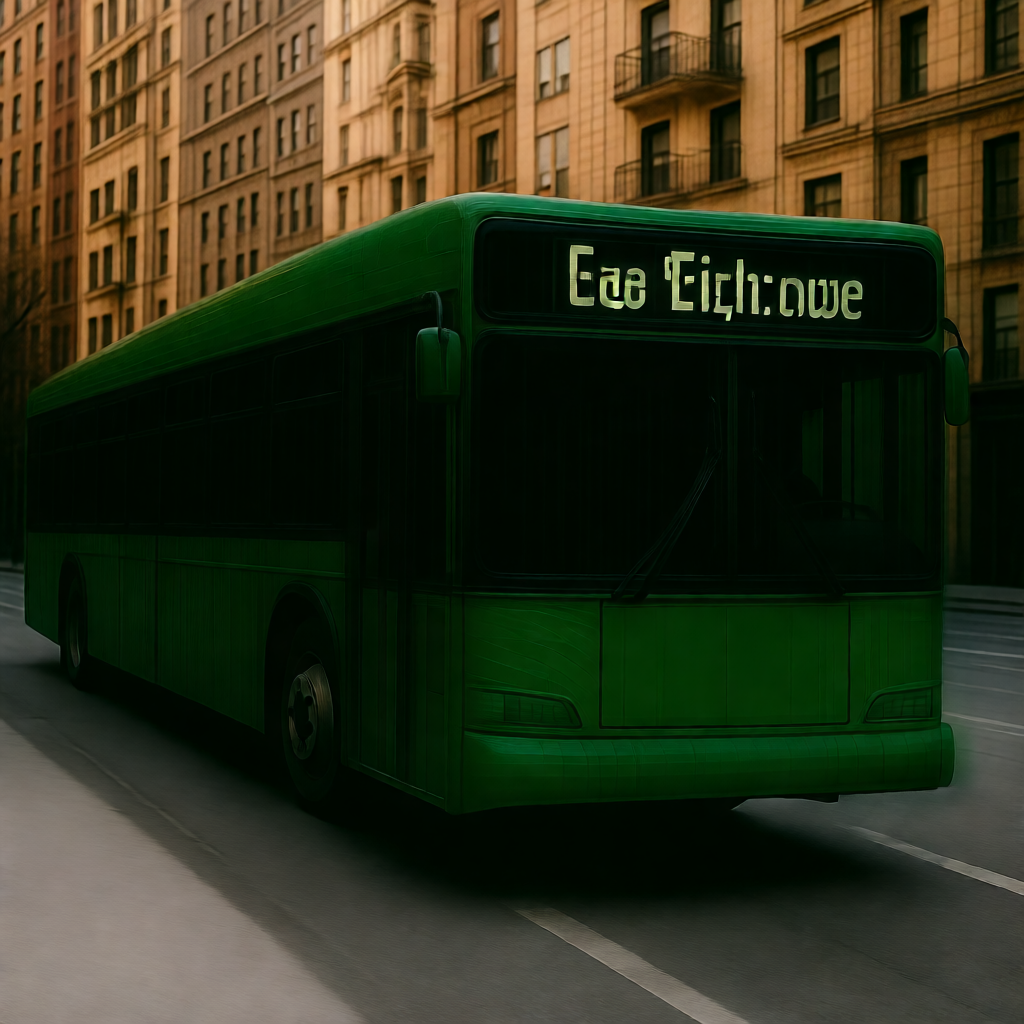}

\vspace{-0.2em}
\caption*{\footnotesize
\textbf{Prompt 1:} \textit{``A green bus with signage on a city street.''}
CAB-2 produces a sharper and detail-enhanced sample at 6 NFEs, while existing samplers either retain strong noise artifacts or oversmooth the sample.
}

\vspace{0.35em}

\includegraphics[width=0.24\textwidth]{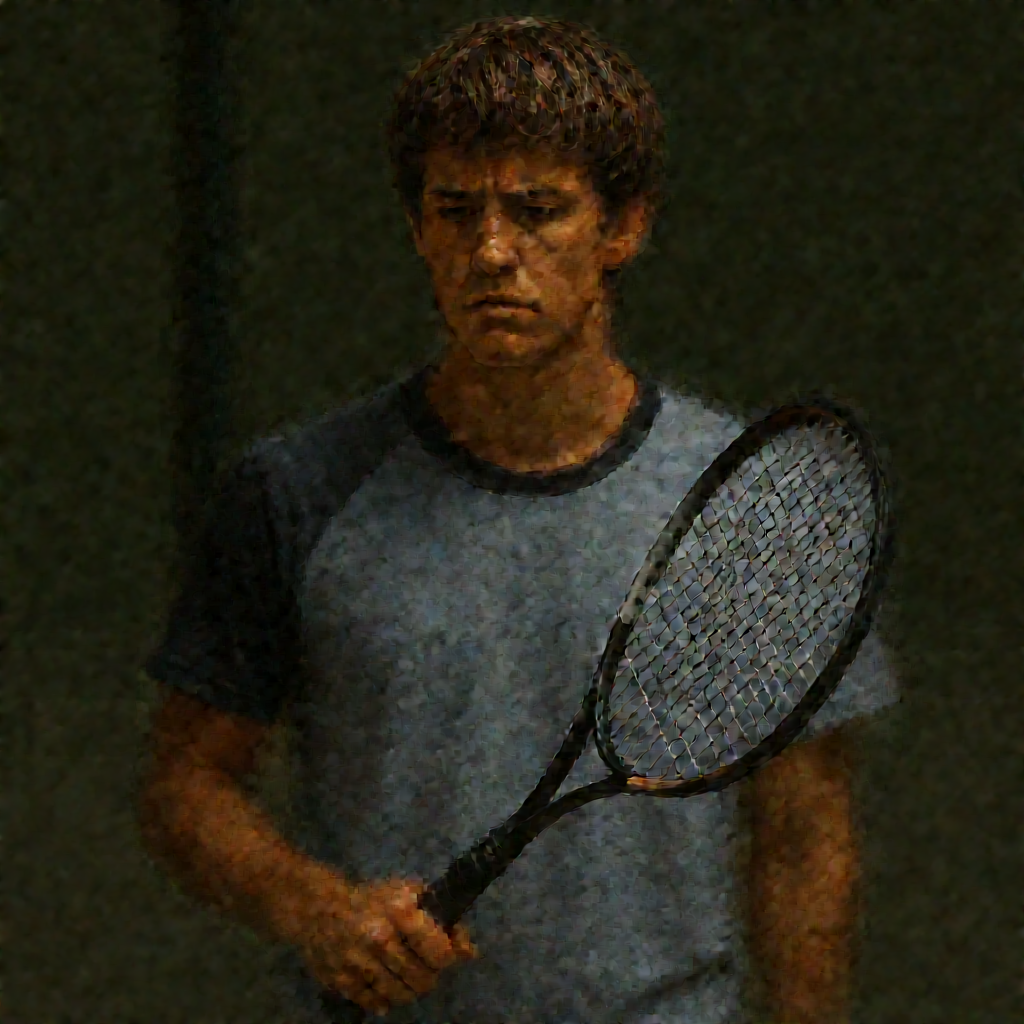}\hfill
\includegraphics[width=0.24\textwidth]{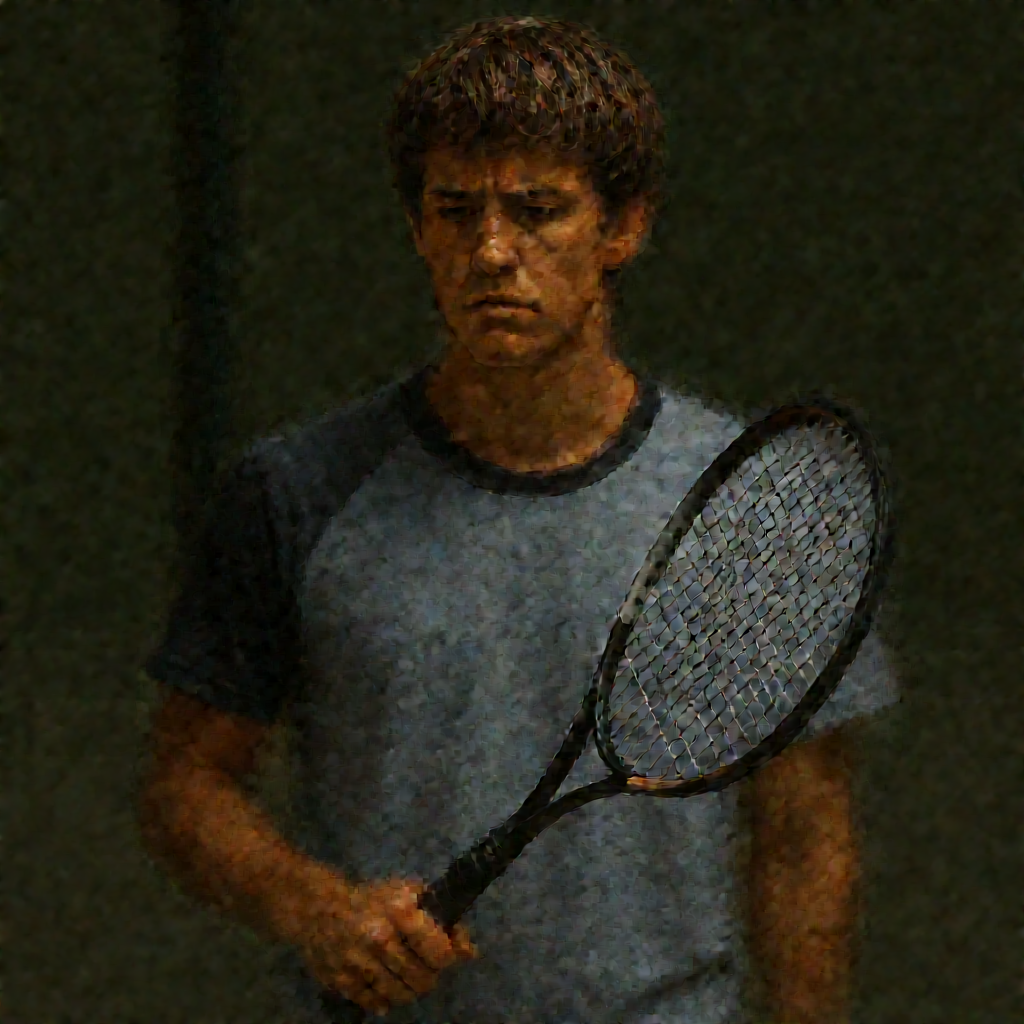}\hfill
\includegraphics[width=0.24\textwidth]{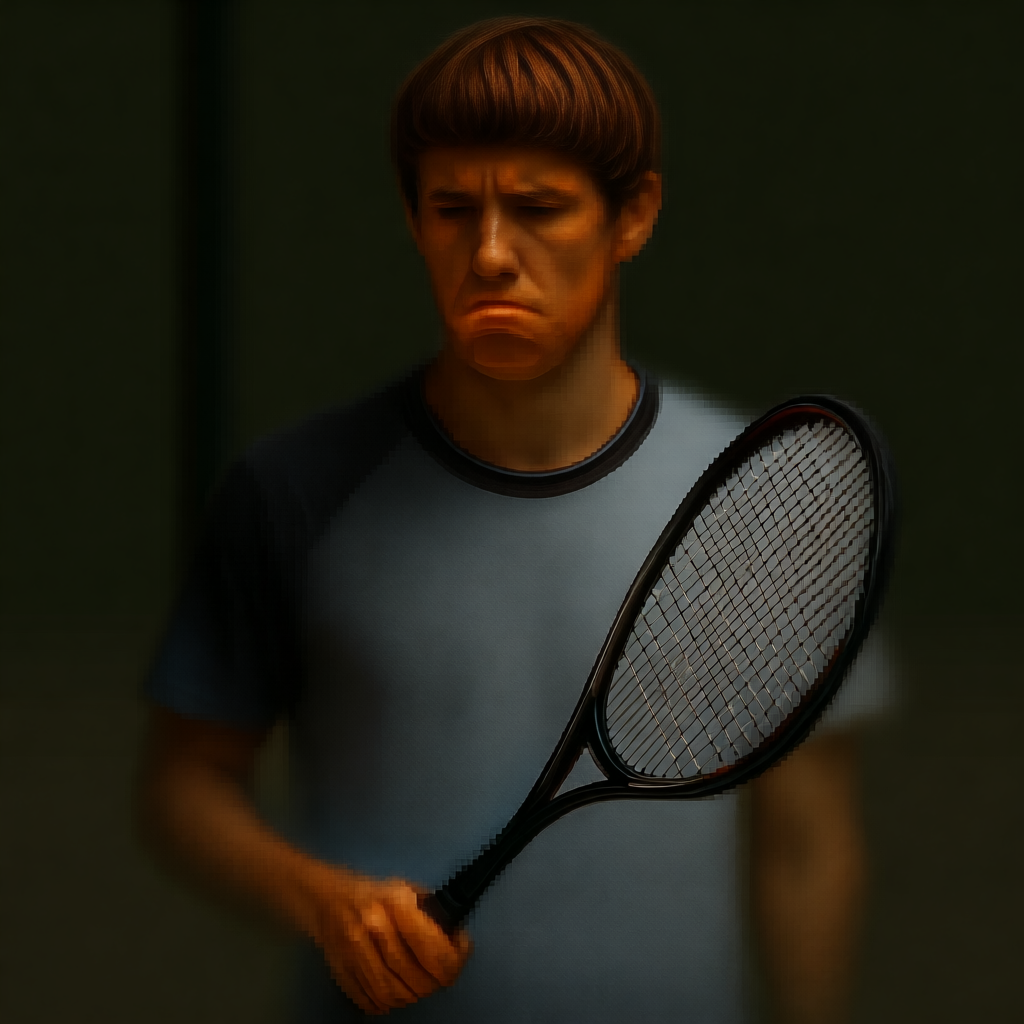}\hfill
\includegraphics[width=0.24\textwidth]{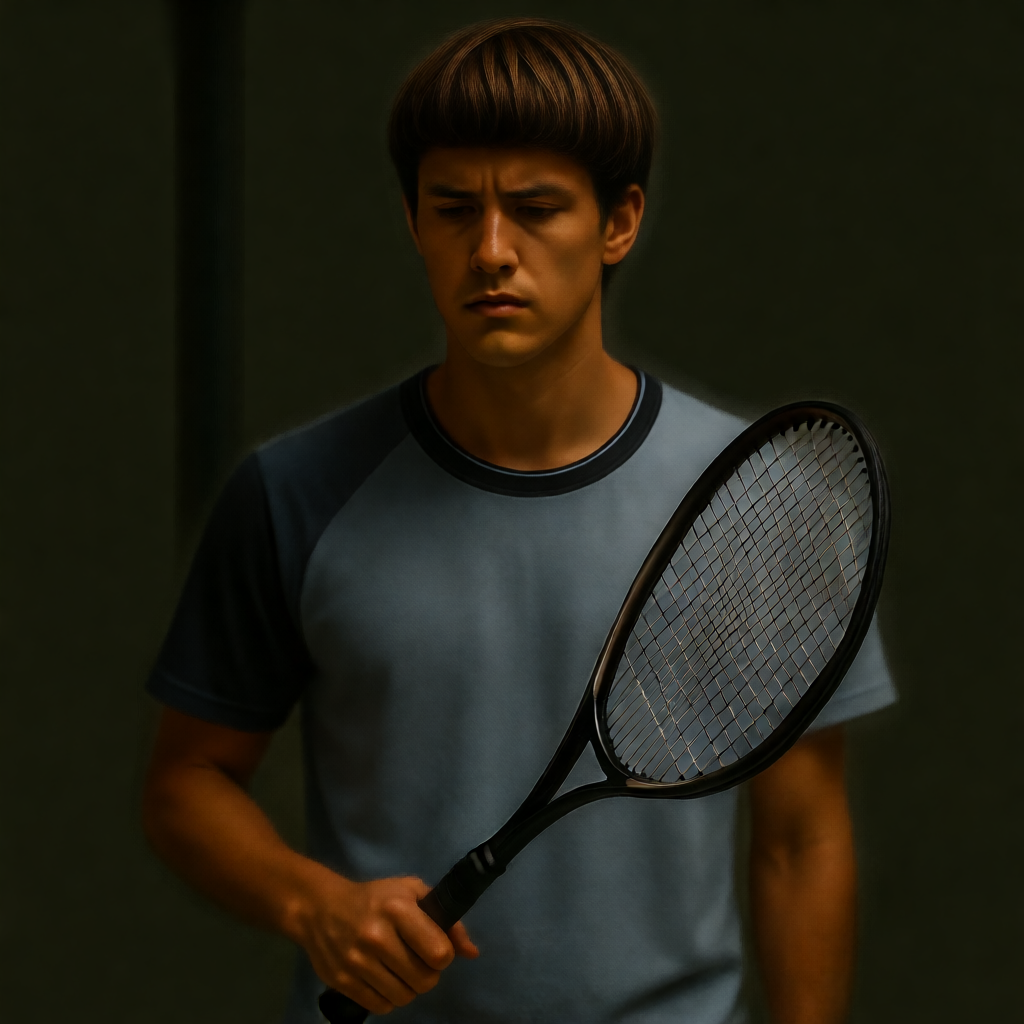}

\vspace{-0.2em}
\caption*{\footnotesize
\textbf{Prompt 2:} \textit{``The tennis player looks solemn holding his racket.''}
CAB-2 preserves facial structure, pose, and racket details better at 6 NFEs, while competing solvers show grain and distortion.
}

\vspace{0.35em}

\includegraphics[width=0.24\textwidth]{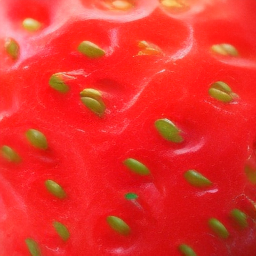}\hfill
\includegraphics[width=0.24\textwidth]{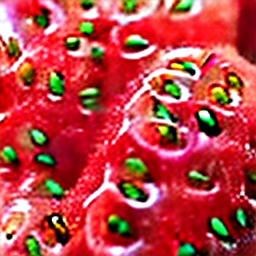}\hfill
\includegraphics[width=0.24\textwidth]{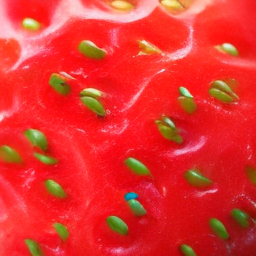}\hfill
\includegraphics[width=0.24\textwidth]{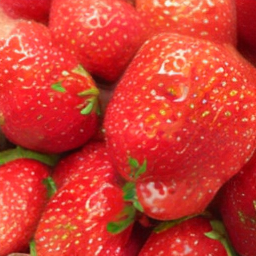}

\vspace{-0.2em}
\caption*{\footnotesize
\textbf{Class-conditional generation:}
CAB-2 produces a more realistic sample with clearer global structure, while competing samplers show oversmoothing, artifacts, or distorted texture at 8 NFEs.
}

\vspace{-0.2em}
\caption{Qualitative comparison of training-free samplers on QWEN-Image \(1024\times1024\) flow generation~\cite{wu2025qwen} and DiT \(256\times256\) VP diffusion generation~\cite{peebles2023scalable}.}
\label{fig:motivation}
\vspace{-1.2em}
\end{figure*}

Training-free ODE solvers have greatly accelerated flow and diffusion sampling, but each family comes with its own limitations. First-order methods such as DDIM are simple and robust, yet typically require many function evaluations for high-quality generation~\cite{lu2023dpmsolverpp}. Multistep methods such as PNDM improve over first-order baselines, but the low-NFE regime remains challenging ~\cite{zhao2023unipc,liu2022pndm}. Dedicated high-order diffusion solvers such as DEIS and DPM-Solver exploit the semi-linear structure of diffusion ODEs, though prior fast high-order samplers can become unstable under large guidance scales, motivating stabilized variants such as DPM-Solver++~\cite{zhang2023fast,lu2022dpmsolver,lu2023dpmsolverpp}. UniPC improves few-step accuracy over DPM-Solver++, but satisfactory sampling below 10 steps remains difficult in general~\cite{zhao2023unipc}. DPM-Solver-v3 improves low-NFE quality by introducing empirical model statistics computed on the pretrained model, which yields a more model-adaptive solver requiring additional model-dependent estimation \cite{zheng2023dpmsolverv3}. STORK reduces the cost of stabilized Runge--Kutta sampling through virtual NFEs based on Taylor approximation, improving efficiency at the expense of added approximation and solver simplicity \cite{tan2026stork}.  Consequently, existing samplers still face a practical trade-off among accuracy, stability, generality, and simplicity in the low-NFE regime~\cite{zhao2023unipc,lu2023dpmsolverpp,tan2026stork}.  Figure~\ref{fig:motivation} illustrates this low-NFE trade-off, where existing training-free samplers can exhibit noise, oversmoothing, or structural distortions at low NFE budget.
\noindent \textbf{Contribution:} In this work, we propose \textbf{Corrected Adams--Bashforth (CAB)}, a simple framework for constructing explicit multistep samplers for flow and diffusion models via an appropriate rectification of the reverse-time ODE. CAB augments Adams--Bashforth (AB) updates with a lightweight correction term that improves low-NFE stability while retaining the key efficiency advantage of a single network evaluation per iteration. Unlike UniPC, which is formulated in log-SNR diffusion coordinates, CAB applies a noise-to-signal rectification to obtain a common explicit learned-field ODE for both diffusion and flow models.   We also prove that the corrected schemes preserve the convergence order of the underlying AB solvers. Empirically, CAB improves quality--NFE trade-offs across academic benchmarks and large-scale text-to-image/video models. For example, CAB produces sharper and more coherent samples in the low-NFE settings illustrated in Figure~\ref{fig:motivation}. 

\section{Background and Related Work}
\label{sec:prelim}

Let $\x_0\in\R^d$ denote a data sample drawn from an unknown distribution $p_0$. Flow and diffusion models define a family of intermediate marginals $\{p_t\}_{t\in[0,T]}$ that transport $p_0$ to a tractable reference distribution $p_T$, typically chosen as $\mathcal N(\mathbf 0,\mathbf I)$ \cite{ho2020ddpm,song2021score_sde,chen2018neural_ode,grathwohl2019ffjord}. We use the forward direction $t:0\to T$ to describe the corruption process, while sampling integrates the corresponding reverse-time dynamics from $t=T$ to $t=0$ starting from the simple prior $p_T$. Since $p_0$ is unknown, the intermediate marginals are specified through a forward transition kernel $p_{0t}(\x_t\mid \x_0)$ \cite{ho2020ddpm,lipman2023flowmatching}, which induces
\[
p_t(\x)=\E_{\x_0\sim p_0}\!\left[p_{0t}(\x\mid \x_0)\right].
\]
A useful unifying viewpoint is that many widely used constructions, including VP/DDPM-type diffusion models~\cite{ho2020ddpm,song2021score_sde}, VE/EDM-type parameterizations~\cite{song2021score_sde,karras2022edm}, and rectified-flow  paths~\cite{liu2023rectified_flow}, admit the affine-Gaussian forward form
\begin{equation}
\label{eq:affine_marginal}
\x_t = s_t \x_0 + \sigma_t \epsilon,
\qquad
\epsilon\sim\mathcal N(\mathbf 0,\mathbf I),
\end{equation}
or equivalently, 
\[
p_{0t}(\x_t\mid \x_0)=\mathcal N(s_t\x_0,\sigma_t^2\mathbf I).
\]
Different model families correspond to different choices of nonnegative scalar schedules \(s_t,\sigma_t\ge 0\)  \cite{ho2020ddpm,song2021score_sde,karras2022edm,liu2023rectified_flow}. On intervals where \(s_t>0\), we define the noise-to-signal ratio
$\rho_t := {\sigma_t}/{s_t}$. 
For all the  commonly used schedules in \cite{ho2020ddpm,song2021score_sde,karras2022edm,liu2023rectified_flow,lipman2023flowmatching}, \(\rho_t\) is differentiable and strictly increasing. Under this regularity condition, the affine path \eqref{eq:affine_marginal} can be  equivalently realized by a linear forward stochastic differential equation (SDE) initialized at \(\x_0\sim p_0\),  \cite{kingma2021variational,lu2022dpmsolver}, 
\begin{equation}
\label{eq:forward_sde_general}
d\x_t = f(t)\x_t\,dt + g(t)\,d\w_t, 
\qquad
f(t)=\frac{\dot s_t}{s_t},
\qquad
g(t)^2 = 2\,\sigma_t\!\left(\dot{\sigma}_t-\frac{\dot s_t}{s_t}\sigma_t\right),
\end{equation}
where $\w_t$ denotes a standard Wiener process. In particular, the monotonicity condition $\dot\rho_t\ge 0$ ensures that the diffusion coefficient $g(t)$ is real-valued and nonnegative. This monotonicity condition is also required for the noise-to-signal rectification used to derive the proposed solver in Section~\ref{sec:method}.

Given the forward SDE in \eqref{eq:forward_sde_general}, there exists a corresponding reverse-time generative process \cite{song2021score_sde}. Starting from the prior \(\x_T\sim\mathcal N(\mathbf 0,\mathbf I)\), samples from the data distribution can be generated by integrating the associated reverse-time ODE from \(t=T\) to \(t=0\). In the noise-prediction parameterization commonly used in diffusion models, the reverse-time ODE is, 
\begin{equation}
\label{eq:reverse_ode_eps}
\frac{d\x_t}{dt}
=
f(t)\x_t+\frac{g(t)^2}{2\sigma_t}\,\epsilon_\theta(\x_t,t),
\qquad t\in[0,T],
\end{equation}
with initialization $\x_T\sim p_T$  \cite{song2021score_sde,lu2023dpmsolverpp,zhao2023unipc,tan2026stork}. Here, $\epsilon_\theta(\x_t,t)$ denotes a neural network trained to predict the noise component $\epsilon$ from $\x_t$ in the forward path \eqref{eq:affine_marginal}. Equivalent parameterizations are also common. In particular, the data prediction model $\hat{\x}_{\theta}(\x_t,t)$ can directly predict $\x_0$ from $\x_t$, where its relation with $\epsilon_\theta(\x_t,t)$ is given as,
\begin{equation}
\hat{\x}_{\theta}(\x_t,t)
=
\frac{\x_t-\sigma_t\epsilon_\theta(\x_t,t)}{s_t}.
\label{eq:eps_prediction}
\end{equation}
while the reverse-time velocity field in \eqref{eq:reverse_ode_eps} may be learned directly as
\[
\boldv_\theta(\x_t,t)=\dot s_t\,\hat{\x}_{\theta}(\x_t,t)+\dot\sigma_t\,\epsilon_\theta(\x_t,t)
\]
as in rectified flow \cite{lipman2023flowmatching,liu2023rectified_flow}. 
Thus, noise prediction, clean-sample prediction, and direct velocity prediction are equivalent parameterizations of the same reverse-time dynamics.

\noindent \textbf{Fast solvers:} One class of existing fast solvers  
accelerates generation through additional training, including progressive distillation, consistency-style models, and one-step diffusion models \cite{salimans2022progressive,starodubcev2025,
song2023consistency_models,lu2025,liu2024instaflow,yin2024,chen2025onestep}. 
In contrast, training-free samplers \cite{song2021ddim,liu2022pndm,zhang2023fast,lu2022dpmsolver,
lu2023dpmsolverpp,zhao2023unipc,xue2023sa,xu2023restart_sampling,
zheng2023dpmsolverv3,shaul2024bespoke,
frankel2025ss,tan2026stork} retain the pretrained model and accelerate generation through improved inference-time solver design. Among these, the most widely used methods solve the reverse-time ODE in \eqref{eq:reverse_ode_eps}. 
Early training-free samplers such as DDIM discretize \eqref{eq:reverse_ode_eps} using a first-order update, while PNDM combines Runge--Kutta startup with linear multistep integration \cite{song2021ddim,liu2022pndm}. DPM-Solver reparameterizes \eqref{eq:reverse_ode_eps} using the log-SNR variable $\lambda_t=\log\!\frac{s_t}{\sigma_t}$ and, under noise prediction, rewrites the ODE \eqref{eq:reverse_ode_eps} in a semi-linear form that admits high-order exponential-integrator updates \cite{lu2022dpmsolver}. DPM-Solver++~\cite{lu2023dpmsolverpp} retains the same log-SNR reparameterization but replaces noise prediction with data-prediction, yielding improved guided sampling and stabilized multistep variants. UniPC develops a unified predictor--corrector framework for the same reverse dynamics, yielding arbitrary-order samplers with zero-extra-NFE correction and improved low-NFE performance \cite{zhao2023unipc}. DPM-Solver-v3 further improves low-NFE quality by incorporating empirical model statistics computed on the pretrained model \cite{zheng2023dpmsolverv3}. When the reverse-time velocity field is learned directly in \eqref{eq:reverse_ode_eps}, as in flow matching and rectified flow, generation is performed by standard ODE solvers, with Euler, Heun, and related Runge--Kutta schemes being the most common choices \cite{lipman2023flowmatching,liu2023rectified_flow}. More recently, STORK introduced stabilized Taylor--Runge--Kutta solvers that address both stiffness and dependence on semi-linear diffusion structure, extending naturally beyond classical diffusion ODEs to flow-matching models \cite{tan2026stork}.

\section{Corrected Adams-Bashforth Solver}
\label{sec:method}

In this section, we develop a training-free sampler by reparameterizing the ODE in \eqref{eq:reverse_ode_eps} to noise-signal coordinates, and designing a corrected Adams-Bashforth method for flow and diffusion sampling.

\subsection{Rectification of the reverse-time ODE in noise-to-signal coordinates}

Our key observation is that the reverse-time ODE in \eqref{eq:reverse_ode_eps} admits a particularly simple form after an appropriate change of state and time variables. Specifically, by rescaling the state as $\y_t=\x_t/s_t$ and reparameterizing time using the noise-to-signal ratio $\rho_t=\sigma_t/s_t$, the linear drift induced by the schedules $(s_t,\sigma_t)$ is removed. In the resulting coordinates, the reverse dynamics reduce to a rectified ODE whose right-hand side is given directly by the learned noise field as shown in Lemma \ref{lem:rho_rectified}.

\begin{lemma}[Rectified form in noise-to-signal coordinates]
\label{lem:rho_rectified}
Assume $s:[0,T]\to \R_{>0}$ and $\sigma:[0,T]\to \R_{>0}$ are $C^1$. Define the noise-to-signal ratio $\rho_t := {\sigma_t}/{s_t}$, and assume that $\rho:[0,T]\to\R_{>0}$ is strictly increasing, so that the inverse map $t=t(\rho)$ exists. Consider the reverse-time ODE in \eqref{eq:reverse_ode_eps}. Define the rescaled state and its parametrization in $\rho$ space as
\[
\y_t = \frac{\x_t}{s_t},
\qquad
\hat{\y}_\rho = \y_{t(\rho)}.
\]
Then the $\rho$-reparameterized trajectory satisfies
\begin{equation}
\label{eq:rectified_rho_ode}
\frac{d\hat{\y}_\rho}{d\rho}
=
\epsilon_\theta \bigl(s_{t(\rho)}\,\hat{\y}_\rho,\; t(\rho)\bigr).
\end{equation}
\end{lemma}
The proof is given in Appendix~\ref{app:rectified_lemma}. As in \eqref{eq:reverse_ode_eps}, the noise-prediction model in \eqref{eq:rectified_rho_ode} can be replaced by equivalent neural parameterizations. In particular,
\begin{equation}
\label{eq:rectified_param_relations}
\epsilon_\theta\!\bigl(s_{t(\rho)}\hat{\y}_\rho,t(\rho)\bigr)
=
\frac{\hat{\y}_\rho-\hat{\x}_{\theta}\!\bigl(s_{t(\rho)}\hat{\y}_\rho,t(\rho)\bigr)}{\rho}
=
\frac{\boldv_\theta\!\bigl(s_{t(\rho)}\hat{\y}_\rho,t(\rho)\bigr)-\dot s_{t(\rho)}\,\hat{\y}_\rho}{s_{t(\rho)}\,\dot\rho_{t(\rho)}},
\end{equation}
where recall that $\hat{\x}_{\theta}$ denotes the data-prediction model and $\boldv_\theta$ the reverse-time velocity field.
The assumption required in Lemma~\ref{lem:rho_rectified} that $\rho_t=\sigma_t/s_t$ is strictly increasing, i.e., $\dot{\rho}_t>0$, is precisely the condition under which the affine-Gaussian path \eqref{eq:affine_marginal} is generated by the linear SDE in Section~\ref{sec:prelim}. 

\subsection{Proposed solver}

We develop our sampler by numerically integrating the rectified reverse-time ODE in \eqref{eq:rectified_rho_ode}. Although the underlying time variable $t$ is typically discretized uniformly, the induced noise-to-signal variable $\rho_t=\sigma_t/s_t$ is in general non-uniformly spaced; accordingly, our solver is formulated directly on the $\rho$-grid. Since \eqref{eq:rectified_rho_ode} has the learned nonlinear field as the right-hand side of the ODE, we can directly apply explicit linear multistep predictors on the \(\rho\)-grid~\cite{hairer1993ode1,butcher2000ode20thcentury}. However, linear-multistep methods such as Adams--Bashforth (AB) may become inaccurate at large step sizes required for low-NFE sampling  \cite{hairer1993ode1,butcher2000ode20thcentury}. Classical predictor--corrector and defect-correction ideas address this issue by using an additional correction stage \cite{hairer1993ode1,ascher1998computer,bohm1984defect,ong2020deferred}. Inspired by this perspective, we correct the Adams-Bashforth prediction using an extrapolation defect, the discrepancy between the current learned velocity and its linear Lagrange extrapolation from previous steps.

\begin{figure*}[t]
\centering
\begin{minipage}[t]{0.56\textwidth}
\begin{algorithm}[H]
\caption{CAB-$p$ sampling ($p = 2$ or $3$)}
\label{alg:cabp_main}
\begin{algorithmic}[1]
\Require Prior sample $\x_{t_0}\sim\mathcal N(\mathbf 0,\mathbf I)$ with $t_0=T$; reverse grid $\{t_i\}_{i=0}^{N}$; schedules $s_t,\sigma_t$; corrector weights $\{\gamma_i\}_{i=2}^{N}$; noise prediction model $\epsilon_\theta(\x_t,t)$.
\Ensure Generated sample $\x_{t_N}$.
\State $\epsilon_0 \leftarrow \epsilon_\theta(\x_{t_0},t_0),\quad \y_{t_0}\leftarrow \x_{t_0}/s_{t_0}$
\State $\rho_0 \leftarrow \sigma_{t_0}/s_{t_0},\quad \rho_{1}\leftarrow \sigma_{t_{1}}/s_{t_{1}},\quad h_0 \leftarrow \rho_{1}-\rho_0$
\State $\y_{t_1}\leftarrow \y_{t_0}+h_0\epsilon_0$
\For{$i=1$ to $N-1$}
    \State $\epsilon_i \leftarrow \epsilon_\theta(\x_{t_i},t_i),\quad \y_{t_i}\leftarrow \x_{t_i}/s_{t_i}$
    \State $\rho_i \leftarrow \sigma_{t_i}/s_{t_i},\quad \rho_{i+1}\leftarrow \sigma_{t_{i+1}}/s_{t_{i+1}}$
    \State $h_i \leftarrow \rho_{i+1}-\rho_i, \quad r_i \leftarrow h_i/h_{i-1}$
    \If{$i=1$}
        \State $r_i \leftarrow h_i/h_{i-1}$
        \State $\y_{t_2}\leftarrow \y_{t_1}+h_i\!\left[\left(1+\frac{r_i}{2}\right)\epsilon_i-\frac{r_i}{2}\epsilon_{i-1}\right]$
    \Else
        \State $\y^{\mathrm{pred}}_{t_{i+1}} \leftarrow \mathrm{ABPredictor}(p,i)$
        \State $\epsilon_i^{\mathrm{ext}} \leftarrow (1+ r_{i-1})\epsilon_{i-1}- r_{i-1}\epsilon_{i-2}$
        \State $\y_{t_{i+1}} \leftarrow \y^{\mathrm{pred}}_{t_{i+1}}+\gamma_i h_i(\epsilon_i-\epsilon_i^{\mathrm{ext}})$
    \EndIf
    \State $\x_{t_{i+1}}\leftarrow s_{t_{i+1}}\y_{t_{i+1}}$
\EndFor
\State \Return $\x_{t_N}$
\end{algorithmic}
\end{algorithm}
\end{minipage}
\hspace{0.1cm}
\begin{minipage}[t]{0.38\textwidth}
\small
\vspace{1.4cm}
\fbox{%
\parbox[t]{0.95\linewidth}{%
\textbf{AB predictors used in Algorithm~\ref{alg:cabp_main}.}

\vspace{0.4em}
\textbf{Order \(p=2\):}
\begin{align*}
&\mathrm{ABPredictor}(2,i) \\
&=
\y_{t_i}
+
h_i\!\left[\left(1+\frac{r_i}{2}\right)\epsilon_i-\frac{r_i}{2}\epsilon_{i-1}\right].
\end{align*}

\textbf{Order \(p=3\):}
\begin{align*}
&\mathrm{ABPredictor}(3,i) \\
&=
\y_{t_i}
+
h_i\left(\beta_0\epsilon_i+\beta_1\epsilon_{i-1}+\beta_2\epsilon_{i-2}\right), \\
& \text{where} \\
&\beta_0 =
1+\frac{r_i(2 r_{i-1}+1)}{2( r_{i-1}+1)}
+\frac{ r_{i-1} r_i^2}{3( r_{i-1}+1)},\\[1mm]
&\beta_1 =
-\frac{r_i}{6}\left(2 r_{i-1} r_i+3 r_{i-1}+3\right),\\[1mm]
&\beta_2 =
\frac{r_{i-1}^2 r_i(2r_i+3)}{6( r_{i-1}+1)}.
\end{align*}
}
}
\end{minipage}
\end{figure*}

We now derive the resulting update on the nonuniform \(\rho\)-grid induced by a reverse-time discretization. Specifically, sampling integrates the rectified ODE in \eqref{eq:rectified_rho_ode} from \(t=T\) to \(t=0\) using grid points
\[
T=t_0>t_1>\cdots>t_N=0.
\]
For any $i\ge 2$, we next derive the update from $\y_{t_i}$ to $\y_{t_{i+1}}$. Let
\[
\epsilon_i:=\epsilon_\theta(s_{t_i}\y_{t_i},t_i),
\qquad
\epsilon_{i-1}:=\epsilon_\theta(s_{t_{i-1}}\y_{t_{i-1}},t_{i-1}),
\qquad
\epsilon_{i-2}:=\epsilon_\theta(s_{t_{i-2}}\y_{t_{i-2}},t_{i-2})
\]
denote the learned noise predictions at timesteps $t_i$, $t_{i-1}$ and $t_{i-2}$. Further, let step sizes be
\[
h_i:=\rho_{t_{i+1}}-\rho_{t_i},
\qquad
h_{i-1}:=\rho_{t_i}-\rho_{t_{i-1}},
\qquad
h_{i-2}:=\rho_{t_{i-1}}-\rho_{t_{i-2}},
\]
and the step ratios be defined as
\[
r_i:=\frac{h_i}{h_{i-1}},
\qquad
r_{i-1}:=\frac{h_{i-1}}{h_{i-2}}.
\]
Then the CAB-2 update from $\y_{t_i}$ to $\y_{t_{i+1}}$ is given by the AB2 predictor
\begin{equation}
\label{eq:cab2_update_main}
\y_{t_{i+1}}^{\,\mathrm{pred}}
=
\y_{t_i}
+
h_i\!\left[\left(1+\frac{r_i}{2}\right)\epsilon_i-\frac{r_i}{2}\epsilon_{i-1}\right],
\end{equation}
followed by the correction
\begin{equation}
\label{eq:cab2_correction_main}
\y_{t_{i+1}}
=
\y_{t_{i+1}}^{\,\mathrm{pred}}
+
\gamma_i h_i\!\left(\epsilon_i-\epsilon_i^{\,\mathrm{ext}}\right),
\end{equation}
where
\begin{equation}
\label{eq:cab2_extrap}
\epsilon_i^{\,\mathrm{ext}}
=
(1+r_{i-1})\epsilon_{i-1}
-
r_{i-1}\epsilon_{i-2}.
\end{equation}
is the linear Lagrange extrapolation of the velocity at $\rho_{t_i}$ from the two previous steps $\rho_{t_{i-1}}$ and $\rho_{t_{i-2}}$, and $\gamma_i\ge 0$ is the corrector weight. The CAB correction is an extrapolation-defect term, which measures how far the current learned velocity deviates from its linear extrapolation using previous velocities, and feeds this error back into the update with weight \(\gamma_i\). Thus, the correction vanishes when the learned field varies linearly across steps, and becomes active when the learned field has high non-linear variation, which is exactly when Adams--Bashforth becomes error-prone (refer Appendix~\ref{sec:correction_advantage}). In particular, the correction is designed to refine the Adams--Bashforth prediction while keeping the solver explicit, lightweight, and single-evaluation per step. The complete CAB procedures with two-step AB2 and three-step AB3 predictors, denoted CAB-2 and CAB-3 respectively, are given in Algorithm~\ref{alg:cabp_main}. We next show that the proposed CAB schemes retain the formal order of accuracy of the underlying Adams--Bashforth methods  \cite{hairer1993ode1,butcher2000ode20thcentury} despite the added correction.

\begin{theorem}[Accuracy of CAB-2 and CAB-3]
\label{thm:cab_accuracy}
Consider the rectified ODE \eqref{eq:rectified_rho_ode}, and let the numerical approximations
$\{\y_{t_i}\}_{i=0}^N$ be generated by Algorithm~\ref{alg:cabp_main} on the induced grid
$
\rho_{t_0}>\rho_{t_1}>\cdots>\rho_{t_N}.
$
Assume that the rectified velocity field $\epsilon_\theta$ in \eqref{eq:rectified_rho_ode} is Lipschitz in $\y$ uniformly in $\rho$, the exact solution $\hat{\y}(\rho)$ is sufficiently smooth, the variable-step ratios are uniformly bounded away from \(0\) and \(\infty\), the corrector weights are uniformly bounded, and the starting values are accurate to the target global order. Then the corrected variable-step CAB schemes satisfy
\begin{equation}
\label{eq:cab_accuracy_result_cases}
\max_{0\le i\le N}\|\y_{t_i}-\hat{\y}(\rho_{t_i})\|
\le
\begin{cases}
C_2\,h_{\max}^{2}, & \text{for CAB-2},\\[1mm]
C_3\,h_{\max}^{3}, & \text{for CAB-3 if } \gamma_i = \mathcal{O}(h_i),
\end{cases}
\end{equation}
where $h_{\max}:=\max_i |\rho_{t_{i+1}}-\rho_{t_i}|$ and the constants \(C_2\) and \(C_3\) are  independent of the step sizes \(\{h_i\}\). If used with a constant corrector weight \(\gamma_i=\mathcal{O}(1)\), CAB-3 is second-order globally.   Moreover, the local truncation errors are \(\mathcal{O}(h_i^{3})\) for CAB-2 and \(\mathcal{O}(h_i^{4})\) for CAB-3 when \(\gamma_i=\mathcal{O}(h_i)\). 
\end{theorem}

We defer the proof to Appendix~\ref{thmproof:cab_accuracy}, where we present a unified analysis in which CAB-2 and CAB-3 arise as special cases. Thus, CAB retains the standard convergence behavior of explicit multistep integration while improving low-NFE generation quality in practice, as shown in Section~\ref{sec:experiments}. We next provide a detailed comparison of our proposed solver with existing solvers in Table~\ref{tab:solver_comparison}, with respect to the evolution of the updates and NFEs required. 

\begin{table}[t]
\centering
\caption{Comparison of fast flow and diffusion samplers with CAB.}
\label{tab:solver_comparison}

\scriptsize
\setlength{\tabcolsep}{3.2pt}
\renewcommand{\arraystretch}{0.88}

\begin{tabularx}{\textwidth}{
@{}
>{\raggedright\arraybackslash}p{2.55cm}
>{\raggedright\arraybackslash}X
>{\raggedright\arraybackslash}p{5.0cm}
@{}
}
\toprule
\textbf{Method (NFE regime)}
& \textbf{Integration space; update mechanism}
& \textbf{Distinction from CAB}
\\
\midrule

\textbf{PNDM / PLMS} (20--50)
& DDIM path; AB/PLMS history extrapolation
& {History extrapolation only}; no field-defect correction
\\

\textbf{DEIS} (10--20)
& $t$ or $\rho$; polynomial noise extrapolation with exponential integration
& Related $\rho$-space prediction, but {no defect-based correction}
\\

\textbf{DPM-Solver} (10--20)
& $-log(\rho)$; high-order exponential / multistep integration
& Uses {high-order integration} rather than history-based defect correction
\\

\textbf{DPM-Solver++} (10--20)
& $-log(\rho)$; data-prediction exponential multistep solver
& No {adaptive defect-controlled correction weight}
\\

\textbf{UniPC} (5--20)
& $-log(\rho)$; multistep predictor + zero-NFE UniC corrector
& Uses an order-dependent weighted combination of field differences; CAB uses one extrapolation residual scaled by a single scalar
\\

\textbf{DPM-Solver-v3} (5--10)
& $-log(\rho)$; multistep residual approximation
& Uses {model-specific offline statistics}
\\

\textbf{STORK} (4--50)
& Original ODE; Taylor-predicted RK stages
& Predicts {stages/derivatives} rather than correcting a field defect
\\

\textbf{Conjugate Integrators} ($\geq 30$)
& Transformed state; time-varying linear transformation
& Requires a {chosen transformation} of the dynamics
\\

\textbf{Bespoke Solvers} (8--20)
& Learned coordinates; model- and NFE-specific optimized solver
& Requires {solver optimization for each setting}
\\

\midrule

\textbf{CAB} (6--20)
& $\mathbf{y}=\mathbf{x}/s,\ \rho=\sigma/s$;
{history-based field-defect correction}
& Uses one defect scaled by a fixed or adaptive scalar;
no evaluation at the predicted state or offline model statistics
\\

\bottomrule
\end{tabularx}

\vspace{-2mm}
\end{table}

\paragraph{Implementation details.}
Since CAB is a multistep method, the initial updates are handled separately as in Algorithm~\ref{alg:cabp_main}, where we use an Euler step for \(i=0\) and an AB2 step for \(i=1\). Though Algorithm~\ref{alg:cabp_main} is written in terms of a noise-prediction field, when the pretrained network uses data prediction, as in DPM-Solver++~\cite{lu2023dpmsolverpp}, or direct velocity prediction, as in rectified flow~\cite{liu2023rectified_flow}, we use the equivalent noise-field expressions in \eqref{eq:rectified_param_relations}. Because sampling proceeds from \(t=T\) to \(t=0\), the induced noise-to-signal coordinate \(\rho_t=\sigma_t/s_t\) can produce a numerically negligible terminal step. We therefore merge the last two \(\rho\)-steps in our implementation. This affects only the grid construction; all reported NFEs are counted by the number of network evaluations, not by the number of grid intervals. Finally, Theorem~\ref{thm:cab_accuracy} shows that CAB-3 preserves third-order global accuracy when the corrector weight is
\(\gamma_i=\mathcal{O}(h_i)\). In practice, however, a constant weight \(\gamma_i=\gamma\) gives better low-NFE sample quality. We therefore select a single model-specific \(\gamma\in(0,1.5)\) and keep it fixed across all sampling steps and generated samples for both CAB-2 and CAB-3.

\begin{figure*}[!htp]
    \centering
    \begin{subfigure}[t]{0.32\textwidth}
        \centering
        \includegraphics[width=\linewidth]{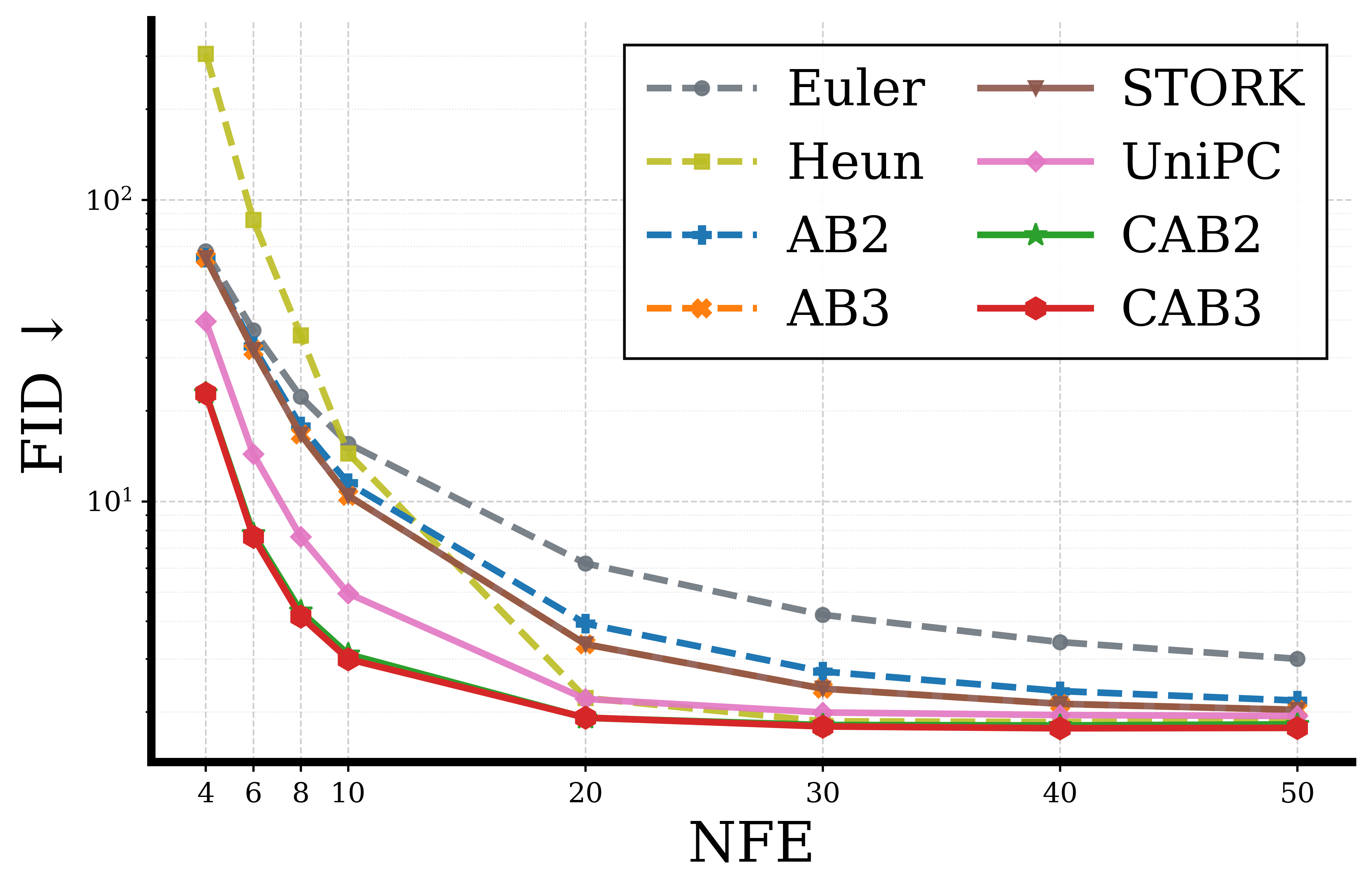}
        \caption{VP schedule on CIFAR-10.}      
        \label{fig:fid-vs-nfe-vp}
    \end{subfigure}
    \hfill
    \begin{subfigure}[t]{0.32\textwidth}
        \centering
        \includegraphics[width=\linewidth]{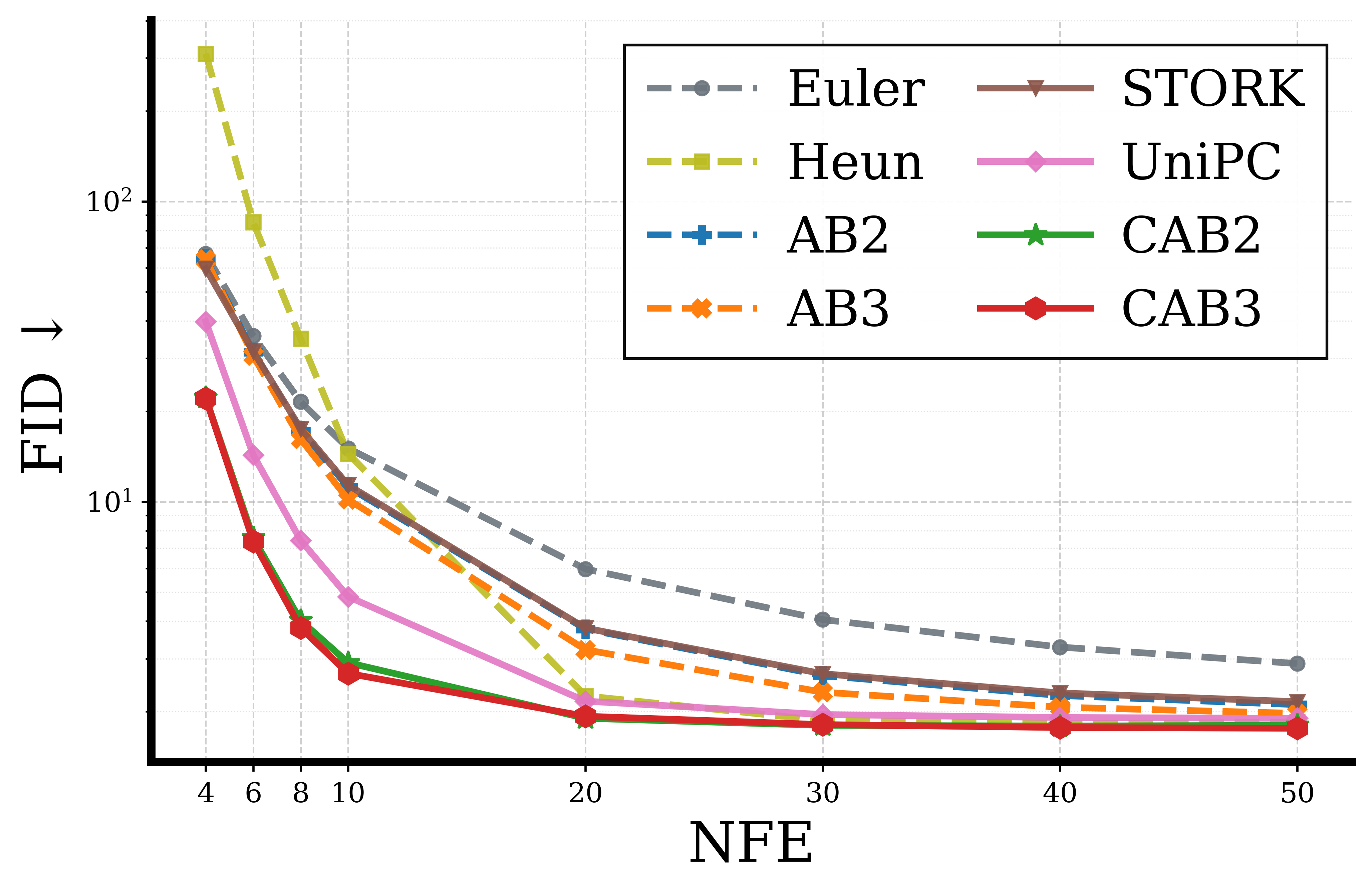}
        \caption{VE schedule on CIFAR-10.}
        \label{fig:fid-vs-nfe-ve}
    \end{subfigure}
    \hfill
    \begin{subfigure}[t]{0.32\textwidth}
        \centering
        \includegraphics[width=\linewidth]{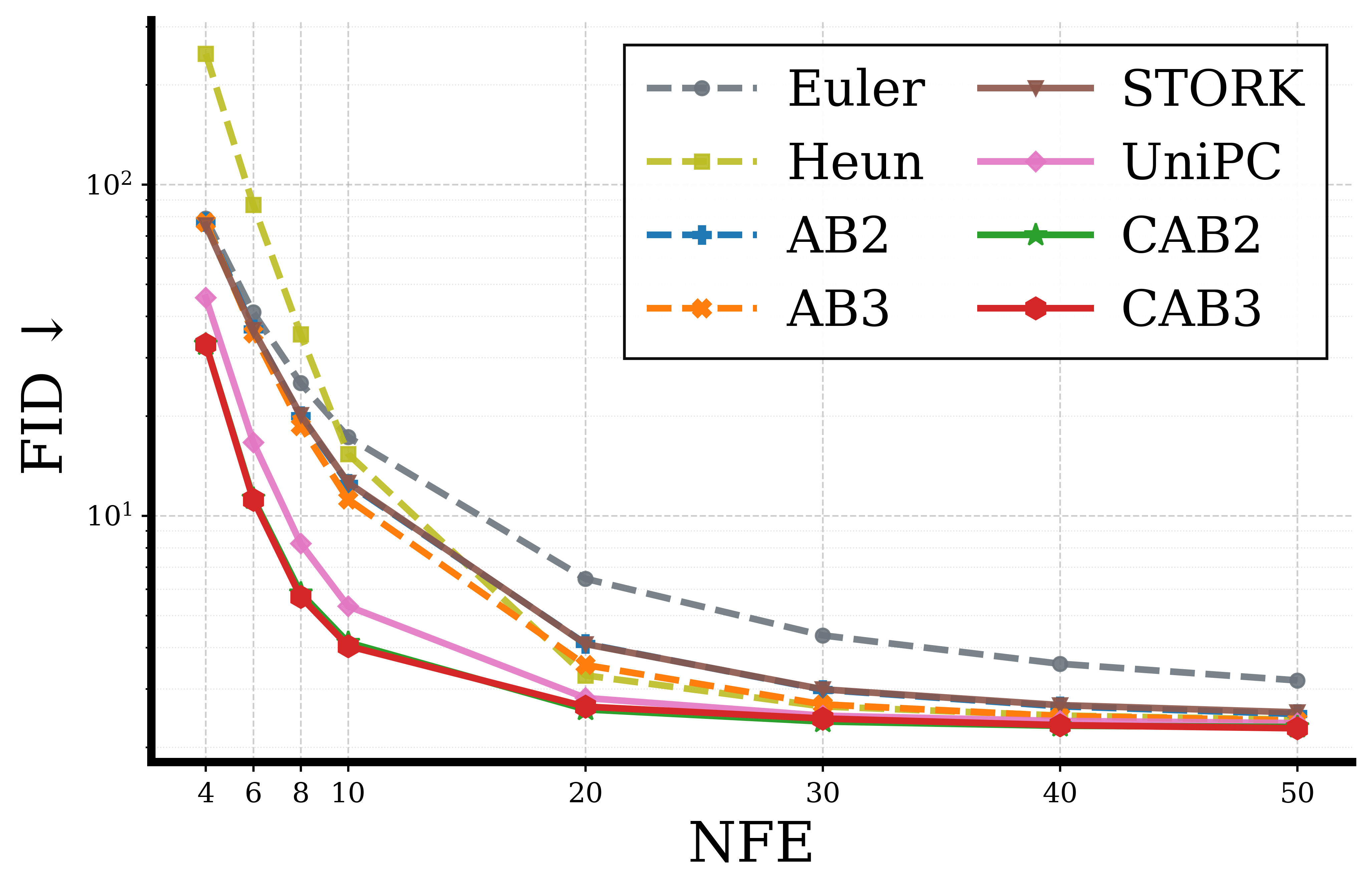}
        \caption{VP schedule on ImageNet.}
        \label{fig:fid-vs-nfe-imagenet}
    \end{subfigure}
\caption{FID versus NFE for CIFAR-10 with VP/VE schedules and ImageNet with the VP schedule. CAB-2 and CAB-3 consistently improve over the corresponding AB baselines in the low-NFE regime. They remain competitive with, and often outperform, strong training-free fast samplers.}
    \label{fig:fid-vs-nfe-three-plots}
\end{figure*}
\section{Experiments}
\label{sec:experiments}
In this section, we compare CAB with fast training-free flow and diffusion samplers such as DPM-Solver++, UniPC, and STORK, across multiple generation settings. We also provide ablations on the roles of rectification and correction, the effect of the corrector weight \(\gamma\) on distributional and perceptual quality, runtime, and memory cost. Additional results, such as empirical verification of Theorem~\ref{thm:cab_accuracy}, additional comparisons, and limitations, are deferred to the Appendix.

\begin{figure}[t]
    \centering

    \begin{subfigure}[b]{0.24\textwidth}
        \centering
        \includegraphics[width=\textwidth]{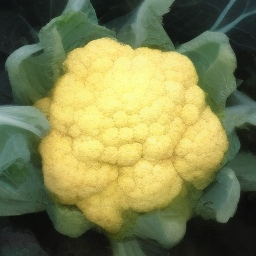}
        \caption{DDIM}
    \end{subfigure}
    \hfill
    \begin{subfigure}[b]{0.24\textwidth}
        \centering
        \includegraphics[width=\textwidth]{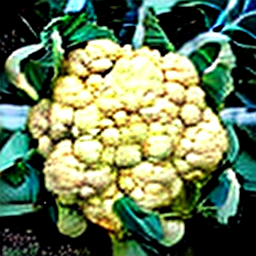}
        \caption{DPM++}
    \end{subfigure}
    \hfill
    \begin{subfigure}[b]{0.24\textwidth}
        \centering
        \includegraphics[width=\textwidth]{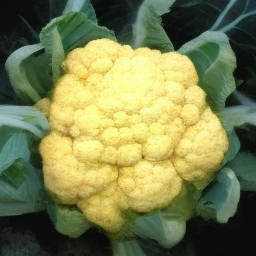}
        \caption{STORK}
    \end{subfigure}
    \hfill
    \begin{subfigure}[b]{0.24\textwidth}
        \centering
        \includegraphics[width=\textwidth]{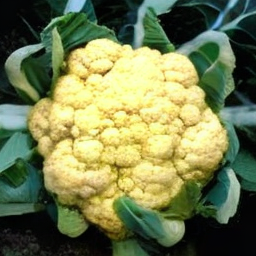}
        \caption{CAB-2}
    \end{subfigure}
\caption*{\textbf{DiT at 8 NFEs.} CAB-2 retains the cauliflower structure with sharper boundaries and no texture artifacts. Other samplers show oversmoothing, high-frequency artifacts, or distorted local texture.}

\vspace{0.2em}
    \begin{subfigure}[b]{0.24\textwidth}
        \centering
        \includegraphics[width=\textwidth]{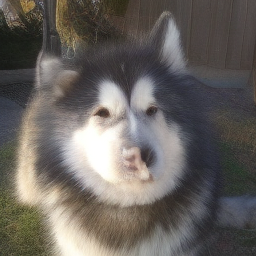}
        \caption{DDIM}
    \end{subfigure}
    \hfill
    \begin{subfigure}[b]{0.24\textwidth}
        \centering
        \includegraphics[width=\textwidth]{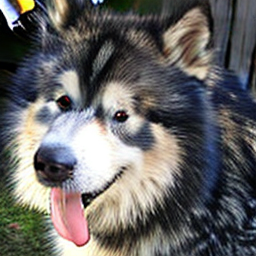}
        \caption{DPM++}
    \end{subfigure}
    \hfill
    \begin{subfigure}[b]{0.24\textwidth}
        \centering
        \includegraphics[width=\textwidth]{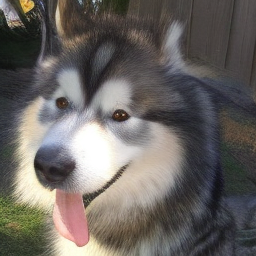}
        \caption{STORK}
    \end{subfigure}
    \hfill
    \begin{subfigure}[b]{0.24\textwidth}
        \centering
        \includegraphics[width=\textwidth]{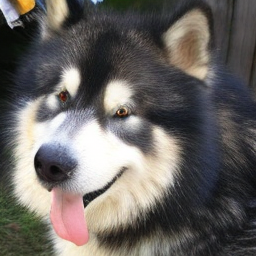}
        \caption{CAB-2}
    \end{subfigure}
\caption*{\textbf{DiT at 10 NFEs.} CAB-2 preserves sharper facial detail and more coherent structure, DDIM and STORK are noticeably blurred, while DPM-Solver++ introduces stronger texture artifacts.}
\caption{Comparison of training-free samplers on \(256\times256\) class-conditional ImageNet generation.}    
    \label{fig:dit_comparison}
\end{figure}

\begin{table*}[t]
\centering
\caption{FID comparison on DiT/ImageNet \(256\times256\), where lower is better. Sampler-O3 denotes the corresponding third-order variant using three past velocity evaluations, analogous to CAB-3.}
\label{tab:dit_imagenet_256}
\scriptsize
\setlength{\tabcolsep}{4pt}
\renewcommand{\arraystretch}{1.0}

\resizebox{\textwidth}{!}{%
\begin{tabular}{c|ccccccccc}
\hline
\textbf{NFE}
& \textbf{Euler}
& \textbf{DPM++-O3}
& \textbf{DPM++}
& \textbf{STORK-O3}
& \textbf{STORK}
& \textbf{AB2}
& \textbf{AB3}
& \textbf{CAB-2}
& \textbf{CAB-3} \\
\hline

$4$
& $120.83$
& $106.64$
& $\mathbf{75.26}$
& $127.99$
& $117.12$
& $116.99$
& $116.77$
& $88.04$
& $87.82$ \\

$6$
& $73.20$
& $96.38$
& $26.64$
& $100.23$
& $52.27$
& $51.52$
& $46.44$
& $15.13$
& $\mathbf{14.81}$ \\

$8$
& $39.71$
& $144.08$
& $24.90$
& $78.00$
& $21.18$
& $20.70$
& $16.75$
& $\mathbf{9.21}$
& $9.72$ \\

$10$
& $24.22$
& $141.58$
& $22.76$
& $57.62$
& $12.65$
& $12.39$
& $10.35$
& $\mathbf{7.99}$
& $8.79$ \\

$20$
& $9.94$
& $40.38$
& $12.19$
& $17.98$
& $7.48$
& $7.44$
& $7.21$
& $\mathbf{6.86}$
& $7.08$ \\

$30$
& $8.21$
& $15.06$
& $9.34$
& $11.49$
& $7.10$
& $7.09$
& $7.05$
& $\mathbf{6.86}$
& $6.93$ \\

$40$
& $7.63$
& $9.59$
& $8.42$
& $9.44$
& $7.02$
& $7.01$
& $6.98$
& $\mathbf{6.91}$
& $6.97$ \\

$50$
& $7.40$
& $8.02$
& $7.99$
& $8.53$
& $6.99$
& $6.98$
& $6.97$
& $\mathbf{6.95}$
& $7.01$ \\
\hline

\end{tabular}%
}
\end{table*}

\paragraph{Unconditional image generation on CIFAR-10 (\(32\times32\)) and ImageNet (\(64\times64\))} 

We evaluate CAB on three pixel-space EDM models from~\cite{karras2022edm}: CIFAR-10 with VP and VE schedules, and ImageNet with VP schedule. For each sampler, we generate \(50\)K images and report Fr\'echet Inception Distance (FID) as a function of NFE in Figure~\ref{fig:fid-vs-nfe-three-plots}. In the low-NFE regime, CAB-2 and CAB-3 consistently improve over their uncorrected Adams-Bashforth counterparts with $\gamma=0.9$, demonstrating that the proposed correction improves finite-step sampling accuracy. CAB remains effective across both VP and VE parameterizations and often outperforms strong training-free baselines such as UniPC and STORK at small NFE budgets. As the number of NFEs increases, CAB remains competitive while retaining a simple one-network-evaluation-per-step update.

\begin{figure}[t]
    \centering

    \begin{subfigure}[b]{0.24\textwidth}
        \centering
        \includegraphics[width=\textwidth]{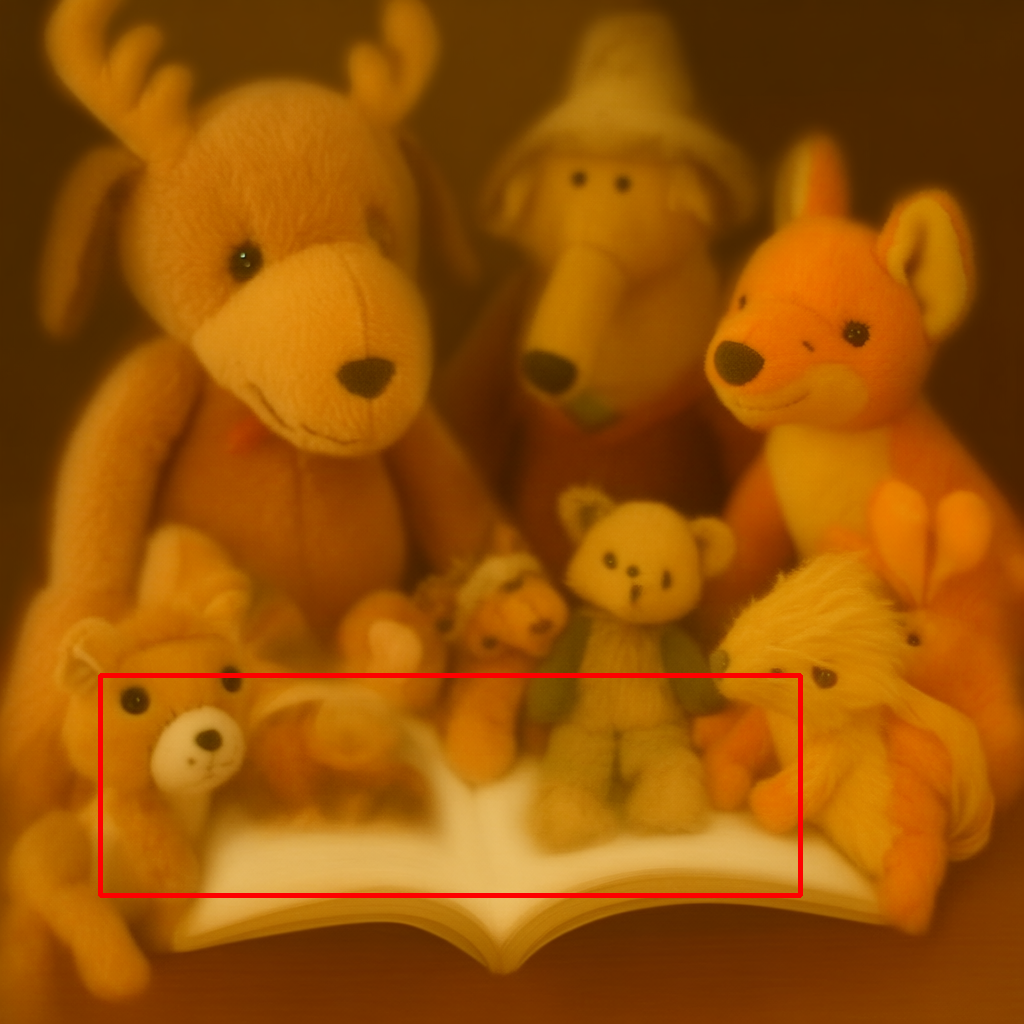}
        \caption{Euler}
    \end{subfigure}
    \hfill
    \begin{subfigure}[b]{0.24\textwidth}
        \centering
        \includegraphics[width=\textwidth]{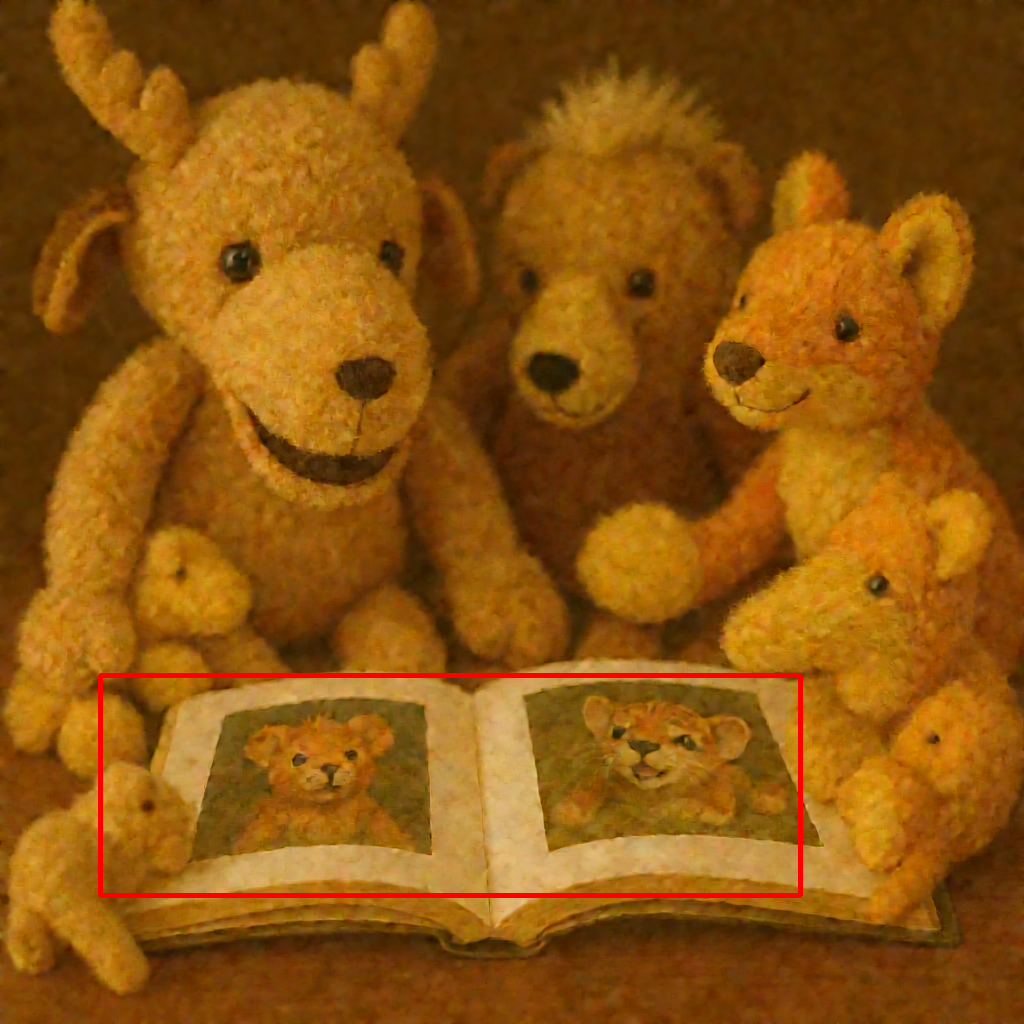}
        \caption{DPM++}
    \end{subfigure}
    \hfill
    \begin{subfigure}[b]{0.24\textwidth}
        \centering
        \includegraphics[width=\textwidth]{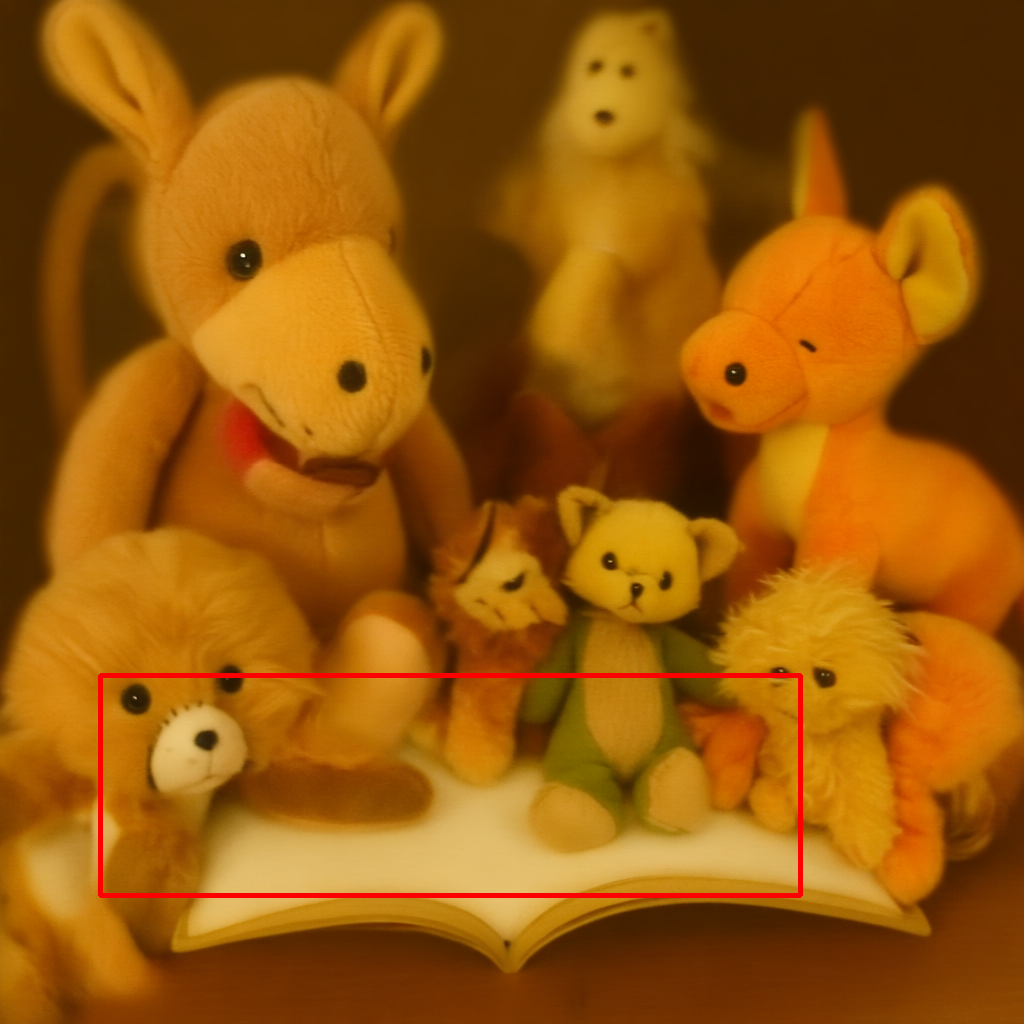}
        \caption{STORK}
    \end{subfigure}
    \hfill
    \begin{subfigure}[b]{0.24\textwidth}
        \centering
        \includegraphics[width=\textwidth]{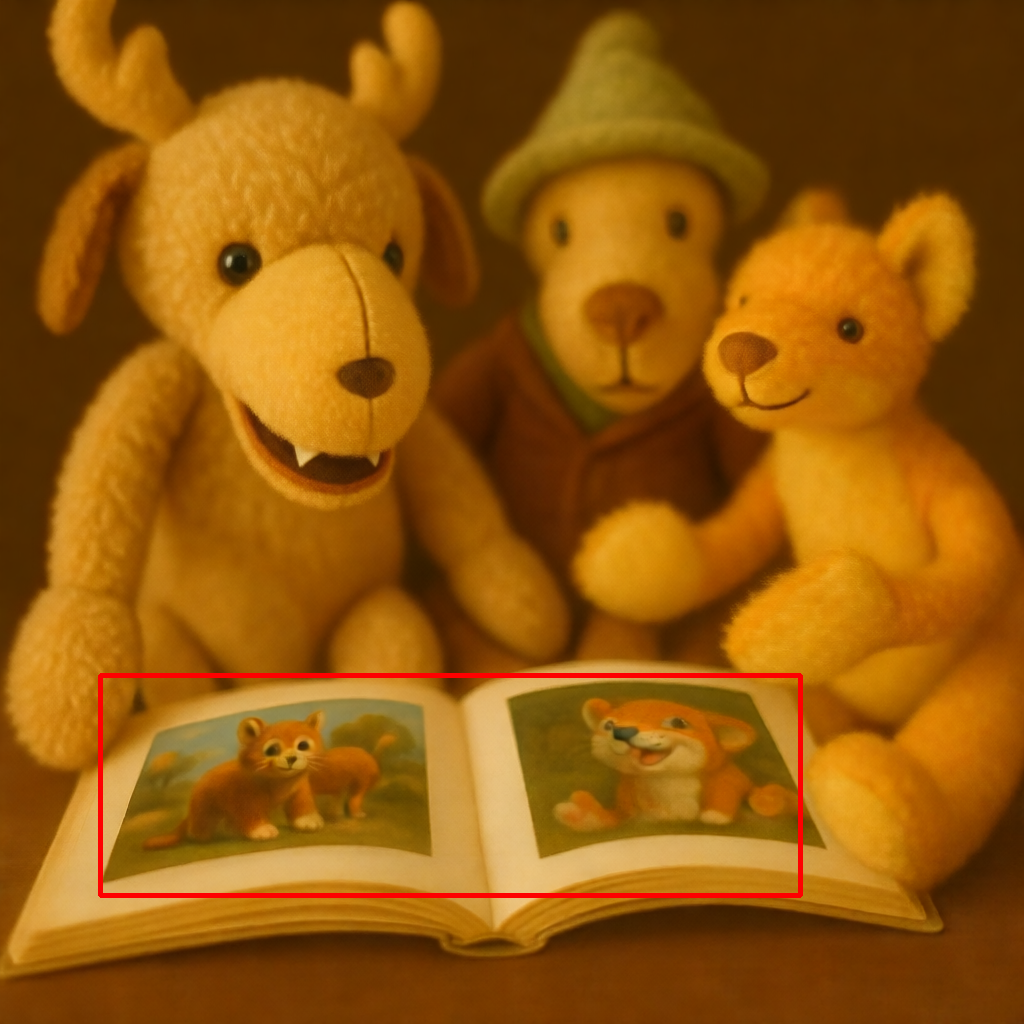}
        \caption{CAB-2}
    \end{subfigure}

    \caption*{\textbf{Prompt}: \textit{``Stuffed animals looking a pictures of other animals in a book.''} At 8 NFEs, CAB-2 preserves the context ' animals in book', while other outputs contain blur, noise, or semantic loss.}

    \begin{subfigure}[b]{0.24\textwidth}
        \centering
        \includegraphics[width=\textwidth]{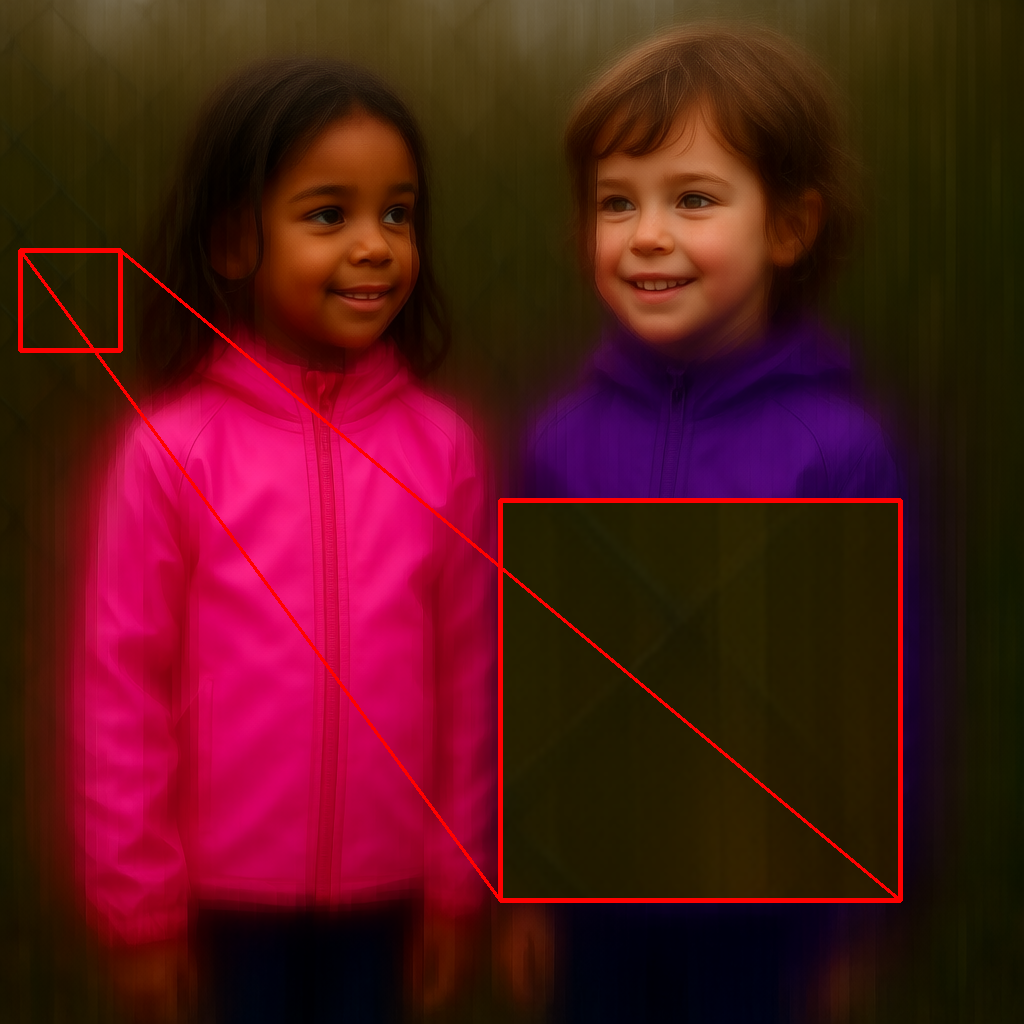}
        \caption{Euler}
    \end{subfigure}
    \hfill
    \begin{subfigure}[b]{0.24\textwidth}
        \centering
        \includegraphics[width=\textwidth]{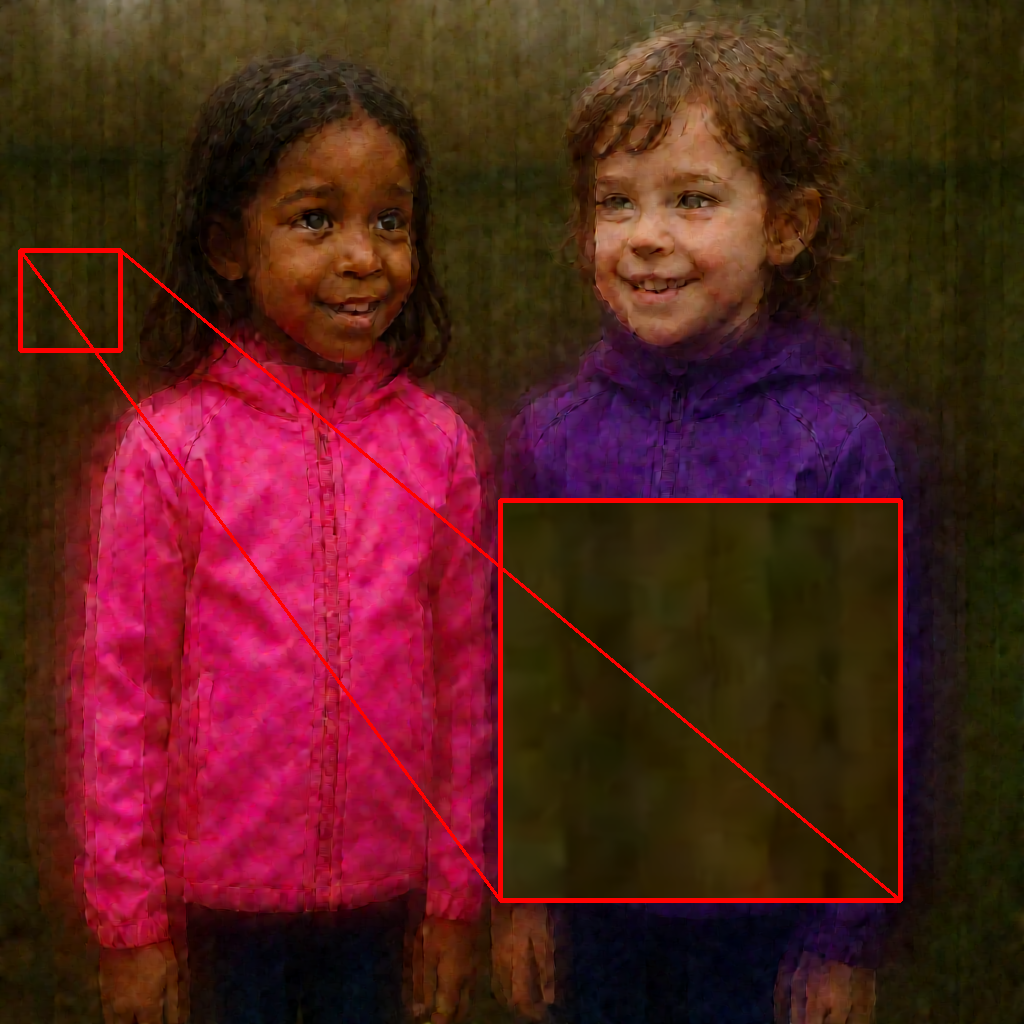}
        \caption{DPM++}
    \end{subfigure}
    \hfill
    \begin{subfigure}[b]{0.24\textwidth}
        \centering
        \includegraphics[width=\textwidth]{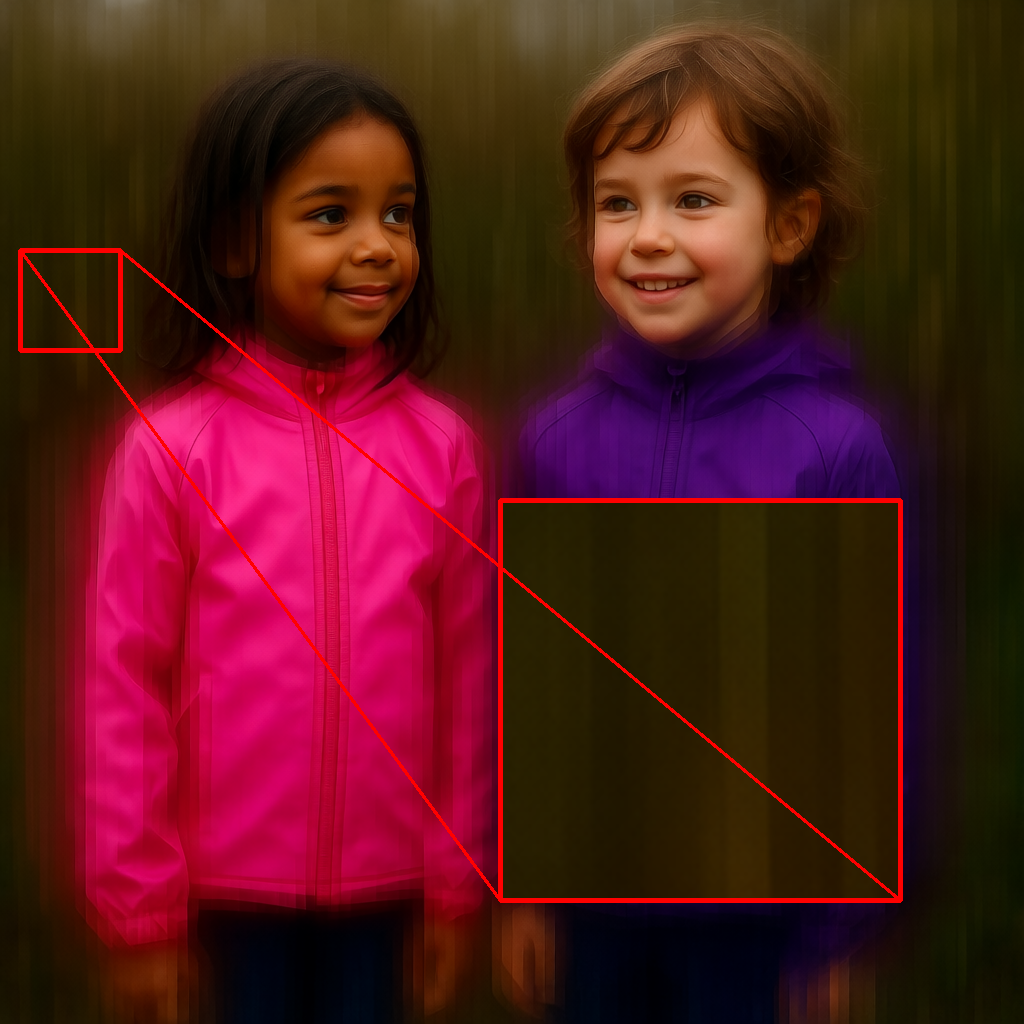}
        \caption{STORK}
    \end{subfigure}
    \hfill
    \begin{subfigure}[b]{0.24\textwidth}
        \centering
        \includegraphics[width=\textwidth]{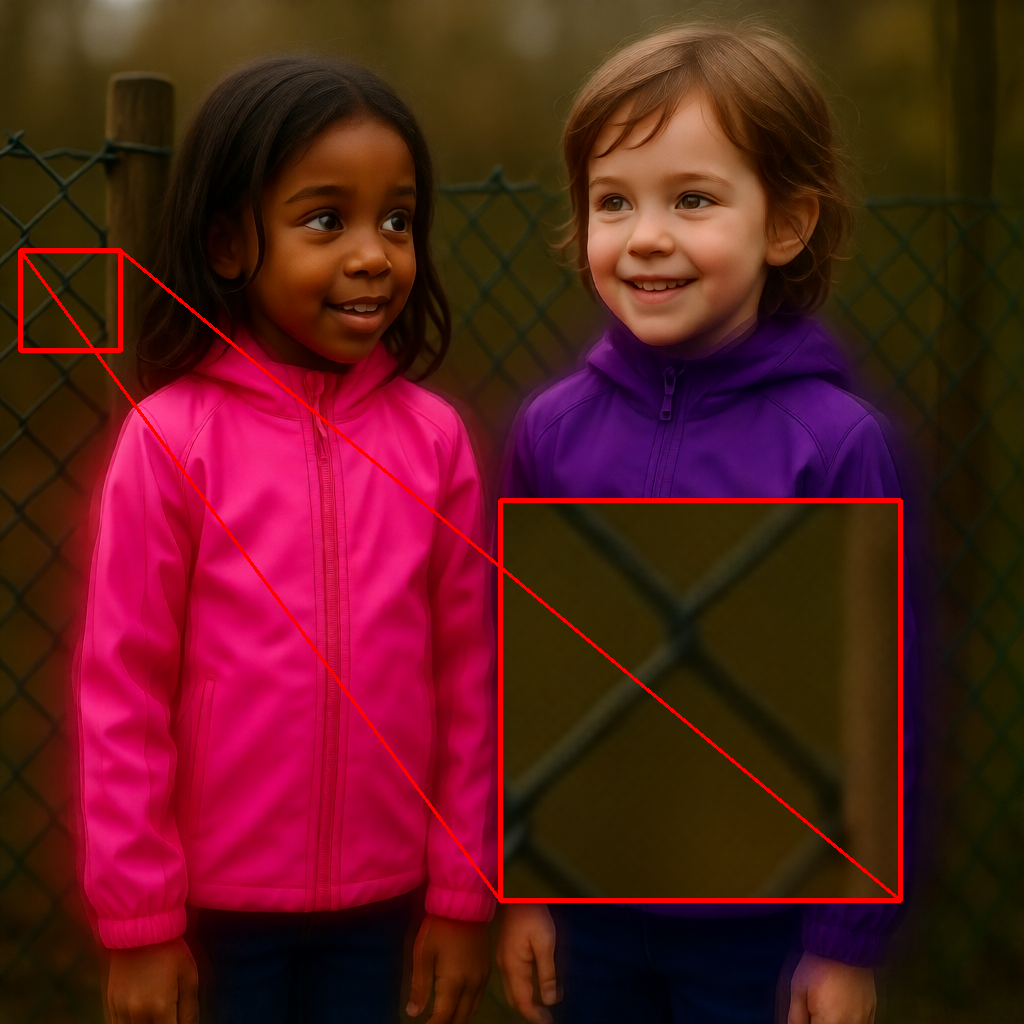}
        \caption{CAB-2}
    \end{subfigure}

    \caption*{\textbf{Prompt}: \textit{``Two kids in pink and purple jackets standing by a fence.''} At 8 NFEs, CAB-2 generates sharper details and preserves 'fence', while competing samplers show blur, noise, or missing structure.}
    
    \caption{Comparison of training-free samplers on QWEN-Image \(1024\times1024\) flow generation~\cite{wu2025qwen}.}
    
    \label{fig:qwen_comparison}

\end{figure}

\begin{table*}[t]
\centering
\caption{ImageReward Comparison on Qwen-Image in the low-NFE regime. CAB-3 (with \(\gamma=0.2\)) is strongest at \(4\!-\!10\) NFEs, while CAB-2 remains highly competitive and achieves the best score at \(N=20\); at larger NFEs, the gap to the strongest baseline becomes small.}
\label{tab:qwen_imagereward}
\vspace{0.3em}
\scriptsize
\setlength{\tabcolsep}{7pt}
\renewcommand{\arraystretch}{1.0}
\resizebox{\textwidth}{!}{%
\begin{tabular}{c|ccccccccc}
\hline
\textbf{NFE}
& \textbf{Euler}
& \textbf{Heun}
& \textbf{DPM-Solver++}
& \textbf{UniPC}
& \textbf{STORK}
& \textbf{AB2}
& \textbf{AB3}
& \textbf{CAB-2}
& \textbf{CAB-3} \\
\hline

$4$
& $-0.7949$
& $-1.1742$
& $-1.3069$
& $-0.6505$
& $-0.5147$
& $-0.6547$
& $-0.6414$
& $-0.3544$
& $\mathbf{-0.3474}$ \\

$6$
& $-0.1279$
& $-0.2009$
& $0.0455$
& $0.1377$
& $0.1306$
& $0.0637$
& $0.0975$
& $0.3393$
& $\mathbf{0.3798}$ \\

$8$
& $0.3439$
& $0.1499$
& $0.5866$
& $0.5471$
& $0.5512$
& $0.4996$
& $0.5183$
& $0.6561$
& $\mathbf{0.6760}$ \\

$10$
& $0.6499$
& $0.4081$
& $0.8069$
& $0.7671$
& $0.7734$
& $0.7482$
& $0.7662$
& $0.8689$
& $\mathbf{0.8856}$ \\

$20$
& $1.1545$
& $1.0057$
& $1.1912$
& $1.1682$
& $1.1756$
& $1.1686$
& $1.1768$
& $\mathbf{1.2026}$
& $1.1921$ \\

$30$
& $1.2644$
& $1.1344$
& $1.2729$
& $1.2651$
& $\mathbf{1.2855}$
& $1.2809$
& $1.2843$
& $1.2655$
& $1.2584$ \\

$40$
& $1.2806$
& $1.2543$
& $1.2669$
& $1.2756$
& $1.2654$
& $1.2838$
& $\mathbf{1.2852}$
& $1.2707$
& $1.2746$ \\

$50$
& $1.2824$
& $1.2738$
& $1.2840$
& $1.2839$
& $\mathbf{1.2934}$
& $1.2818$
& $1.2844$
& $1.2866$
& $1.2920$ \\
\hline

\end{tabular}
}
\vspace{-0.4em}
\end{table*}

\paragraph{Conditional ImageNet \(256\times256\) generation with DiT.}
We next evaluate CAB on DiT for class-conditional ImageNet \(256\times256\) generation. Table~\ref{tab:dit_imagenet_256} reports FID using \(10\)K generated samples across NFE budgets with classifier guidance scale \(1.25\). For UniPC, DPM-Solver++, and STORK, we report the best-performing settings together with higher-order variants; in this setting, aggressive multi-evaluation corrections can overcorrect and degrade FID (also shown visually in Figure.~\ref{fig:qualitative_nfe} in  Appendix). CAB uses a more controlled correction and achieves the strongest low-NFE performance: CAB-3 is best at \(6\) NFEs, while setting $\gamma=0.9$ in CAB-2, it obtains the best FID from \(8\) to \(50\) NFEs.  Figure~\ref{fig:motivation} and Figure~\ref{fig:dit_comparison} provide qualitative comparisons using classifier guidance scale \(4\). CAB-2 with $\gamma=0.5$ produces sharper and more coherent samples, and baselines often exhibit oversmoothing or high-frequency texture artifacts, consistent with the quantitative trends in Figure~\ref{fig:fid-vs-nfe-three-plots} and  Table~\ref{tab:dit_imagenet_256}.

\paragraph{High-Resolution ($1024\times1024$) Text-to-Image Generation (QWEN-Image)}

We evaluate CAB for high-resolution text-to-image synthesis using the recent QWEN-Image model~\cite{wu2025qwen}. Table~\ref{tab:qwen_imagereward} reports ImageReward~\cite{xu2023imagereward}, a learned human-preference score, averaged over \(200\) prompts. CAB-3 achieves the best ImageReward score at \(N=4,6,8,\) and \(10\), while CAB-2 is best at \(N=20\). Both CAB variants with $\gamma=0.2$ substantially improve over the uncorrected AB2 baseline, highlighting the contribution of the correction term. As NFE increases, the gap narrows, and strong baselines such as STORK, DPM-Solver++, and UniPC marginally outperform CAB. The qualitative results in Figures~\ref{fig:motivation} and~\ref{fig:qwen_comparison} are consistent with the ImageReward trends, where, CAB-2 produces sharper samples with fewer noise artifacts, less oversmoothing, and better preservation of prompt-relevant objects such as the bus, racket, book, and fence  at \(6\)--\(8\) NFEs. In contrast, competing samplers often exhibit grainy textures, blurred structure, or partial semantic loss.

\paragraph{Text-to-video generation with HunyuanVideo.}
We evaluate CAB on HunyuanVideo, a text-to-video generator trained with flow matching~\cite{kong2024hunyuanvideo}. Table~\ref{tab:hunyuan_video_eval_ext} reports low-NFE results averaged over \(200\) EvalCrafter prompts~\cite{liu2024evalcrafter}, where  \(20\)-frame videos at \(512\times320\) resolution are generated. We evaluate visual quality, temporal consistency, and text--video alignment, where alignment combines CLIP similarity~\cite{radford2021learning} with BLIP-BLEU~\cite{li2022blip}. CAB-2 (\(\gamma=0.9\)) achieves the best final score for most NFE budgets, despite variation in individual metrics. It is particularly strong in text--video alignment at \(4\) and \(6\) NFEs, while matching strong baselines in visual quality and temporal consistency. These results suggest that CAB  improves low-NFE sampling for video generation, beyond image settings.

\begin{table*}[t]
\centering
\caption{HunyuanVideo evaluation. CAB-2 gives the best overall final score across most step budgets.}
\label{tab:hunyuan_video_eval_ext}
\vspace{0.2em}
\scriptsize
\setlength{\tabcolsep}{5pt}
\renewcommand{\arraystretch}{1.05}

\begin{tabular}{c l ccccc | c l ccccc}
\toprule
\textbf{Metric} & \textbf{Method}
& \multicolumn{5}{c|}{\textbf{NFE}}
& \textbf{Metric} & \textbf{Method}
& \multicolumn{5}{c}{\textbf{NFE}} \\
\cmidrule(lr){3-7}
\cmidrule(lr){10-14}
& & \textbf{4} & \textbf{6} & \textbf{8} & \textbf{10} & \textbf{20}
& & & \textbf{4} & \textbf{6} & \textbf{8} & \textbf{10} & \textbf{20} \\
\midrule

\multirow{4}{*}{\textbf{Visual}}
& DPM++ & 45.07 & 44.88 & 45.78 & 47.35 & 51.74
&
\multirow{4}{*}{\textbf{Align.}}
& DPM++ & 30.87 & 36.88 & 40.13 & 41.04 & 42.50 \\

& STORK & 45.18 & 47.44 & \textbf{49.35} & 50.47 & \textbf{53.09}
&
& STORK & 35.30 & 37.61 & 40.00 & 40.53 & 42.06 \\

& CAB-2 & \textbf{46.08} & \textbf{47.70} & 49.03 & \textbf{50.54} & 52.71
&
& CAB-2 & \textbf{37.92} & \textbf{39.80} & 40.97 & 42.12 & \textbf{43.17} \\

& CAB-3 & 46.03 & 47.05 & 47.88 & 49.33 & 51.53
&
& CAB-3 & 37.72 & 39.67 & \textbf{41.41} & \textbf{42.34} & 42.73 \\

\midrule

\multirow{4}{*}{\textbf{Temp.}}
& DPM++ & 62.22 & 62.88 & 63.04 & 63.13 & \textbf{63.13}
&
\multirow{4}{*}{\textbf{Final}}
& DPM++ & 138 & 146 & 149 & 152 & 157 \\

& STORK & \textbf{63.71} & \textbf{63.47} & \textbf{63.22} & \textbf{63.30} & 63.03
&
& STORK & 144 & 149 & 152 & 154 & 158 \\

& CAB-2 & 62.84 & 62.72 & 62.78 & 63.04 & 62.86
&
& CAB-2 & \textbf{147} & \textbf{150} & \textbf{153} & \textbf{156} & \textbf{159} \\

& CAB-3 & 62.73 & 62.42 & 62.45 & 62.73 & 62.72
&
& CAB-3 & 146 & 149 & 151 & 154 & 157 \\

\bottomrule
\end{tabular}

\vspace{-0.4em}
\end{table*}

\begin{figure*}[!htp]
\centering

\begin{minipage}[t]{0.66\textwidth}
\centering
\vspace{0pt}

\begin{subfigure}[t]{0.49\linewidth}
    \centering
    \includegraphics[width=\linewidth]{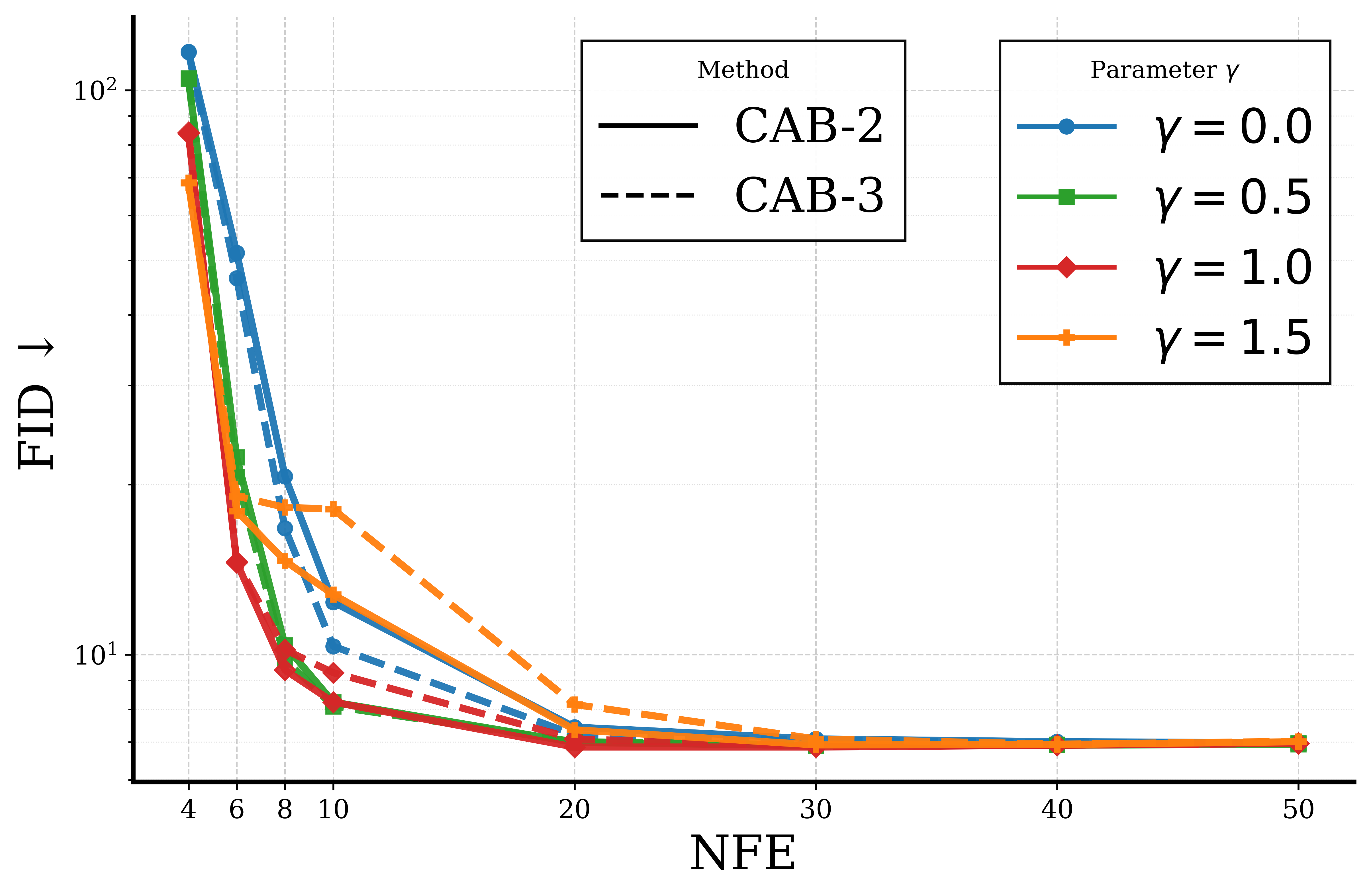}
    \caption{DiT on ImageNet \(256\times256\).}
\end{subfigure}
\hfill
\begin{subfigure}[t]{0.49\linewidth}
    \centering
    \includegraphics[width=\linewidth]{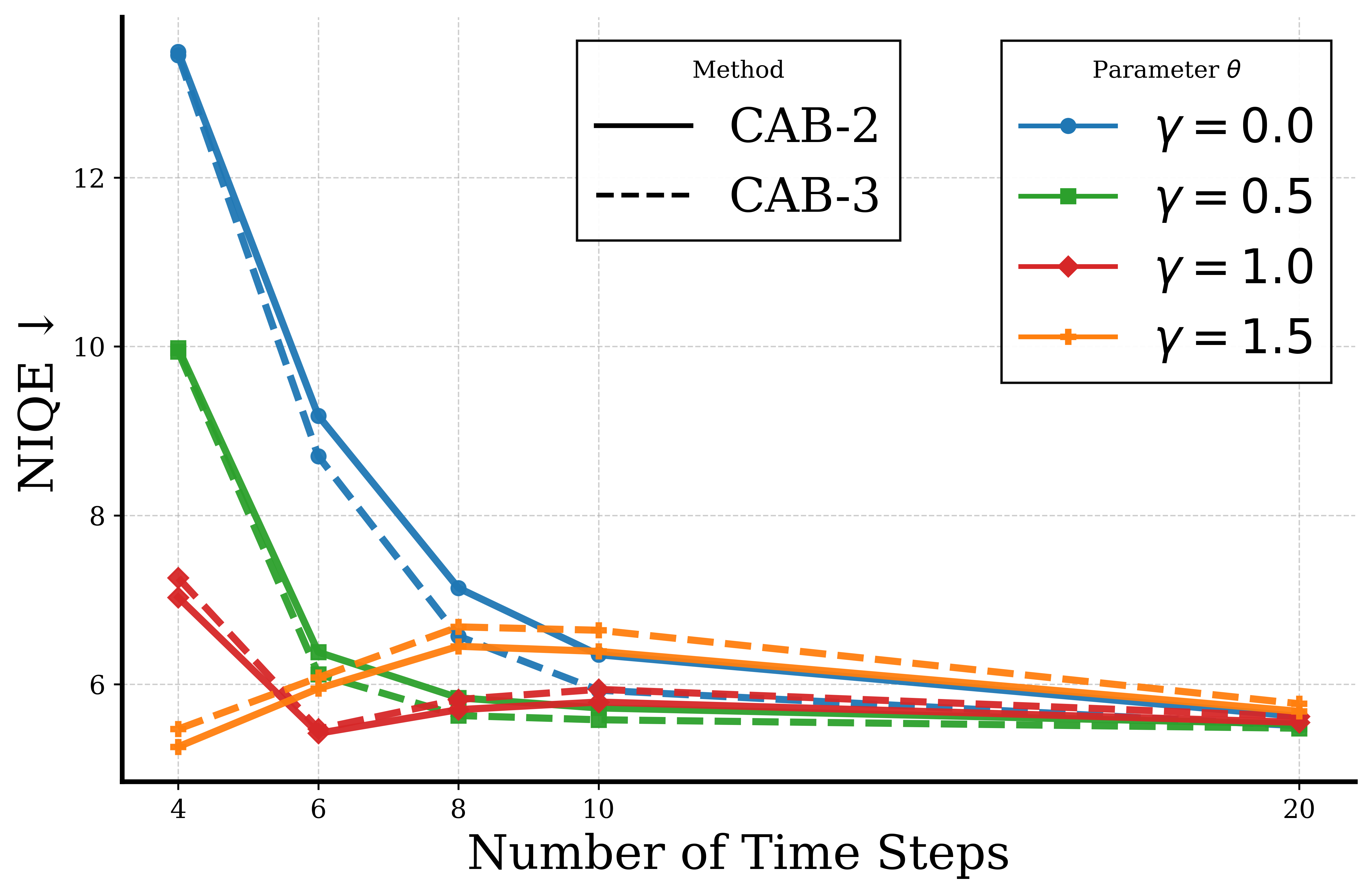}
    \caption{FID--NIQE trade-off.}
\end{subfigure}

\vspace{0.4em}
\captionof{figure}{Ablation of the CAB correction weight \(\gamma\) on DiT/ImageNet \(256\times256\). Stronger correction improves low-NFE FID, while moderate correction yields better NIQE.}
\label{fig:theta_ablation}
\end{minipage}
\hfill
\begin{minipage}[t]{0.30\textwidth}
\centering
\vspace{0pt}

\captionof{table}{Runtime (in sec) and maximum GPU memory allocated (in GB) across samplers for QWEN-Image.}
\label{tab:runtime_memory}

\vspace{0.3em}
\scriptsize
\setlength{\tabcolsep}{6pt}
\renewcommand{\arraystretch}{1.35}
\begin{tabular}{lcc}
\toprule
\textbf{Sampler} & \textbf{Time} & \textbf{GPU Mem.} \\
\midrule
STORK & $9.9620$ & $57.949$ \\
DPM++ & $9.9679$ & $57.940$ \\
AB2   & $9.9539$ & $57.941$ \\
AB3   & $9.9710$ & $57.942$ \\
CAB-2  & $9.9556$ & $57.942$ \\
CAB-3  & $9.9731$ & $57.942$ \\
\bottomrule
\end{tabular}

\end{minipage}

\end{figure*}

\begin{table*}[!htp]
\centering
\caption{
Transition-wise variation of the learned ODE field before and after noise-to-signal rectification.
Entries report normalized change per grid interval, with consecutive field change in parentheses.
}
\label{tab:rectification_field_variation}

\vspace{0.2em}

\tiny
\renewcommand{\arraystretch}{1.12}

\setlength{\tabcolsep}{12pt}
\begin{tabular}{c c | c c c c}

\toprule

\textbf{NFE}
& \textbf{Grid}
& $\mathbf{2\!\rightarrow\!3}$
& $\mathbf{3\!\rightarrow\!4}$
& $\mathbf{4\!\rightarrow\!5}$
& $\mathbf{5\!\rightarrow\!6}$ \\

\midrule

\multirow{2}{*}{\raisebox{-0.8ex}{$6$}}

& $t$-space
& $\;541.03\;(90.17)$
& $\;470.22\;(78.37)$
& $\;482.09\;(80.35)$
& $\;646.12\;(107.69)$ \\

& $\rho$-space
& $\mathbf{12.79\;(38.37)}$
& $\mathbf{39.61\;(39.61)}$
& $\mathbf{103.56\;(51.78)}$
& $\mathbf{280.98\;(84.29)}$ \\

\bottomrule
\end{tabular}

\vspace{0.01em}

\setlength{\tabcolsep}{2.3pt}
\begin{tabular}{c c | c c c c c c}
\toprule

\textbf{NFE}
& \textbf{Grid}
& $\mathbf{2\!\rightarrow\!3}$
& $\mathbf{3\!\rightarrow\!4}$
& $\mathbf{4\!\rightarrow\!5}$
& $\mathbf{5\!\rightarrow\!6}$
& $\mathbf{6\!\rightarrow\!7}$
& $\mathbf{7\!\rightarrow\!8}$ \\

\midrule

\multirow{2}{*}{\raisebox{-0.8ex}{$8$}}

& $t$-space
& $\;981.57\;(122.70)$
& $\;929.34\;(116.17)$
& $\;687.91\;(85.99)$
& $\;588.78\;(73.60)$
& $\;587.45\;(73.43)$
& $\;838.06\;(104.76)$ \\

& $\rho$-space
& $\mathbf{8.83\;(35.33)}$
& $\mathbf{32.98\;(43.98)}$
& $\mathbf{62.86\;(41.90)}$
& $\mathbf{119.28\;(47.71)}$
& $\mathbf{212.67\;(56.71)}$
& $\mathbf{486.83\;(92.73)}$ \\

\bottomrule
\end{tabular}

\vspace{-0.6em}

\end{table*}
\paragraph{Ablation studies}

Figure~\ref{fig:theta_ablation} studies the effect of the CAB correction weight \(\gamma\) on DiT/ImageNet model. Nonzero correction substantially improves low-NFE performance over the uncorrected case \(\gamma=0\), especially in the aggressive \(4\)-\(10\) NFE regime.  This supports the interpretation of the correction as an extrapolation-defect adjustment. The best value of \(\gamma\) is however metric-dependent. Stronger correction, e.g., \(\gamma=1.0\), can improve FID suggesting better agreement with  dataset-level distributional statistics under this metric.
In our NIQE evaluations, moderate correction $\gamma = 0.5$ is frequently preferred, indicating a trade-off between distributional metrics and perceptual metrics. Table~\ref{tab:rectification_field_variation} quantifies the effect of noise-to-signal rectification on field variation along QWEN-Image sampling trajectories, averaged over \(10\) prompts. In the original coordinates, ODE uses the learned velocity field, whereas after rectification the right-hand side becomes the learned noise field in Lemma~\ref{lem:rho_rectified}. The rectified \(\rho\)-space dynamics exhibit substantially smaller, more structured transition-wise changes at \(6\) and \(10\) NFEs, indicating a smoother field for multistep updates and supporting the role of rectification on improved low-NFE behavior of CAB (refer Appendix~\ref{effect:rectcorr} for more details). We compare runtime and memory on an A100 GPU for QWEN-Image \(1024\times1024\) generation at 10 NFEs, averaged over $10$ prompts with batch size $1$ and \textit{torch.bfloat16}. CAB has the same cost range as other samplers, with negligible memory overhead from storing one additional past evaluation.

\paragraph{Extension to Stochastic Sampling.}
We extend CAB to stochastic sampling by considering the
variance-controlled family in noise-to-signal coordinates,
\begin{equation}
d\mathbf{y}_{\rho}
= (1+\eta^2)\boldsymbol{\epsilon}_{\theta}
(s_{t(\rho)}\mathbf{y}_{\rho},t(\rho))\,d\rho
+ \eta\sqrt{2\rho}\,d\bar{\mathbf{W}}_{\rho},
\end{equation}
where $\eta=0$ recovers the deterministic probability-flow ODE,
$\eta=1$ corresponds to the full reverse SDE, and
$0<\eta<1$ interpolates between them.
We discretize the drift using CAB-2 and sample the Gaussian
increments analytically, retaining one network evaluation per step.
Table~\ref{tab:cab_stochastic_main} shows that CAB-2-SDE
substantially improves over DDIM-SDE~\cite{song2021ddim} on DiT/ImageNet-256
at $\eta=0.1$, reducing FID from $39.83$ to $19.45$
at $8$ NFEs and from $24.40$ to $12.00$ at $10$ NFEs.
Deterministic CAB-2 remains stronger, indicating that
stochastic perturbations can make few-step multistep
correction more challenging.

\begin{table}[t]
\centering

\begin{minipage}[t]{0.39\linewidth}
\vspace{0pt}
\centering
\captionsetup{justification=raggedright,
singlelinecheck=false}

\caption{Stochastic sampling on DiT/ImageNet-256
($\eta=0.1$). FID $\downarrow$.}
\label{tab:cab_stochastic_main}
\vspace{2pt}

{\scriptsize
\setlength{\tabcolsep}{3pt}
\renewcommand{\arraystretch}{1.24}

\begin{tabular}{@{}c|ccc@{}}
\toprule
\multirow{2}{*}{\textbf{NFE}}
& \multicolumn{3}{c}{\textbf{Sampler}} \\
\cmidrule(l){2-4}
& \textbf{DDIM}
& \textbf{CAB-SDE}
& \textbf{CAB-ODE} \\
\midrule
$8$  & $39.83$ & $19.45$ & $\mathbf{9.21}$ \\
$10$ & $24.40$ & $12.00$ & $\mathbf{7.99}$ \\
$15$ & $12.90$ & $8.09$  & $\mathbf{7.09}$ \\
\bottomrule
\end{tabular}
}
\end{minipage}%
\hfill
\begin{minipage}[t]{0.59\linewidth}
\vspace{0pt}
\centering
\captionsetup{justification=raggedright,
singlelinecheck=false}

\caption{Generalizability of the correction on
CIFAR-10 (VP). FID $\downarrow$.}
\label{tab:predictor_correction_main}
\vspace{2pt}

{\scriptsize
\setlength{\tabcolsep}{2pt}
\renewcommand{\arraystretch}{1.00}

\resizebox{\linewidth}{!}{%
\begin{tabular}{@{}c|cc|cc|cc|cc@{}}
\toprule
\multirow{2}{*}{\textbf{NFE}}
& \multicolumn{2}{c|}{\textbf{Euler}}
& \multicolumn{2}{c|}{\textbf{Heun}}
& \multicolumn{2}{c|}{\textbf{STORK}}
& \multicolumn{2}{c}{\textbf{AB}} \\
\cmidrule(lr){2-3}
\cmidrule(lr){4-5}
\cmidrule(lr){6-7}
\cmidrule(l){8-9}
& \textbf{Base} & \textbf{+Corr.}
& \textbf{Base} & \textbf{+Corr.}
& \textbf{Base} & \textbf{+Corr.}
& \textbf{AB2} & \textbf{CAB-2} \\
\midrule
$6$
& $36.94$ & $12.90$
& $85.83$ & $85.76$
& $32.66$ & $23.21$
& $23.78$ & $\mathbf{7.81}$ \\
$8$
& $22.21$ & $5.92$
& $35.57$ & $35.27$
& $18.03$ & $6.02$
& $14.42$ & $\mathbf{4.32}$ \\
$10$
& $15.54$ & $3.91$
& $14.44$ & $14.34$
& $11.65$ & $3.25$
& $9.99$ & $\mathbf{3.11}$ \\
\bottomrule
\end{tabular}%
}
}
\end{minipage}

\vspace{-0.5em}
\end{table}


\paragraph{Generalizability of the Correction.}
We evaluate the extrapolation-defect correction with Euler, Heun, and STORK predictors on CIFAR-10 (VP).
Table~\ref{tab:predictor_correction_main} shows substantial improvements for Euler and STORK, reducing FID from $15.54$ to $3.91$ and $11.65$ to $3.25$, respectively, at $10$ NFEs. Heun shows negligible gains, while CAB-2 consistently achieves the lowest FID, highlighting the effectiveness of the correction with history-based Adams--Bashforth prediction.


\section{Conclusion}

We introduced CAB, a simple training-free framework for accelerating flow and diffusion sampling by rectifying the reverse-time ODE associated with affine-Gaussian paths. In the resulting noise-to-signal coordinates, CAB applies corrected Adams--Bashforth updates with one network evaluation per step while preserving the second- or third-order accuracy of the underlying predictors. Across our evaluated image and video settings, CAB provides a favorable low-NFE trade-off, often improving perceptual quality or distributional metrics over strong training-free baselines.

\begin{ack}
This work was supported by the Anusandhan National Research Foundation (ANRF),
Government of India, under the Prime Minister Early Career Research Grant
(PM-ECRG), Grant No.~\texttt{ANRF/ECRG/2024/001835/ENS}.
\end{ack}

\paragraph{Broader impact.}
\label{sec:broader_impact}
CAB reduces the inference cost of flow and diffusion generation by improving sample quality at low NFE budgets. This can lower latency, energy use, and hardware requirements, making generative models more accessible in resource-constrained settings. However, faster sampling can also increase the scale at which synthetic or misleading visual content is generated. Deployment should therefore be accompanied by watermarking or misuse-detection mechanisms.


\bibliographystyle{unsrtnat}
\bibliography{refs}

@article{ho2020ddpm,
  title={Denoising diffusion probabilistic models},
  author={Ho, Jonathan and Jain, Ajay and Abbeel, Pieter},
  journal={Proc. Advances in neural information processing systems},
  volume={33},
  pages={6840--6851},
  year={2020}
}

@article{lipman2023flowmatching,
title={Flow Matching for Generative Modeling},
author={Yaron Lipman and Ricky T. Q. Chen and Heli Ben-Hamu and Maximilian Nickel and Matthew Le},
journal={Proc. International Conference on Learning Representations },
year={2023}
}

@article{ho2022videodiffusion,
  title     = {Video Diffusion Models},
  author    = {Ho, Jonathan and Salimans, Tim and Gritsenko, Alexey and Chan, William and Norouzi, Mohammad and Fleet, David J.},
  journal={Proc. Advances in neural information processing systems},
  volume={35},
  pages={8633--8646},
  year={2022}
}

@article{huang2025diffusioneditingsurvey,
  title   = {Diffusion Model-Based Image Editing: A Survey},
  author  = {Huang, Yi and Huang, Jiancheng and Liu, Yifan and Yan, Mingfu and Lv, Jiaxi and Liu, Jianzhuang and Xiong, Wei and Zhang, He and Chen, Shifeng and Cao, Liangliang},
  journal = {IEEE Transactions on Pattern Analysis and Machine Intelligence},
  volume={47},
  number={6},
  pages={4409-4437},
  year    = {2025}
}

@article{saharia2021palette,
  title   = {Palette: Image-to-Image Diffusion Models},
  author  = {Saharia, Chitwan and Chan, William and Chang, Huiwen and Lee, Chris A. and Ho, Jonathan and Salimans, Tim and Fleet, David J. and Norouzi, Mohammad},
  journal = {NeurIPS 2021 Workshop on Deep Generative Models and Downstream Applications},
  year    = {2021}
}

@article{lugmayr2022repaint,
  title     = {RePaint: Inpainting using Denoising Diffusion Probabilistic Models},
  author    = {Lugmayr, Andreas and Danelljan, Martin and Romero, Andr{\'e}s and Yu, Fisher and Timofte, Radu and Van Gool, Luc},
  journal = {Proc. IEEE/CVF Conference on Computer Vision and Pattern Recognition},
  year      = {2022}
}

@article{wang2025videodiffusionsurvey,
title={Survey of Video Diffusion Models: Foundations, Implementations, and Applications},
author={Yimu Wang and Xuye Liu and Wei Pang and Li Ma and Shuai Yuan and Paul Debevec and Ning Yu},
journal={Transactions on Machine Learning Research},
issn={2835-8856},
year={2025}
}

@article{li2023restorationenhancement,
  title={Diffusion models for image restoration and enhancement: A comprehensive survey},
  author={Li, Xin and Ren, Yulin and Jin, Xin and Lan, Cuiling and Wang, Xingrui and Zeng, Wenjun and Wang, Xinchao and Chen, Zhibo},
  journal={International Journal of Computer Vision},
  volume={133},
  number={11},
  pages={8078--8108},
  year={2025}
}

@article{song2021score_sde,
  title   = {Score-Based Generative Modeling through Stochastic Differential Equations},
  author={Yang Song and Jascha Sohl-Dickstein and Diederik P Kingma and Abhishek Kumar and Stefano Ermon and Ben Poole},
  journal = {Proc. International Conference on Learning Representations},
  year    = {2021}
}

@article{song2021ddim,
  title   = {Denoising Diffusion Implicit Models},
  author={Jiaming Song and Chenlin Meng and Stefano Ermon},
  journal = {Proc. International Conference on Learning Representations},
  year    = {2021}
}

@article{karras2022edm,
  title={Elucidating the design space of diffusion-based generative models},
  author={Karras, Tero and Aittala, Miika and Aila, Timo and Laine, Samuli},
  journal={Proc. Advances in neural information processing systems},
  volume={35},
  pages={26565--26577},
  year={2022}
}

@article{chen2018neural_ode,
  title   = {Neural Ordinary Differential Equations},
  author={Chen, Ricky TQ and Rubanova, Yulia and Bettencourt, Jesse and Duvenaud, David K},
  journal = {Proc. Advances in Neural Information Processing Systems},
  volume = {31},
  year = {2018}
}

@article{grathwohl2019ffjord,
  title   = {FFJORD: Free-form Continuous Dynamics for Scalable Reversible Generative Models},
  author={Will Grathwohl and Ricky T. Q. Chen and Jesse Bettencourt and David Duvenaud},
  journal = {Proc. International Conference on Learning Representations},
  year    = {2019}
}

@article{liu2023rectified_flow,
  title   = {Flow Straight and Fast: Learning to Generate and Transfer Data with Rectified Flow},
  author={Xingchao Liu and Chengyue Gong and qiang liu},
  journal = {Proc. International Conference on Learning Representations},
  year    = {2023}
}

@article{liu2024instaflow,
title={InstaFlow: One Step is Enough for High-Quality Diffusion-Based Text-to-Image Generation},
author={Xingchao Liu and Xiwen Zhang and Jianzhu Ma and Jian Peng and Qiang liu},
journal={Proc. International Conference on Learning Representations},
year={2024}
}

@article{starodubcev2025,
  title   = {Scale-wise Distillation of Diffusion Models},
  author={Starodubcev, Nikita and Drobyshevskiy, Ilya and Kuznedelev, Denis and Babenko, Artem and Baranchuk, Dmitry},
  journal={Proc. International Conference on Learning Representations},
  year    = {2026}
}

@article{lu2025,
  title   = {Simplifying, Stabilizing and Scaling Continuous-Time Consistency Models},
  author  = {Lu, Cheng and Song, Yang},
  journal={Proc. International Conference on Learning Representations},
  year    = {2025}
}

@article{yin2024,
  title     = {One-step Diffusion with Distribution Matching Distillation},
  author    = {Yin, Tianwei and Gharbi, Micha{\"e}l and Zhang, Richard and Shechtman, Eli and Durand, Fr{\'e}do and Freeman, William T. and Park, Taesung},
  journal = {Proc. IEEE/CVF Conference on Computer Vision and Pattern Recognition},
  pages={6613--6623},
  year      = {2024}
}

@article{chen2025onestep,
  title={SANA-Sprint: One-step diffusion with continuous-time consistency distillation},
  author={Chen, Junsong and Xue, Shuchen and Zhao, Yuyang and Yu, Jincheng and Paul, Sayak and Chen, Junyu and Cai, Han and Han, Song and Xie, Enze},
  journal={Proc. IEEE/CVF International Conference on Computer Vision},
  pages={16185--16195},
  year={2025}
}

@article{kingma2021variational,
  title={Variational diffusion models},
  author={Kingma, Diederik and Salimans, Tim and Poole, Ben and Ho, Jonathan},
  journal={Proc. Advances in neural information processing systems},
  volume={34},
  pages={21696--21707},
  year={2021}
}

@article{salimans2022progressive,
title={Progressive Distillation for Fast Sampling of Diffusion Models},
author={Tim Salimans and Jonathan Ho},
journal={Proc. International Conference on Learning Representations},
year={2022}
}

@article{song2023consistency_models,
  title={Consistency Models},
  author={Song, Yang and Dhariwal, Prafulla and Chen, Mark and Sutskever, Ilya},
  journal={Proc. International Conference on Machine Learning},
  pages={32211--32252},
  year={2023}
}

@article{lu2022dpmsolver,
  title={DPM-Solver: A fast ode solver for diffusion probabilistic model sampling in around 10 steps},
  author={Lu, Cheng and Zhou, Yuhao and Bao, Fan and Chen, Jianfei and Li, Chongxuan and Zhu, Jun},
  journal={Proc. Advances in neural information processing systems},
  volume={35},
  pages={5775--5787},
  year={2022}
}

@article{lu2023dpmsolverpp,
  title={DPM-Solver++: Fast solver for guided sampling of diffusion probabilistic models},
  author={Lu, Cheng and Zhou, Yuhao and Bao, Fan and Chen, Jianfei and Li, Chongxuan and Zhu, Jun},
  journal={Machine Intelligence Research},
  pages={1--22},
  year={2025}
}

@article{zhang2023fast,
title={Fast Sampling of Diffusion Models with Exponential Integrator},
author={Qinsheng Zhang and Yongxin Chen},
journal={Proc. International Conference on Learning Representations },
year={2023}
}

@article{zhao2023unipc,
  title={Unipc: A unified predictor-corrector framework for fast sampling of diffusion models},
  author={Zhao, Wenliang and Bai, Lujia and Rao, Yongming and Zhou, Jie and Lu, Jiwen},
  journal={Proc. Advances in Neural Information Processing Systems},
  volume={36},
  pages={49842--49869},
  year={2023}
}

@article{xue2023sa,
  title={Sa-solver: Stochastic adams solver for fast sampling of diffusion models},
  author={Xue, Shuchen and Yi, Mingyang and Luo, Weijian and Zhang, Shifeng and Sun, Jiacheng and Li, Zhenguo and Ma, Zhi-Ming},
  journal={Proc. Advances in Neural Information Processing Systems},
  volume={36},
  pages={77632--77674},
  year={2023}
}

@article{tan2026stork,
title={{STORK}: Faster Diffusion and Flow Matching Sampling by Resolving both Stiffness and Structure-Dependence},
author={Zheng Tan and Weizhen Wang and Andrea L. Bertozzi and Ernest K. Ryu},
journal={Proc. International Conference on Learning Representations},
year={2026}
}

@article{xu2023restart_sampling,
  title={Restart sampling for improving generative processes},
  author={Xu, Yilun and Deng, Mingyang and Cheng, Xiang and Tian, Yonglong and Liu, Ziming and Jaakkola, Tommi},
  journal={Proc. Advances in Neural Information Processing Systems},
  volume={36},
  pages={76806--76838},
  year={2023}
}

@article{liu2022pndm,
  title   = {Pseudo Numerical Methods for Diffusion Models on Manifolds},
  author={Luping Liu and Yi Ren and Zhijie Lin and Zhou Zhao},
  journal = {Proc. International Conference on Learning Representations},
  year    = {2022}
}

@article{zheng2023dpmsolverv3,
  title={Dpm-solver-v3: Improved diffusion ode solver with empirical model statistics},
  author={Zheng, Kaiwen and Lu, Cheng and Chen, Jianfei and Zhu, Jun},
  journal={Proc. Advances in Neural Information Processing Systems},
  volume={36},
  pages={55502--55542},
  year={2023}
}

@article{shaul2024bespoke,
title={Bespoke Solvers for Generative Flow Models},
author={Neta Shaul and Juan Perez and Ricky T. Q. Chen and Ali Thabet and Albert Pumarola and Yaron Lipman},
journal={Proc. International Conference on Learning Representations},
year={2024}
}

@article{frankel2025ss,
title={S4S: Solving for a Fast Diffusion Model Solver},
author={Eric Frankel and Sitan Chen and Jerry Li and Pang Wei Koh and Lillian J. Ratliff and Sewoong Oh},
journal = {Proc. International Conference on Machine Learning},
year={2025}
}

@article{peebles2023scalable,
  title={Scalable diffusion models with transformers},
  author={Peebles, William and Xie, Saining},
  journal = {Proc. IEEE/CVF International Conference on Computer Vision},
  pages={4195--4205},
  year={2023}
}

@book{hairer1993ode1,
  author    = {Ernst Hairer and Syvert P. N{\o}rsett and Gerhard Wanner},
  title     = {Solving Ordinary Differential Equations I: Nonstiff Problems},
  edition   = {2},
  publisher = {Springer},
  year      = {1993}
}

@misc{uic_variablestep_notes,
  title        = {Variable Step Methods},
  author       = {Verschelde, Jan},
  year         = {2022},
  note         = {Lecture notes for MCS 471, University of Illinois Chicago}
}

@article{butcher2000ode20thcentury,
  author  = {J. C. Butcher},
  title   = {Numerical methods for ordinary differential equations in the 20th century},
  journal = {Journal of Computational and Applied Mathematics},
  volume  = {125},
  number  = {1--2},
  pages   = {1--29},
  year    = {2000}
}

@book{ascher1998computer,
  title     = {Computer Methods for Ordinary Differential Equations and Differential-Algebraic Equations},
  author    = {Ascher, Uri M. and Petzold, Linda R.},
  publisher = {SIAM},
  year      = {1998}
}

@incollection{bohm1984defect,
  author    = {B{\"o}hmer, Klaus and Hemker, P. W. and
               Stetter, H. J.},
  title     = {The Defect Correction Approach},
  booktitle = {Defect Correction Methods:
               Theory and Applications},
  editor    = {B{\"o}hmer, Klaus and Stetter, H. J.},
  series    = {Computing Supplementum},
  volume    = {5},
  pages     = {1--32},
  year      = {1984},
  publisher = {Springer, Vienna},
  doi       = {10.1007/978-3-7091-7023-6_1}
}

@article{ong2020deferred,
  author  = {Ong, Benjamin W. and
             Spiteri, Raymond J.},
  title   = {Deferred Correction Methods for
             Ordinary Differential Equations},
  journal = {Journal of Scientific Computing},
  volume  = {83},
  number  = {3},
  pages   = {60},
  year    = {2020},
  doi     = {10.1007/s10915-020-01235-8}
}

@article{kong2024hunyuanvideo,
  title={Hunyuanvideo: A systematic framework for large video generative models},
  author={Kong, Weijie and Tian, Qi and Zhang, Zijian and Min, Rox and Dai, Zuozhuo and Zhou, Jin and Xiong, Jiangfeng and Li, Xin and Wu, Bo and Zhang, Jianwei and others},
  journal={arXiv preprint arXiv:2412.03603},
  year={2024}
}

@article{radford2021learning,
  title={Learning transferable visual models from natural language supervision},
  author={Radford, Alec and Kim, Jong Wook and Hallacy, Chris and Ramesh, Aditya and Goh, Gabriel and Agarwal, Sandhini and Sastry, Girish and Askell, Amanda and Mishkin, Pamela and Clark, Jack and others},
  journal={Proc. International Conference on Machine Learning},
  pages={8748--8763},
  year={2021}
}

@article{li2022blip,
  title={Blip: Bootstrapping language-image pre-training for unified vision-language understanding and generation},
  author={Li, Junnan and Li, Dongxu and Xiong, Caiming and Hoi, Steven},
  journal={Proc. International Conference on Machine Learning},
  pages={12888--12900},
  year={2022}
}

@article{liu2024evalcrafter,
  title={Evalcrafter: Benchmarking and evaluating large video generation models},
  author={Liu, Yaofang and Cun, Xiaodong and Liu, Xuebo and Wang, Xintao and Zhang, Yong and Chen, Haoxin and Liu, Yang and Zeng, Tieyong and Chan, Raymond and Shan, Ying},
  journal={Proc. IEEE/CVF Conference on Computer Vision and Pattern Recognition},
  pages={22139--22149},
  year={2024}
}

@article{wu2025qwen,
  title={Qwen-image technical report},
  author={Wu, Chenfei and Li, Jiahao and Zhou, Jingren and Lin, Junyang and Gao, Kaiyuan and Yan, Kun and Yin, Sheng-ming and Bai, Shuai and Xu, Xiao and Chen, Yilei and others},
  journal={arXiv preprint arXiv:2508.02324},
  year={2025}
}

@article{xu2023imagereward,
  title={Imagereward: Learning and evaluating human preferences for text-to-image generation},
  author={Xu, Jiazheng and Liu, Xiao and Wu, Yuchen and Tong, Yuxuan and Li, Qinkai and Ding, Ming and Tang, Jie and Dong, Yuxiao},
  journal={Proc. Advances in Neural Information Processing Systems},
  volume={36},
  pages={15903--15935},
  year={2023}
}

@inproceedings{liu2025taylorseer,
  title={From Reusing to Forecasting: Accelerating Diffusion Models with TaylorSeers},
  author={Liu, Jiacheng and Zou, Chang and Lyu, Yuanhuiyi and Chen, Junjie and Zhang, Linfeng},
  booktitle={Proceedings of the IEEE/CVF International Conference on Computer Vision (ICCV)},
  pages={15853--15863},
  year={2025}
}

@misc{fan2025qwenimagelightning,
  title={Qwen-Image-Lightning},
  author={Fan, Xiangyu and Qiu, Zesong and Wang, Fanzhou and Gao, Peng and Yang, Lei},
  year={2025},
  howpublished={\url{https://github.com/ModelTC/LightX2V-Qwen-Image-Lightning}}
}

\appendix
\section{Theoretical results and additional experiments}
\label{sec:appendix}

\subsection{Proof of Lemma \ref{lem:rho_rectified}}
\label{app:rectified_lemma}

Starting from the reverse-time ODE in \eqref{eq:reverse_ode_eps},
\begin{equation}
\label{eq:reverse_ode_expanded_appendix}
\frac{d\x_t}{dt}
=
\frac{\dot s_t}{s_t}\x_t
+
\left(
\dot\sigma_t-\frac{\dot s_t}{s_t}\sigma_t
\right)\epsilon_\theta(\x_t,t).
\end{equation}

Consider the rescaled state.
\[
\y_t:=\frac{\x_t}{s_t},
\qquad\text{so that}\qquad
\x_t=s_t\y_t.
\]
Differentiating \(\x_t=s_t\y_t\) with respect to \(t\) gives
\[
\frac{d\x_t}{dt}
=
\dot s_t\,\y_t+s_t\frac{d\y_t}{dt}.
\]
Substituting this into \eqref{eq:reverse_ode_expanded_appendix}, we obtain
\[
\dot s_t\,\y_t+s_t\frac{d\y_t}{dt}
=
\frac{\dot s_t}{s_t}(s_t\y_t)
+
\left(
\dot\sigma_t-\frac{\dot s_t}{s_t}\sigma_t
\right)\epsilon_\theta(s_t\y_t,t).
\]
Canceling out common terms and rearranging gives,
\begin{equation}
\label{eq:y_t_dynamics_appendix}
\frac{d\y_t}{dt}
=
\frac{1}{s_t}
\left(
\dot\sigma_t-\frac{\dot s_t}{s_t}\sigma_t
\right)\epsilon_\theta(s_t\y_t,t).
\end{equation}

The noise-to-signal ratio is defined as
\[
\rho_t:=\frac{\sigma_t}{s_t}.
\]
Differentiating \(\rho_t\) with respect to \(t\),
\begin{equation}
\label{rho_derivative}
\dot\rho_t
=
\frac{\dot\sigma_t s_t-\sigma_t\dot s_t}{s_t^2}
=
\frac{1}{s_t}
\left(
\dot\sigma_t-\frac{\dot s_t}{s_t}\sigma_t
\right).
\end{equation}
Using this identity in \eqref{eq:y_t_dynamics_appendix}, we get
\begin{equation}
\label{eq:y_t_dt_appendix}
\frac{d\y_t}{dt}
=
\dot\rho_t\,\epsilon_\theta(s_t\y_t,t).
\end{equation}

By assumption, \(\rho:[0,T]\to\mathbb R_{>0}\) is strictly increasing, so the inverse map \(t=t(\rho)\) exists. The \(\rho\)-reparameterized trajectory is defined as
\[
\hat{\y}_\rho:=\y_{t(\rho)}.
\]
Applying the chain rule,
\[
\frac{d\hat{\y}_\rho}{d\rho}
=
\frac{d\y_t}{dt}\Big|_{t=t(\rho)}\cdot \frac{dt}{d\rho}.
\]
Since \(\frac{dt}{d\rho}=\frac{1}{\dot\rho_{t(\rho)}}\), we get
\[
\frac{d\hat{\y}_\rho}{d\rho}
=
\frac{1}{\dot\rho_{t(\rho)}}\,
\frac{d\y_t}{dt}\Big|_{t=t(\rho)}.
\]
Using \eqref{eq:y_t_dt_appendix} evaluated at \(t=t(\rho)\),
\[
\frac{d\y_t}{dt}\Big|_{t=t(\rho)}
=
\dot\rho_{t(\rho)}\,
\epsilon_\theta\bigl(s_{t(\rho)}\y_{t(\rho)},t(\rho)\bigr).
\]
Substituting the above,
\[
\frac{d\hat{\y}_\rho}{d\rho}
=
\frac{1}{\dot\rho_{t(\rho)}}
\dot\rho_{t(\rho)}
\epsilon_\theta\bigl(s_{t(\rho)}\y_{t(\rho)},t(\rho)\bigr)
=
\epsilon_\theta\bigl(s_{t(\rho)}\y_{t(\rho)},t(\rho)\bigr)
=
\epsilon_\theta\bigl(s_{t(\rho)}\hat{\y}_\rho,t(\rho)\bigr),
\]
which proves the lemma.

\subsection{Relation between noise/data/velocity parameterizations \eqref{eq:rectified_param_relations} in rectified coordinates}
\label{sec:A2}
From the standard data-prediction parameterization, we have
\[
\epsilon_\theta(\x_t,t)=\frac{\x_t-s_t\hat{\x}_\theta(\x_t,t)}{\sigma_t}.
\]
Now substituting \(\x_t=s_{t(\rho)}\hat{\y}_\rho\) and using
$\rho=\frac{\sigma_{t(\rho)}}{s_{t(\rho)}}$,
gives
\[
\epsilon_\theta(s_{t(\rho)}\hat{\y}_\rho,t(\rho))
=
\frac{s_{t(\rho)}\hat{\y}_\rho-s_{t(\rho)}\hat{\x}_\theta(s_{t(\rho)}\hat{\y}_\rho,t(\rho))}{\sigma_{t(\rho)}}
=
\frac{\hat{\y}_\rho-\hat{\x}_\theta(s_{t(\rho)}\hat{\y}_\rho,t(\rho))}{\rho}.
\]

Similarly, starting from
\[
\boldv_\theta(\x_t,t)=\dot s_t\,\hat{\x}_\theta(\x_t,t)+\dot\sigma_t\,\epsilon_\theta(\x_t,t),
\]
we substitute \(\x_t=s_{t(\rho)}\hat{\y}_\rho\) and $\hat{\x}_\theta(s_{t(\rho)}\hat{\y}_\rho,t(\rho)) = \hat{\y}_\rho-\rho\,\epsilon_\theta(s_{t(\rho)}\hat{\y}_\rho,t(\rho))$ to obtain
\begin{align*}
\boldv_\theta(s_{t(\rho)}\hat{\y}_\rho,t(\rho))
&= \dot s_{t(\rho)}\,\hat{\x}_\theta(s_{t(\rho)}\hat{\y}_\rho,t(\rho))+\dot\sigma_{t(\rho)}\,\epsilon_\theta(s_{t(\rho)}\hat{\y}_\rho,t(\rho)) \\
&= \dot s_{t(\rho)}\bigl(\hat{\y}_\rho-\rho\,\epsilon_\theta(s_{t(\rho)}\hat{\y}_\rho,t(\rho))\bigr)+\dot\sigma_{t(\rho)}\,\epsilon_\theta(s_{t(\rho)}\hat{\y}_\rho,t(\rho)) \\
&= \dot s_{t(\rho)}\,\hat{\y}_\rho + (\dot\sigma_{t(\rho)}-\dot s_{t(\rho)}\rho)\epsilon_\theta(s_{t(\rho)}\hat{\y}_\rho,t(\rho)) \\
&= \dot s_{t(\rho)}\,\hat{\y}_\rho + s_{t(\rho)}\dot\rho_{t(\rho)}\,\epsilon_\theta(s_{t(\rho)}\hat{\y}_\rho,t(\rho)),
\end{align*}
The last equality is obtained from the form of $\dot{\rho_{t(\rho)}}$ in \eqref{rho_derivative}. Rearranging terms, we get, 
\[
\epsilon_\theta(s_{t(\rho)}\hat{\y}_\rho,t(\rho))
=
\frac{\boldv_\theta(s_{t(\rho)}\hat{\y}_\rho,t(\rho))-\dot s_{t(\rho)}\hat{\y}_\rho}
{s_{t(\rho)}\dot\rho_{t(\rho)}}.
\]

\subsection{Derivation of CAB-2 and CAB-3 updates}
\label{sec:A3}
The CAB-2 prediction step corresponds to the standard variable-step AB2 formula, whose derivation can be found, for example, in \cite{uic_variablestep_notes}. Likewise, the CAB-3 prediction step follows the standard variable-step AB3 construction; see \cite{hairer1993ode1}. For completeness, we present below the derivations of the CAB-2 predictor, the CAB-3 predictor, and the proposed correction term. 

\paragraph{Prediction step for CAB-2.}

To derive the CAB-2 predictor, we approximate the velocity in \eqref{eq:rectified_rho_ode} over the interval \([\rho_{t_i},\rho_{t_{i+1}}]\) by the degree-1 Lagrange interpolant through the two most recent points \((\rho_{t_i},\epsilon_i)\) and \((\rho_{t_{i-1}},\epsilon_{i-1})\), namely
\[
\mathcal{L}_1(\rho)
=
\frac{\rho-\rho_{t_{i-1}}}{\rho_{t_i}-\rho_{t_{i-1}}}\,\epsilon_i
+
\frac{\rho-\rho_{t_i}}{\rho_{t_{i-1}}-\rho_{t_i}}\,\epsilon_{i-1}.
\]
Integrating the rectified ODE \eqref{eq:rectified_rho_ode} with this approximation gives
\[
\y_{t_{i+1}}
\approx
\y_{t_i}
+
\int_{\rho_{t_i}}^{\rho_{t_{i+1}}}\mathcal{L}_1(\rho)\,d\rho.
\]
Since
\[
\rho_{t_i}-\rho_{t_{i-1}}=h_{i-1},
\qquad
\rho_{t_{i+1}}-\rho_{t_i}=h_i,
\qquad
r_i=\frac{h_i}{h_{i-1}},
\]
we obtain
\begin{align*}
\int_{\rho_{t_i}}^{\rho_{t_{i+1}}}\mathcal{L}_1(\rho)\,d\rho
&=
\left(
\int_{\rho_{t_i}}^{\rho_{t_{i+1}}}
\frac{\rho-\rho_{t_{i-1}}}{h_{i-1}}\,d\rho
\right)\epsilon_i
-
\left(
\int_{\rho_{t_i}}^{\rho_{t_{i+1}}}
\frac{\rho-\rho_{t_i}}{h_{i-1}}\,d\rho
\right)\epsilon_{i-1} \\
&=
\left(h_i+\frac{h_i^2}{2h_{i-1}}\right)\epsilon_i
-
\frac{h_i^2}{2h_{i-1}}\epsilon_{i-1} \\
&=
h_i\left[\left(1+\frac{r_i}{2}\right)\epsilon_i-\frac{r_i}{2}\epsilon_{i-1}\right].
\end{align*}
This is exactly the standard non-uniform Adams--Bashforth prediction as derived in \cite{uic_variablestep_notes}.

\paragraph{Prediction step for CAB-3.}
Similarly, CAB-3 is obtained by approximating the velocity in \eqref{eq:rectified_rho_ode} over the interval \([\rho_{t_i},\rho_{t_{i+1}}]\) by the degree-2 Lagrange interpolant through the three most recent points \((\rho_{t_i},\epsilon_i)\), \((\rho_{t_{i-1}},\epsilon_{i-1})\), and \((\rho_{t_{i-2}},\epsilon_{i-2})\), namely
\[
\mathcal{L}_2(\rho)
=
\ell_i(\rho)\epsilon_i+\ell_{i-1}(\rho)\epsilon_{i-1}+\ell_{i-2}(\rho)\epsilon_{i-2},
\]
where
\[
\ell_i(\rho)
=
\frac{(\rho-\rho_{t_{i-1}})(\rho-\rho_{t_{i-2}})}
{(\rho_{t_i}-\rho_{t_{i-1}})(\rho_{t_i}-\rho_{t_{i-2}})},
\]
\[
\ell_{i-1}(\rho)
=
\frac{(\rho-\rho_{t_i})(\rho-\rho_{t_{i-2}})}
{(\rho_{t_{i-1}}-\rho_{t_i})(\rho_{t_{i-1}}-\rho_{t_{i-2}})},
\qquad
\ell_{i-2}(\rho)
=
\frac{(\rho-\rho_{t_i})(\rho-\rho_{t_{i-1}})}
{(\rho_{t_{i-2}}-\rho_{t_i})(\rho_{t_{i-2}}-\rho_{t_{i-1}})}.
\]
Integrating the rectified ODE \eqref{eq:rectified_rho_ode} with this approximation gives
\[
\y_{t_{i+1}}
\approx
\y_{t_i}
+
\int_{\rho_{t_i}}^{\rho_{t_{i+1}}}\mathcal{L}_2(\rho)\,d\rho.
\]
Using
\[
h_i=\rho_{t_{i+1}}-\rho_{t_i},
\qquad
h_{i-1}=\rho_{t_i}-\rho_{t_{i-1}},
\qquad
h_{i-2}=\rho_{t_{i-1}}-\rho_{t_{i-2}},
\]
The integral can be written as
\[
\int_{\rho_{t_i}}^{\rho_{t_{i+1}}}\mathcal{L}_2(\rho)\,d\rho
=
h_i\left(\beta_0\epsilon_i+\beta_1\epsilon_{i-1}+\beta_2\epsilon_{i-2}\right),
\]
where
\[
\beta_0=
\frac{
\frac{h_i^2}{3}+\frac{h_i}{2}(2h_{i-1}+h_{i-2})+h_{i-1}(h_{i-1}+h_{i-2})
}{
h_{i-1}(h_{i-1}+h_{i-2})
},
\]
\[
\beta_1=
-\frac{h_i(2h_i+3h_{i-1}+3h_{i-2})}{6h_{i-1}h_{i-2}},
\qquad
\beta_2=
\frac{h_i(2h_i+3h_{i-1})}{6h_{i-2}(h_{i-1}+h_{i-2})}.
\]
Equivalently, in terms of the step ratios
\[
r_i:=\frac{h_i}{h_{i-1}},
\qquad
r_{i-1}:=\frac{h_{i-1}}{h_{i-2}},
\]
These coefficients become
\[
\beta_0=
1+\frac{r_i(2r_{i-1}+1)}{2(r_{i-1}+1)}
+\frac{r_{i-1}r_i^2}{3(r_{i-1}+1)},
\]
\[
\beta_1=
-\frac{r_i}{6}\left(2r_{i-1}r_i+3r_{i-1}+3\right),
\qquad
\beta_2=
\frac{r_{i-1}^2r_i(2r_i+3)}{6(r_{i-1}+1)}.
\]
This yields the explicit variable-step AB3 predictor. The derivation follows the standard variable-step Adams construction on non-uniform grids, obtained by integrating the quadratic Lagrange interpolant of past drift evaluations; we include the resulting closed-form coefficients here for completeness and refer to \cite{hairer1993ode1} for the general derivation.

\paragraph{Correction step.}
To derive the correction term, we approximate the velocity at the current point \(\rho_{t_i}\) by the degree-1 Lagrange extrapolant built from the two previous evaluations \((\rho_{t_{i-1}},\epsilon_{i-1})\) and \((\rho_{t_{i-2}},\epsilon_{i-2})\). This extrapolant is
\[
\mathcal{L}^{\mathrm{ext}}_1(\rho)
=
\frac{\rho-\rho_{t_{i-2}}}{\rho_{t_{i-1}}-\rho_{t_{i-2}}}\,\epsilon_{i-1}
+
\frac{\rho-\rho_{t_{i-1}}}{\rho_{t_{i-2}}-\rho_{t_{i-1}}}\,\epsilon_{i-2}.
\]
Evaluating at \(\rho=\rho_{t_i}\) gives
\[
\epsilon_i^{\mathrm{ext}}
=
\mathcal{L}^{\mathrm{ext}}_1(\rho_{t_i})
=
\frac{\rho_{t_i}-\rho_{t_{i-2}}}{\rho_{t_{i-1}}-\rho_{t_{i-2}}}\,\epsilon_{i-1}
-
\frac{\rho_{t_i}-\rho_{t_{i-1}}}{\rho_{t_{i-1}}-\rho_{t_{i-2}}}\,\epsilon_{i-2}.
\]
Using
\[
\rho_{t_i}-\rho_{t_{i-1}}=h_{i-1},
\qquad
\rho_{t_{i-1}}-\rho_{t_{i-2}}=h_{i-2},
\qquad
\rho_{t_i}-\rho_{t_{i-2}}=h_{i-1}+h_{i-2},
\]
we obtain
\[
\epsilon_i^{\mathrm{ext}}
=
\left(1+\frac{h_{i-1}}{h_{i-2}}\right)\epsilon_{i-1}
-
\frac{h_{i-1}}{h_{i-2}}\,\epsilon_{i-2}
=
(1+r_{i-1})\epsilon_{i-1}-r_{i-1}\epsilon_{i-2},
\qquad
r_{i-1}=\frac{h_{i-1}}{h_{i-2}}.
\]
The correction is then defined by the extrapolation error,
\[
\epsilon_i-\epsilon_i^{\mathrm{ext}},
\]
scaled by the current step size \(h_i\). 

\subsection{Proof of Theorem~\ref{thm:cab_accuracy}}
\label{thmproof:cab_accuracy}

We use a unified argument for CAB-2 and CAB-3. Let
\[
\z_i:=\hat{\y}(\rho_{t_i}),
\qquad
\bu_i:=\hat{\y}'(\rho_{t_i}),
\qquad
e_i:=\y_{t_i}-\z_i.
\]
The update in Algorithm~\ref{alg:cabp_main} of the form
\begin{equation}
\label{eq:generic_scheme_proof}
\y_{t_{i+1}}
=
\y_{t_i}
+
h_i\sum_{m=0}^{2} a_{m,i}\,\epsilon_{i-m},
\end{equation}
where CAB-2 and CAB-3 have different coefficients \(a_{m,i}\). The corresponding exact-solution error is
\begin{equation}
\label{eq:generic_tau_definition}
\tau_{i+1}
:=
\z_{i+1}-\z_i-h_i\sum_{m=0}^{q} a_{m,i}\bu_{i-m}.
\end{equation}

For CAB-2, the coefficients are
\begin{equation}
\label{eq:cab2_coeffs_generic}
a_{0,i}=1+\frac{r_i}{2}+\gamma_i,\qquad
a_{1,i}=-\Bigl(\frac{r_i}{2}+(1+r_{i-1})\gamma_i\Bigr),\qquad
a_{2,i}=r_{i-1}\gamma_i,
\end{equation}
where
\[
r_i=\frac{h_i}{h_{i-1}},
\qquad
r_{i-1}=\frac{h_{i-1}}{h_{i-2}}.
\]

For CAB-3, letting \(\beta_{0,i},\beta_{1,i},\beta_{2,i}\) denote the variable-step AB3 coefficients in the companion predictor box of Algorithm~\ref{alg:cabp_main}, the corrected coefficients are
\begin{equation}
\label{eq:cab3_coeffs_generic}
a_{0,i}=\beta_{0,i}+\gamma_i,\qquad
a_{1,i}=\beta_{1,i}-(1+r_{i-1})\gamma_i,\qquad
a_{2,i}=\beta_{2,i}+r_{i-1}\gamma_i.
\end{equation}

\paragraph{Step 1: local truncation error.}
Expand the exact increment about \(\rho_{t_i}\) using Taylor series:
\begin{equation}
\label{eq:generic_Y_increment}
\z_{i+1}-\z_i
=
h_i \z_i'
+
\frac{h_i^2}{2} \z_i''
+
\frac{h_i^3}{6} \z_i'''
+
\frac{h_i^4}{24} \z_i''''
+
O(h_i^5).
\end{equation}
where \(p=2\) for CAB-2 and \(p=3\) for CAB-3. Next expand the past velocities \(\bu_{i-1}\) and \(\bu_{i-2}\) about \(\rho_{t_i}\). Since
\[
\rho_{t_{i-1}}-\rho_{t_i}=-\frac{h_i}{r_i},
\qquad
\rho_{t_{i-2}}-\rho_{t_i}=-\frac{h_i(1+r_{i-1})}{r_ir_{i-1}},
\]
we obtain
\begin{equation}
\label{eq:generic_Vim1}
\bu_{i-1}
=
\bu_i-\frac{h_i}{r_i}\z''_i+\frac{h_i^2}{2r_i^2}\z'''_i-\frac{h_i^3}{6r_i^3}\z''''_i+O(h_i^4),
\end{equation}
\begin{equation}
\label{eq:generic_Vim2}
\bu_{i-2}
=
\bu_i-\frac{h_i(1+r_{i-1})}{r_ir_{i-1}}\z''_i
+\frac{h_i^2(1+r_{i-1})^2}{2r_i^2r_{i-1}^2}\z'''_i
-\frac{h_i^3(1+r_{i-1})^3}{6r_i^3r_{i-1}^3}\z''''_i
+O(h_i^4).
\end{equation}

Substituting \eqref{eq:generic_Vim1}--\eqref{eq:generic_Vim2} into \eqref{eq:generic_tau_definition} yields
\begin{align}
&\tau_{i+1}
=
h_i\Bigl[1-\sum_{m=0}^{2}a_{m,i}\Bigr]\z'_i 
+h_i^2\left[\frac12+\frac{a_{1,i}}{r_i}+\frac{a_{2,i}(1+r_{i-1})}{r_ir_{i-1}}\right]\z''_i \notag\\
&+h_i^3\left[\frac16-\frac{a_{1,i}}{2r_i^2}-\frac{a_{2,i}(1+r_{i-1})^2}{2r_i^2r_{i-1}^2}\right]\z'''_i 
+h_i^4\left[\frac1{24}+\frac{a_{1,i}}{6r_i^3}+\frac{a_{2,i}(1+r_{i-1})^3}{6r_i^3r_{i-1}^3}\right]\z''''_i
+O(h_i^5)
\label{eq:generic_tau_expansion}
\end{align}

\medskip
\noindent\textbf{CAB-2.}
Using \eqref{eq:cab2_coeffs_generic}, we can verify that
\[
a_{0,i}+a_{1,i}+a_{2,i}=1,
\]
and
\[
\frac{a_{1,i}}{r_i}+\frac{a_{2,i}(1+r_{i-1})}{r_ir_{i-1}}=-\frac12.
\]
Hence the \(O(h_i)\) and \(O(h_i^2)\) terms in \eqref{eq:generic_tau_expansion} cancel, leaving
\[
\tau_{i+1}=O(h_i^3).
\]
Thus, CAB-2 has local truncation order \(3\).

\medskip
\noindent\textbf{CAB-3.}
For the variable-step AB3 predictor coefficients \(\beta_{0,i},\beta_{1,i},\beta_{2,i}\), the third-order consistency conditions hold:
\[
\beta_{0,i}+\beta_{1,i}+\beta_{2,i}=1,
\]
\[
\frac{\beta_{1,i}}{r_i}+\frac{\beta_{2,i}(1+r_{i-1})}{r_ir_{i-1}}=-\frac12,
\]
\[
\frac{\beta_{1,i}}{2r_i^2}+\frac{\beta_{2,i}(1+r_{i-1})^2}{2r_i^2r_{i-1}^2}=-\frac16.
\]
The correction modifies the AB3 coefficients according to
\[
a_{0,i}=\beta_{0,i}+\gamma_i,\qquad
a_{1,i}=\beta_{1,i}-(1+r_{i-1})\gamma_i,\qquad
a_{2,i}=\beta_{2,i}+r_{i-1}\gamma_i.
\]
\(O(h_i)\) and \(O(h_i^2)\) terms still cancel since the additional terms satisfy
\[
\gamma_i - (1+r_{i-1})\gamma_i + r_{i-1}\gamma_i = 0,
\]
\[
\frac{-(1+r_{i-1})\gamma_i}{r_i}+\frac{r_{i-1}\gamma_i(1+r_{i-1})}{r_ir_{i-1}}=0
\]

Substituting  $a_{1,i}, a_{2,i}$ into the coefficient of \(h_i^3\z_i'''\) in \eqref{eq:generic_tau_expansion}, we obtain
\begin{equation*}
\frac{1}{6}
-\frac{a_{1,i}}{2r_i^2}
-\frac{a_{2,i}(1+r_{i-1})^2}{2r_i^2r_{i-1}^2}
\notag =
-\gamma_i\,
\frac{1+r_{i-1}}{2r_i^2r_{i-1}}.
\label{eq:cab3_third_order_coeff}
\end{equation*}
Therefore,
\begin{equation}
\label{eq:cab3_tau_theta}
\tau_{i+1}
=
\gamma_i\,h_i^3\,\kappa_i\,\z_i'''
+
O(h_i^4),
\qquad
\kappa_i:=
-\frac{1+r_{i-1}}{2r_i^2r_{i-1}}.
\end{equation}
Since the step ratios are uniformly bounded away from \(0\) and \(\infty\), \(\kappa_i\) is uniformly bounded. Consequently, if \(\gamma_i=O(1)\), then \(\tau_{i+1}=O(h_i^3)\), whereas if \(\gamma_i=O(h_i)\), then \(\tau_{i+1}=O(h_i^4)\). Hence CAB-3 has local truncation order \(4\) under the condition \(\gamma_i=O(h_i)\).

\paragraph{Step 2: global error.}
The exact solution satisfies
\[
\z_{i+1}
=
\z_i+h_i\sum_{m=0}^{2} a_{m,i}\bu_{i-m}+\tau_{i+1},
\]
while the numerical scheme satisfies
\[
\y_{t_{i+1}}
=
\y_{t_i}+h_i\sum_{m=0}^{2} a_{m,i}\epsilon_{i-m}.
\]
Subtracting gives
\begin{equation}
\label{eq:generic_error_recurrence}
e_{i+1}
=
e_i+h_i\sum_{m=0}^{2} a_{m,i}\Delta_{i-m}-\tau_{i+1},
\end{equation}
where
\[
\Delta_j:=\epsilon_\theta(s_{t_j}\y_{t_j},t_j)-\epsilon_\theta(s_{t_j}\z_j,t_j).
\]
By the Lipschitz assumption,
\[
\|\Delta_j\|\le L\|e_j\|.
\]
Since the step ratios are uniformly bounded and the corrector weights are bounded, the coefficients \(a_{m, i}\) are uniformly bounded. Thus there exists \(B>0\) such that
\[
\sum_{m=0}^{2}|a_{m,i}|\le B.
\]
Taking norms in \eqref{eq:generic_error_recurrence}, we obtain
\[
\|e_{i+1}\|
\le
\|e_i\|+BLh_i\max\{\|e_i\|,\|e_{i-1}\|,\|e_{i-2}\|\}+\|\tau_{i+1}\|.
\]
Define
\[
E_i:=\max_{0\le j\le i}\|e_j\|.
\]
Then
\[
E_{i+1}\le (1+BLh_i)E_i+\|\tau_{i+1}\|.
\]
Iterating this recurrence and using \(1+x\le e^x\) to form  
\[
\prod_{j=k+1}^{n-1} \left(1 + BLh_j\right)
\le
\exp\!\bigl(BL(\rho_{t_0}-\rho_{t_N})\bigr),
\]
yields a discrete Gr\"onwall bound of the form
\begin{equation}
\label{eq:En_after_exp}
E_n
\le
C_G\left(E_2+\sum_{k=2}^{n-1}\|\tau_{k+1}\|\right).
\end{equation}

\medskip
\noindent\textbf{CAB-2.}
For CAB-2, the local truncation bound is
\[
\|\tau_{k+1}\|\le C_\tau h_k^3.
\]

Substituting this into \eqref{eq:En_after_exp} gives
\[
E_n
\le
C_G E_2 + C_G C_\tau \sum_{k=2}^{n-1} h_k^3.
\]
Since \(h_k\le h_{\max}\),
\[
\sum_{k=2}^{n-1} h_k^3
\le
h_{\max}^2 \sum_{k=2}^{n-1} h_k
\le
h_{\max}^2(\rho_{t_N}-\rho_{t_0}).
\]
By assumption, the starting iterates satisfy the target global-order bound, so we get
\[
E_n
\le
C h_{\max}^2,
\]
for some constant \(C>0\) independent of step sizes $\{h_i\}$. 

\medskip
\noindent\textbf{CAB-3.}
For CAB-3, under the condition \(\gamma_i=O(h_i)\), the local truncation bound is
\[
\|\tau_{k+1}\|\le C_\tau h_k^4.
\]
Similar to CAB-2, using \(h_k\le h_{\max}\),
\[
\sum_{k=2}^{n-1} h_k^4
\le
h_{\max}^3 \sum_{k=2}^{n-1} h_k
\le
h_{\max}^3(\rho_{t_N}-\rho_{t_0}).
\]
and since the starting iterates are assumed accurate to the target global order, we get 
\[
E_n
\le
C h_{\max}^3,
\]
for some constant \(C>0\) independent of step sizes $\{h_i\}$. 

\subsection{Empirical demonstration of Theorem~\ref{thm:cab_accuracy}}
\begin{figure}[t]
    \centering  
    \begin{subfigure}[t]{0.415\textwidth}
        \centering
        \includegraphics[width=\linewidth]{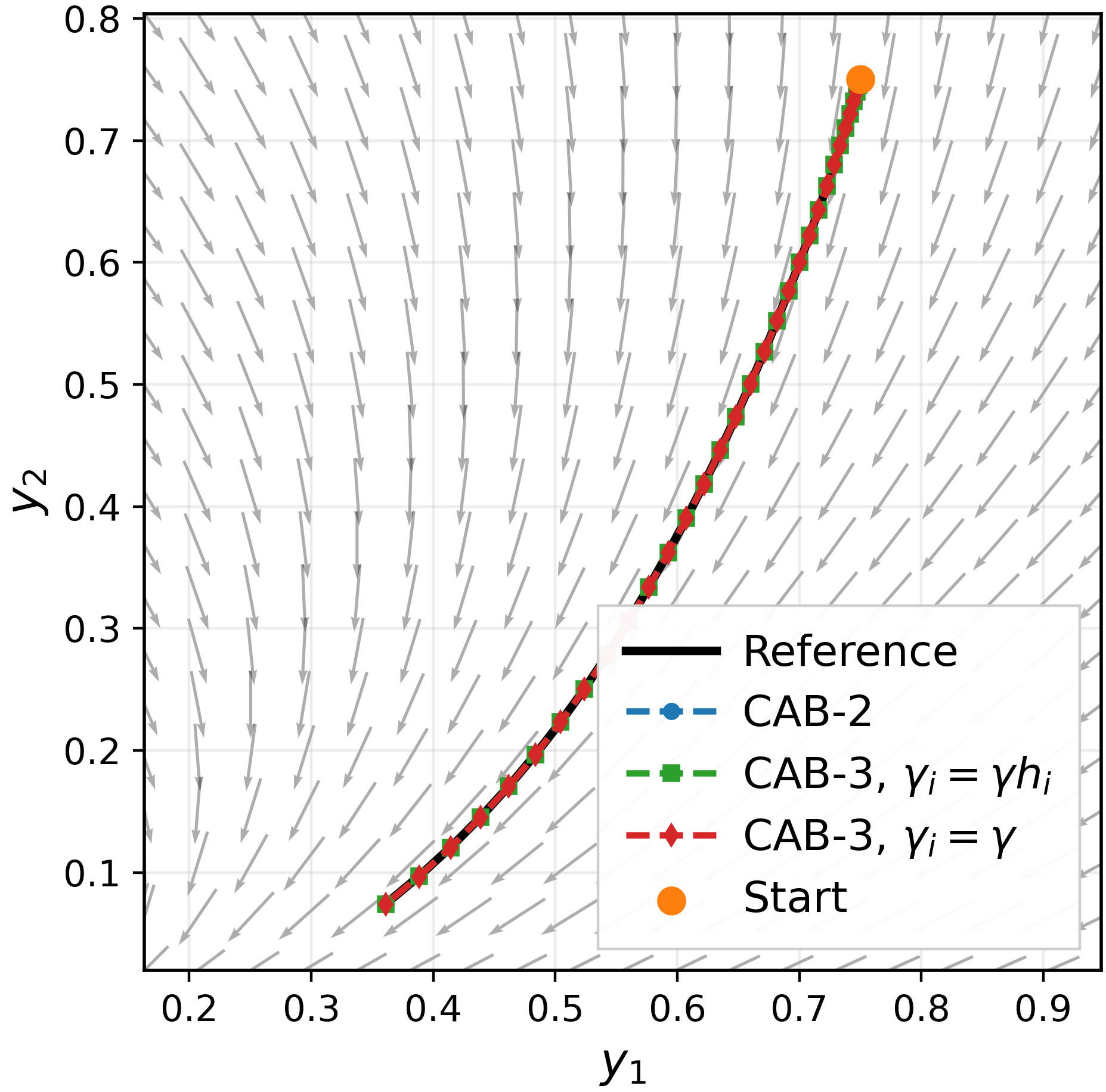}
        \caption{Trajectory for velocity field $v_1$.}
    \end{subfigure}
    \hfill
    \begin{subfigure}[t]{0.535\textwidth}
        \centering
        \includegraphics[width=\linewidth]{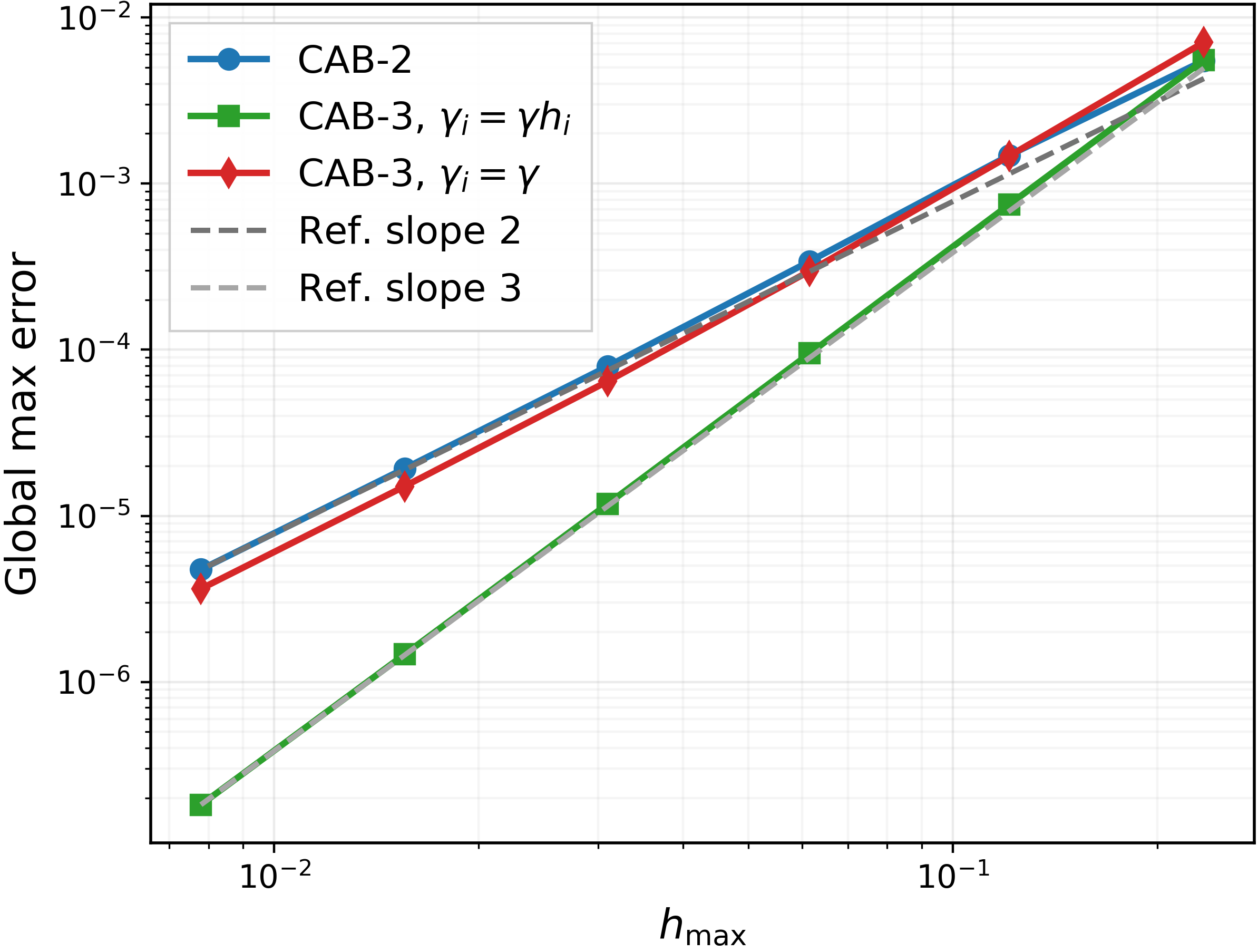}
        \caption{Global max error versus step size.}
    \end{subfigure}

    \vspace{0.4em}
    
    \begin{subfigure}[t]{0.42\textwidth}
        \centering
        \includegraphics[width=\linewidth]{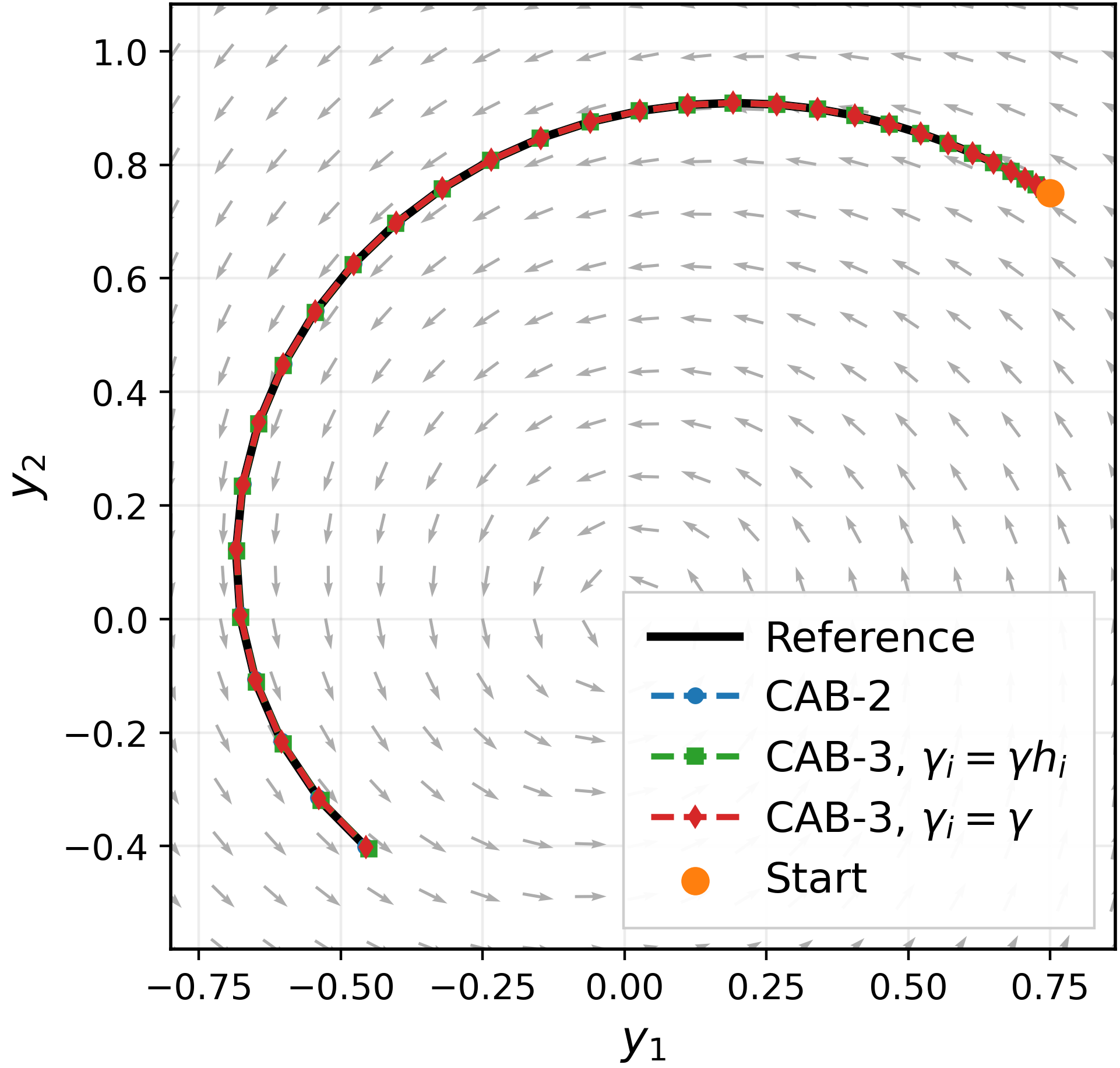}
        \caption{Trajectory for velocity field $v_2$.}
    \end{subfigure}
    \hfill
    \begin{subfigure}[t]{0.53\textwidth}
        \centering
        \includegraphics[width=\linewidth]{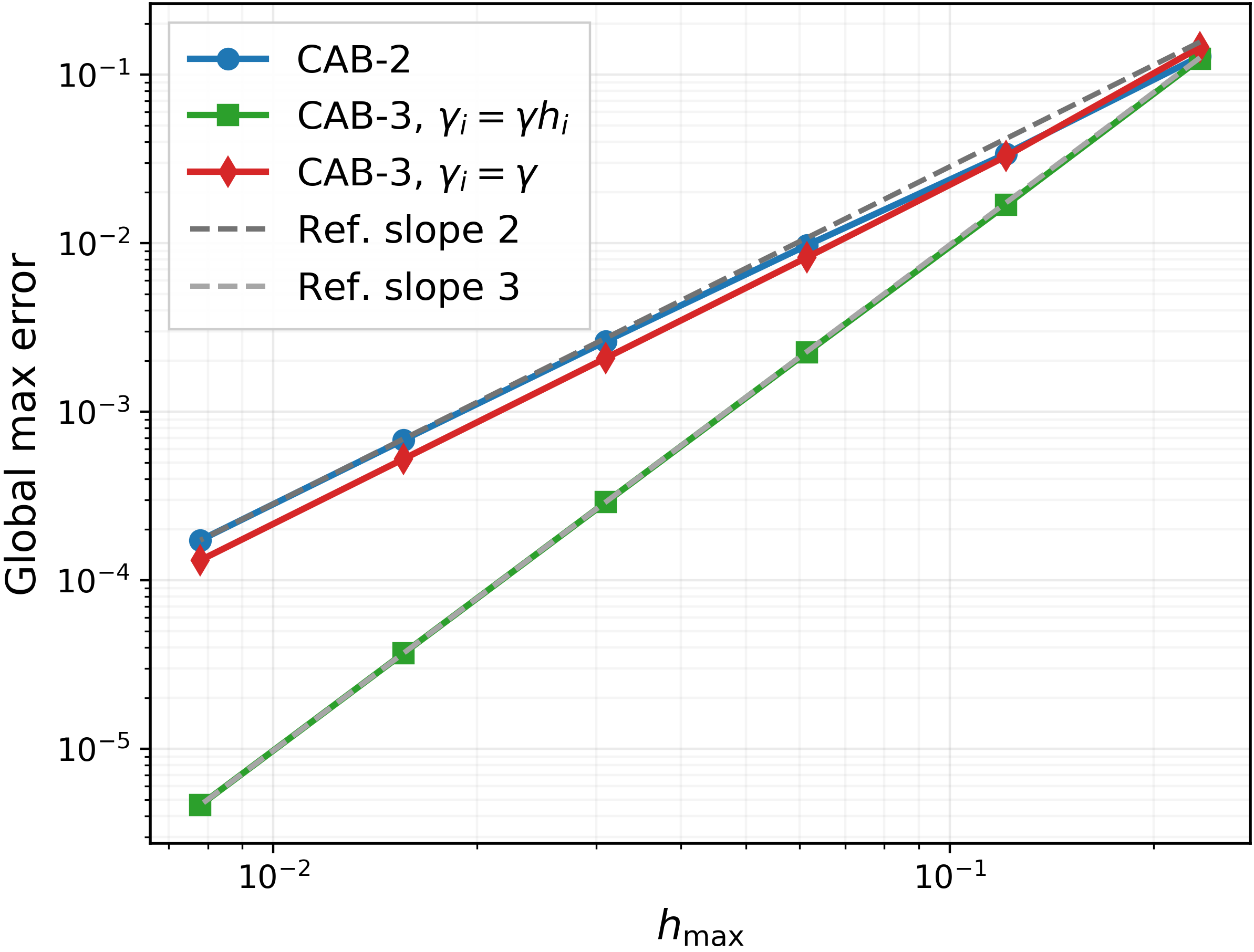}
        \caption{Global max error versus step size.}
    \end{subfigure}
\caption{Empirical verification of the accuracy results in Theorem~\ref{thm:cab_accuracy} on two representative nonlinear velocity fields. (a) and (c) show sample trajectories and normalized velocity fields, while (b) and (d) plot the global maximum error against the maximum step size \(h_{\max}\) on non-uniform \(\rho\)-grids. In both examples, the observed slopes match the theoretical predictions; CAB-2 exhibits second-order global accuracy, whereas CAB-3 exhibits third-order global accuracy .}
\label{fig:traj_rectified}
\end{figure}

We next verify the theoretical convergence results in Theorem~\ref{thm:cab_accuracy} on two representative nonlinear test problems. Specifically, we apply CAB-2 and CAB-3 to the initial-value problem
\begin{equation}
\label{eq:test_ode_accuracy}
\frac{dy}{d\rho}=v(y,\rho),
\qquad
y(\rho_0)=y_0,
\end{equation}
using the two velocity fields shown in Figure.~\ref{fig:traj_rectified}(a) and Figure.~\ref{fig:traj_rectified}(c). The first field is
\begin{equation}
\label{eq:v1_field_accuracy}
v_1(y,\rho)=
\begin{bmatrix}
-1.5\,y_1+0.9\,y_2+0.2\,y_1y_2+0.15\sin(3\rho)\\[1mm]
-1.0\,y_2-0.7\,y_1+0.1\,y_1^2-0.08\,y_2^2+0.1\cos(2\rho)
\end{bmatrix},
\end{equation}
and the second is
\begin{equation}
\label{eq:v2_field_accuracy}
v_2(y,\rho)=
\begin{bmatrix}
-(0.3+0.4(y_1^2+y_2^2))\,y_1-(3.0+0.2\sin\rho)\,y_2\\[1mm]
(3.0+0.2\sin\rho)\,y_1-(0.3+0.4(y_1^2+y_2^2))\,y_2
\end{bmatrix}.
\end{equation}

As the reference solution, we use a high-accuracy adaptive DOP853 integrator with stringent error tolerances using SciPy.  Figures~\ref{fig:traj_rectified}(b) and~\ref{fig:traj_rectified}(d) plot the maximum pointwise trajectory error over the time steps as a function of the maximum step size \(h_{\max}\). For CAB-2, we use \(\gamma=0.75\), while for CAB-3 we consider two choices: a constant parameter \(\gamma=0.25\) and a step-dependent parameter \(\gamma=0.75\,h_i\). In both settings, the empirical convergence behavior is consistent with the theory: the global error scales as \(O(h_{\max}^2)\) for CAB-2 and for CAB-3 with constant \(\gamma\), whereas it improves to \(O(h_{\max}^3)\) for CAB-3 when \(\gamma=\mathcal{O}(h_i)\).

\subsection{Effect of rectification and correction.}
\label{effect:rectcorr}

CAB's gains come from both rectification and correction. Table~5 shows that the noise-to-signal coordinate yields a smoother learned field, supporting the role of rectification in making multistep extrapolation easier. The AB baselines in Tables~1--4 isolate the correction effect: CAB improves over the corresponding AB solver at the same NFE and under the same sampling setup, showing that the gains are not merely due to Adams--Bashforth history. The $\gamma$ ablation further shows that correction strength controls the FID--NIQE trade-off, with moderate values giving robust perceptual behavior. Thus, rectification improves the integration geometry, while CAB's extrapolation-defect correction provides an additional zero-extra-NFE improvement.

\label{sec:correction_advantage}

\begin{figure*}[t]
    \centering
    \setlength{\tabcolsep}{3pt}
    
    \begin{subfigure}[t]{0.48\textwidth}
        \centering
        \includegraphics[width=\linewidth]{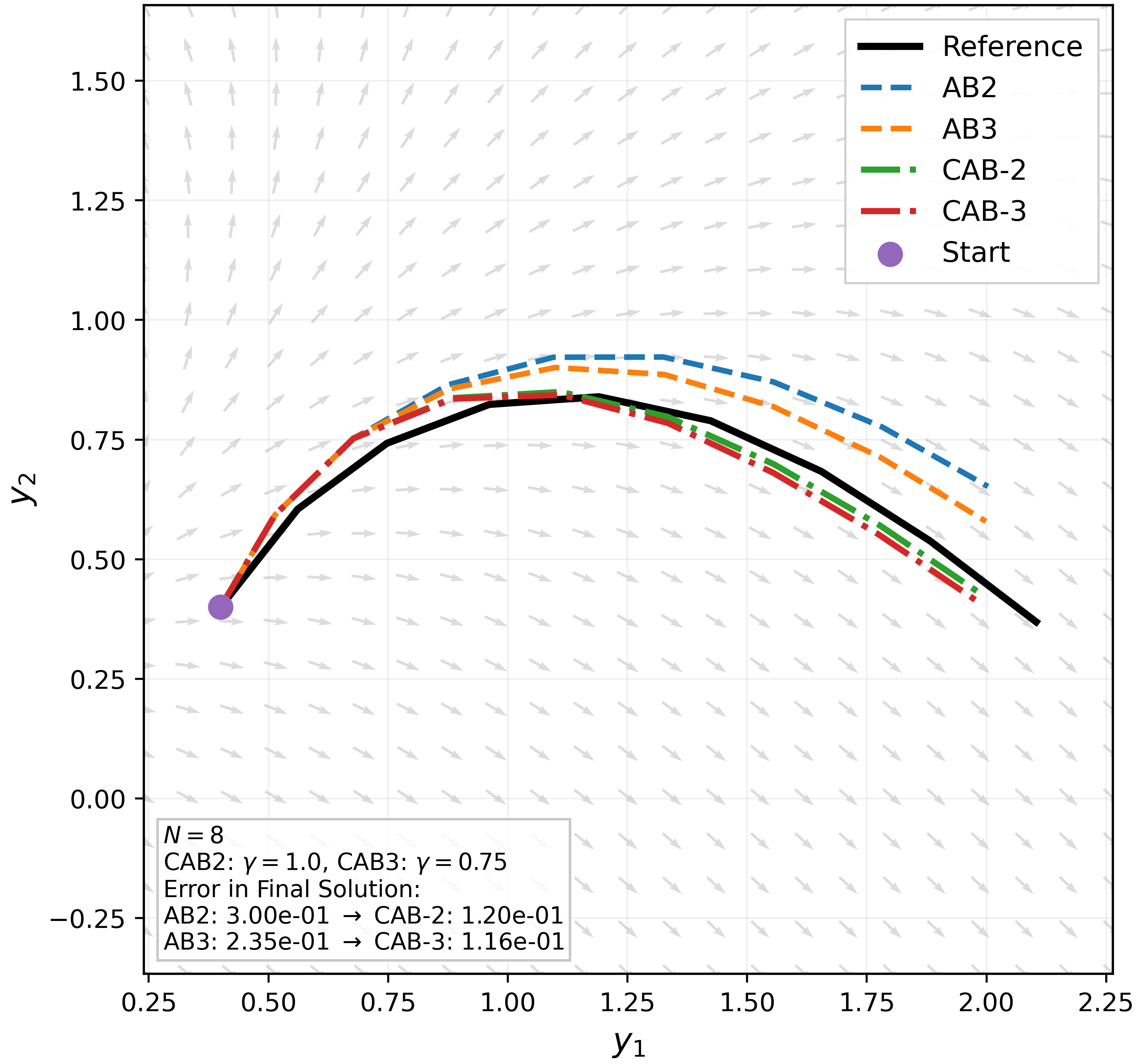}
        \caption{Trajectory comparison for Example 1.}
        \label{fig:cab_vs_ab_traj1}
    \end{subfigure}
    \hfill
    \begin{subfigure}[t]{0.48\textwidth}
        \centering
        \includegraphics[width=0.97\linewidth]{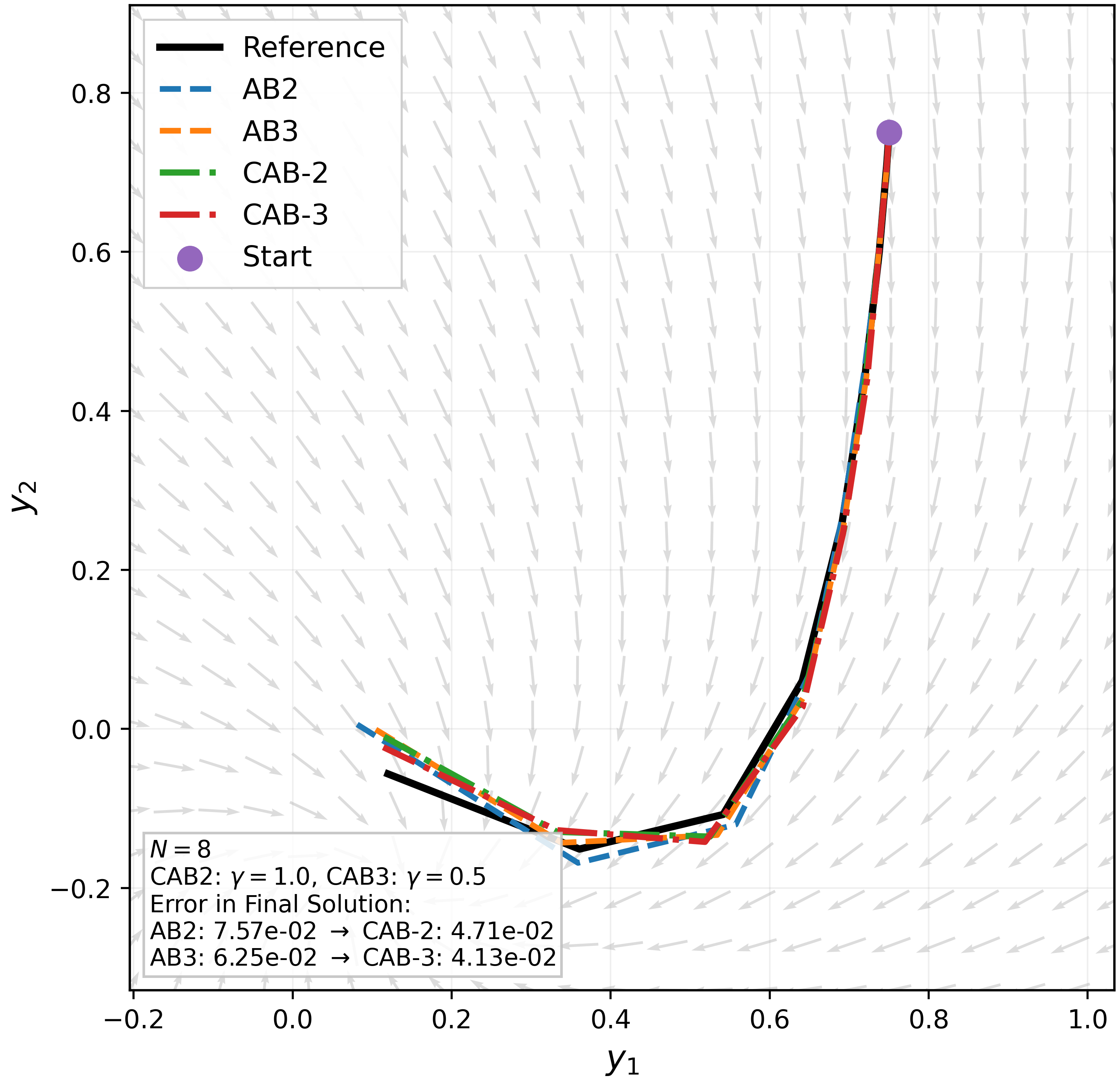}
        \caption{Trajectory comparison for Example 2.}
        \label{fig:cab_vs_ab_traj2}
    \end{subfigure}

    \vspace{0.3em}

    \begin{subfigure}[t]{0.48\textwidth}
        \centering
        \includegraphics[width=\linewidth]{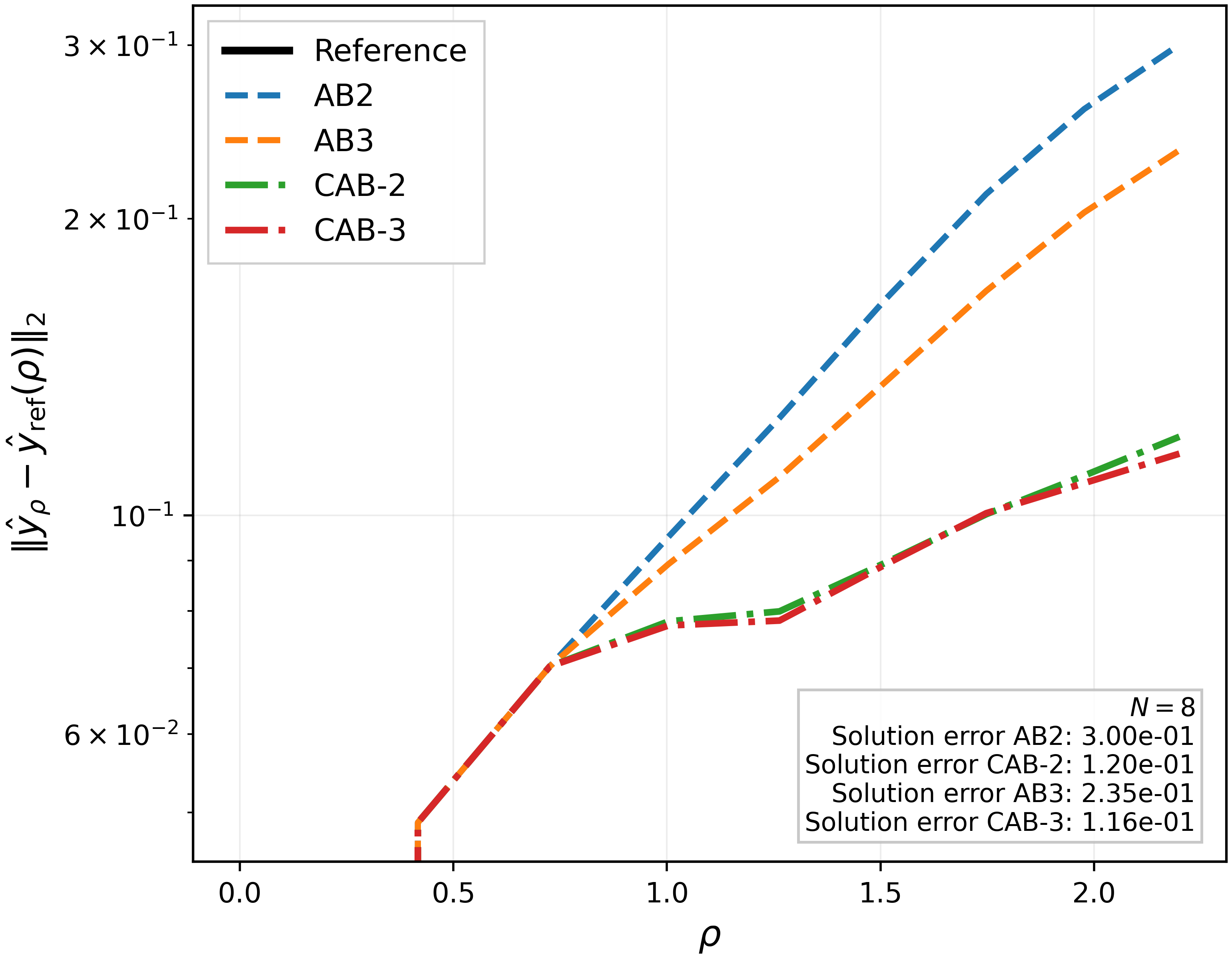}
        \caption{Trajectory error for Example 1.}
        \label{fig:cab_vs_ab_err1}
    \end{subfigure}
    \hfill
    \begin{subfigure}[t]{0.48\textwidth}
        \centering
        \includegraphics[width=\linewidth]{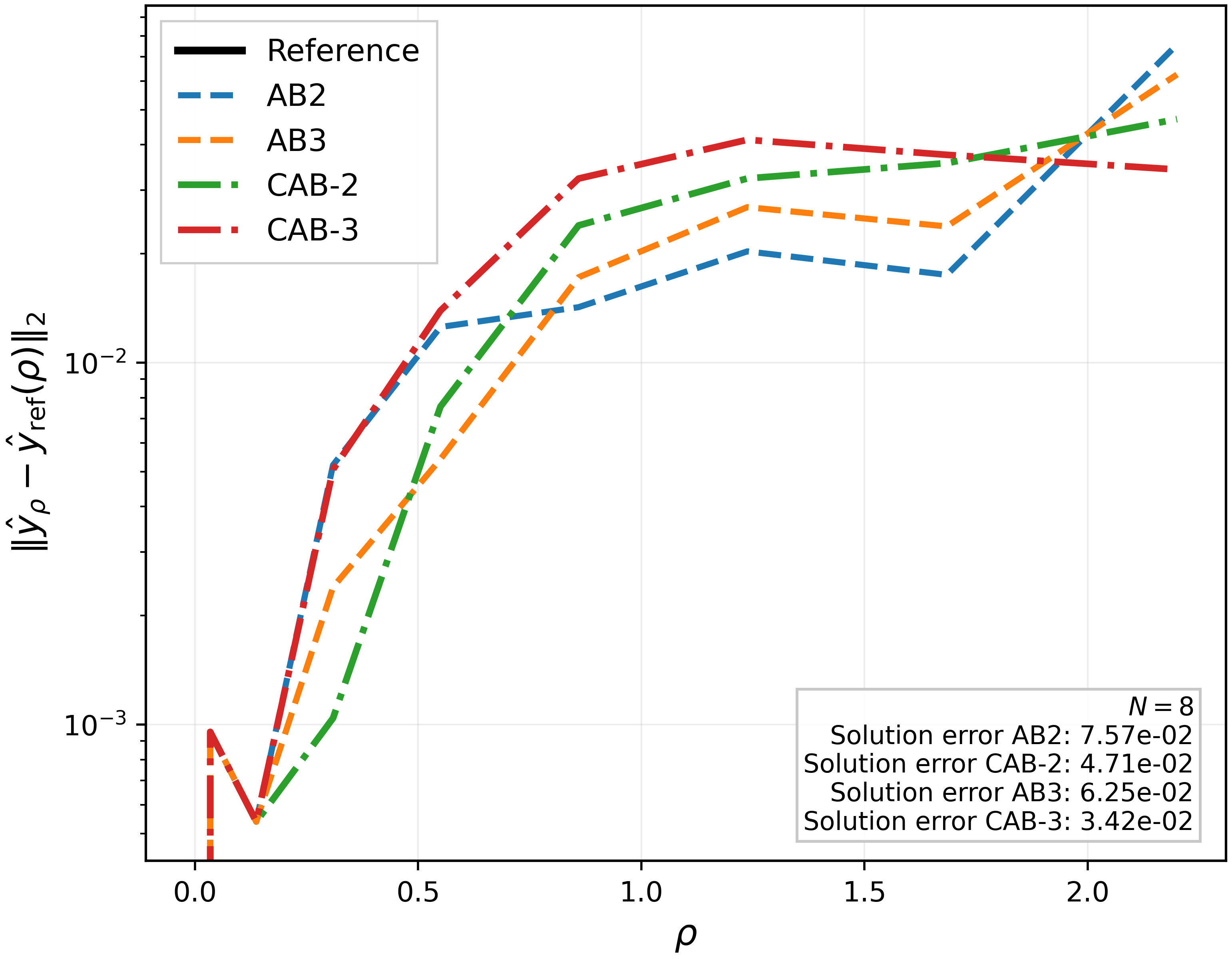}
        \caption{Trajectory error for Example 2.}
        \label{fig:cab_vs_ab_err2}
    \end{subfigure}

    \caption{
    Comparison of AB2/AB3 and the proposed CAB-2/CAB-3 on two representative nonlinear rectified ODEs with a small sampling budget (\(N=8\)). 
    The top row shows the numerical trajectories overlaid on the corresponding velocity fields, and the bottom row shows the trajectory error 
    \(\|\hat{\y}_{\rho}-\hat{\y}_{\mathrm{ref}}(\rho)\|_2\) as a function of \(\rho\).  These visualizations highlight the benefit of the proposed correction in improving low-step integration accuracy to match the exact output.
    }
    \label{fig:cab_vs_ab_examples}
\end{figure*}

Figure~\ref{fig:cab_vs_ab_examples} compares AB2/AB3 and the proposed CAB-2/CAB-3 on two nonlinear rectified ODEs, both using  (\(N=8\)). The top row visualizes the trajectories and the underlying velocity fields, while the bottom row plots the trajectory errors \(\|\hat{\y}_{\rho}-\hat{\y}_{\mathrm{ref}}(\rho)\|_2\) across the steps. In both examples, the corrected schemes track the reference trajectory more faithfully than their uncorrected AB counterparts. In particular, CAB-2 and CAB-3 attain lower final solution errors than AB2 and AB3, respectively. These plots illustrate the practical effect of the correction term: it suppresses the error accumulation visible in AB2/AB3 and provides near-exact output in the low-NFE regime.

\subsection{Additional Experiments}
\begin{figure*}[!htp]
\centering
\setlength{\tabcolsep}{3pt}

\begin{tabular}{ccc}
\includegraphics[width=0.30\linewidth]{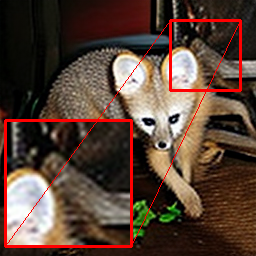} &
\includegraphics[width=0.30\linewidth]{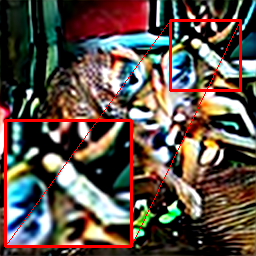} &
\includegraphics[width=0.30\linewidth]{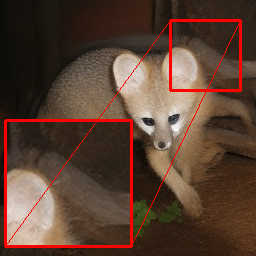} \\

\small \textbf{DPM++} &
\small \textbf{DPM++ (O3)} &
\small \textbf{STORK}
\end{tabular}

\vspace{0.5em}

\begin{tabular}{ccc}
\includegraphics[width=0.30\linewidth]{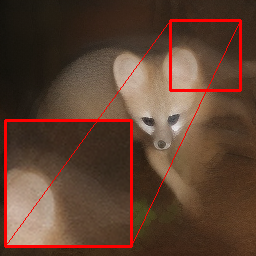} &
\includegraphics[width=0.30\linewidth]{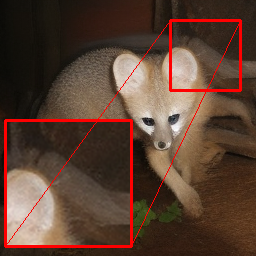} &
\includegraphics[width=0.30\linewidth]{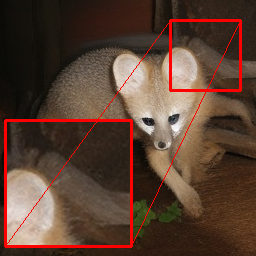} \\

\small \textbf{STORK (O2)} &
\small \textbf{AB2} &
\small \textbf{AB3}
\end{tabular}

\vspace{0.5em}

\begin{tabular}{ccc}
\includegraphics[width=0.30\linewidth]{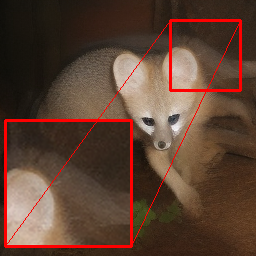} &
\includegraphics[width=0.30\linewidth]{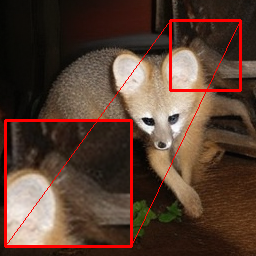} &
\includegraphics[width=0.30\linewidth]{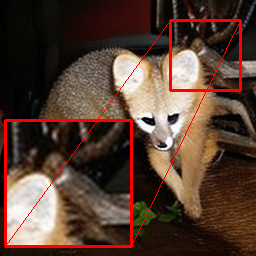} \\

\small \textbf{DDIM} &
\small \textbf{CAB-2} &
\small \textbf{CAB-3}
\end{tabular}

\vspace{0.5em}

\caption{
Qualitative comparison of samplers in the 8-NFE regime for DiT. Increasing the solver order of DPM++ and  STORK can lead to over-correction of the predictor trajectory, producing amplified artifacts and structural distortions in the Grey Fox image, consistent with the quantitative trends in Table~\ref{tab:dit_imagenet_256}. AB predictor algorithms generate images that oversmooth, whereas CAB-based methods maintain improved visual consistency and cleaner structural fidelity under the same sampling budget.
}

\label{fig:order_comparison}
\end{figure*}

In this section, we provide additional qualitative comparisons to complement the main results. The examples show that CAB often produces sharper and more coherent samples than competing tr aining-free solvers, especially in the low-NFE regime where visual artifacts are more pronounced. We also include Figure~\ref{fig:order_comparison}, which illustrates that aggressive higher-order corrections based on previous velocity evaluations can become unstable for strong baseline solvers, in contrast to the controlled correction used by CAB. For DPM-Solver++, UniPC, and STORK, we use implementations from the \texttt{diffusers} library whenever available; otherwise, we use the official code released by the authors, with only minimal changes required to match our evaluation protocol.

\begin{figure*}[!htp]
\centering

\setlength{\tabcolsep}{2pt}
\renewcommand{\arraystretch}{1.0}

\setlength{\fboxrule}{2pt}
\setlength{\fboxsep}{0pt}

\resizebox{\textwidth}{!}{
\begin{tabular}{c c c c}

& \textbf{NFE-8}
& \textbf{NFE-10}
& \textbf{NFE-12} \\[0.4em]

\raisebox{2.0\height}{\rotatebox{90}{\textbf{Euler}}}
&
\fbox{\includegraphics[width=0.31\textwidth]{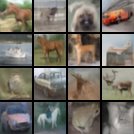}}
&
\fbox{\includegraphics[width=0.31\textwidth]{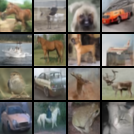}}
&
\fbox{\includegraphics[width=0.31\textwidth]{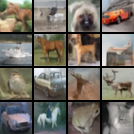}}
\\[0.6em]

\raisebox{1.7\height}{\rotatebox{90}{\textbf{UniPC}}}
&
\fbox{\includegraphics[width=0.31\textwidth]{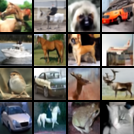}}
&
\fbox{\includegraphics[width=0.31\textwidth]{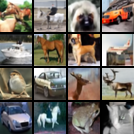}}
&
\fbox{\includegraphics[width=0.31\textwidth]{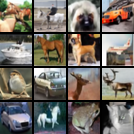}}
\\[0.6em]

\raisebox{1.2\height}{\rotatebox{90}{\textbf{STORK}}}
&
\fbox{\includegraphics[width=0.31\textwidth]{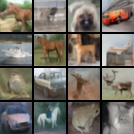}}
&
\fbox{\includegraphics[width=0.31\textwidth]{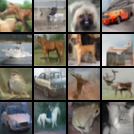}}
&
\fbox{\includegraphics[width=0.31\textwidth]{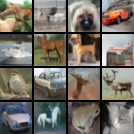}}
\\[0.6em]

\raisebox{1.9\height}{\rotatebox{90}{\textbf{CAB-2}}}
&
\fbox{\includegraphics[width=0.31\textwidth]{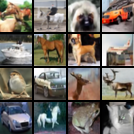}}
&
\fbox{\includegraphics[width=0.31\textwidth]{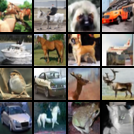}}
&
\fbox{\includegraphics[width=0.31\textwidth]{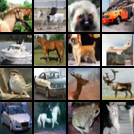}}
\\

\end{tabular}
}

\vspace{0.3em}

\caption{ Unconditional generation on CIFAR-10 using EDM model.}
\label{fig:qualitative_nfe}

\end{figure*}

\begin{figure}[!htp]
    \centering

    \begin{subfigure}[b]{0.235\textwidth}
        \centering
        \includegraphics[width=\textwidth]{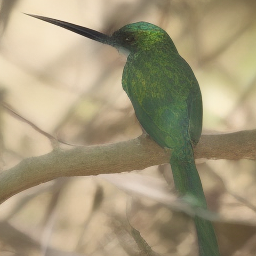}
        \caption{DDIM}
    \end{subfigure}
    \hspace{0.008\textwidth}
    \begin{subfigure}[b]{0.235\textwidth}
        \centering
        \includegraphics[width=\textwidth]{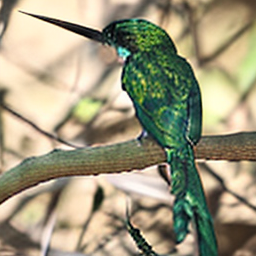}
        \caption{DPM++}
    \end{subfigure}
    \hspace{0.008\textwidth}
    \begin{subfigure}[b]{0.235\textwidth}
        \centering
        \includegraphics[width=\textwidth]{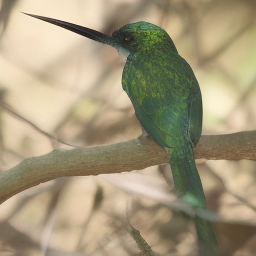}
        \caption{STORK}
    \end{subfigure}
    \hspace{0.008\textwidth}
    \begin{subfigure}[b]{0.235\textwidth}
        \centering
        \includegraphics[width=\textwidth]{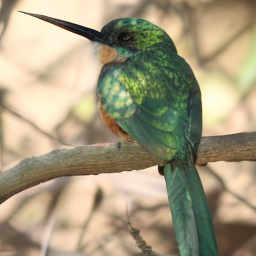}
        \caption{CAB-3}
    \end{subfigure}

\caption*{\textbf{10 NFEs.} CAB-3 produces the most faithful reconstruction with sharper beak structure, consistent texture, and reduced oversmoothing in just 10 NFEs.}

\vspace{0.4em}

    \begin{subfigure}[b]{0.235\textwidth}
        \centering
        \includegraphics[width=\textwidth]{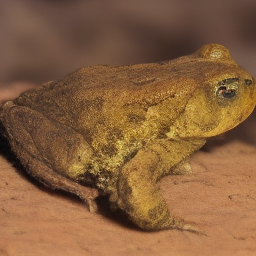}
        \caption{DDIM}
    \end{subfigure}
    \hspace{0.008\textwidth}
    \begin{subfigure}[b]{0.235\textwidth}
        \centering
        \includegraphics[width=\textwidth]{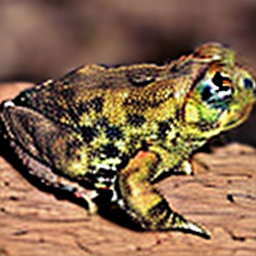}
        \caption{DPM++}
    \end{subfigure}
    \hspace{0.008\textwidth}
    \begin{subfigure}[b]{0.235\textwidth}
        \centering
        \includegraphics[width=\textwidth]{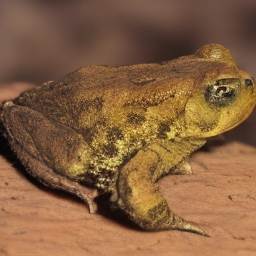}
        \caption{STORK}
    \end{subfigure}
    \hspace{0.008\textwidth}
    \begin{subfigure}[b]{0.235\textwidth}
        \centering
        \includegraphics[width=\textwidth]{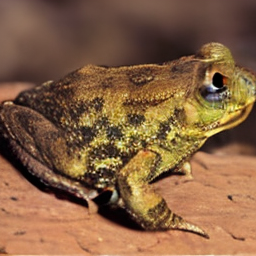}
        \caption{CAB-2}
    \end{subfigure}

\caption*{\textbf{8 NFEs.} CAB-2 best preserves texture, shape, and color, while DDIM/STORK oversmooth and DPM++ introduces artifacts.}

\vspace{0.4em}



    \begin{subfigure}[b]{0.235\textwidth}
        \centering
        \includegraphics[width=\textwidth]{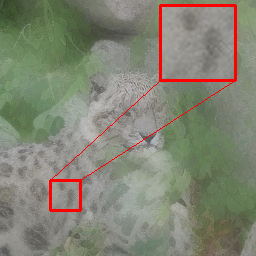}
        \caption{DDIM}
    \end{subfigure}
    \hspace{0.008\textwidth}
    \begin{subfigure}[b]{0.235\textwidth}
        \centering
        \includegraphics[width=\textwidth]{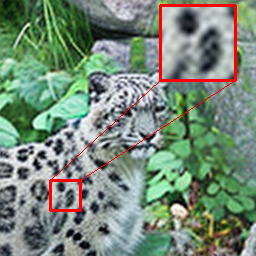}
        \caption{DPM++}
    \end{subfigure}
    \hspace{0.008\textwidth}
    \begin{subfigure}[b]{0.235\textwidth}
        \centering
        \includegraphics[width=\textwidth]{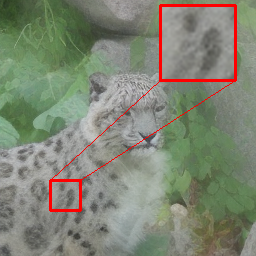}
        \caption{STORK}
    \end{subfigure}
    \hspace{0.008\textwidth}
    \begin{subfigure}[b]{0.235\textwidth}
        \centering
        \includegraphics[width=\textwidth]{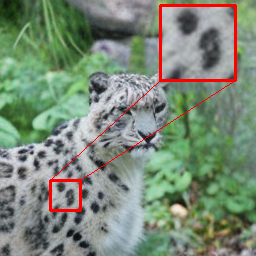}
        \caption{CAB-3}
    \end{subfigure}

\caption*{\textbf{10 NFEs.} DDIM oversmooths, DPM++ is noisy and over-sharpened, STORK generated image is blurred, while CAB-3 gives the most natural image.}

\vspace{0.4em}

    \begin{subfigure}[b]{0.235\textwidth}
        \centering
        \includegraphics[width=\textwidth]{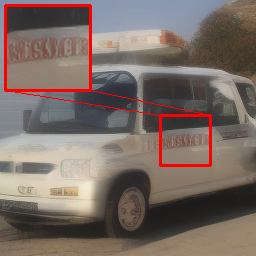}
        \caption{DDIM}
    \end{subfigure}
    \hspace{0.008\textwidth}
    \begin{subfigure}[b]{0.235\textwidth}
        \centering
        \includegraphics[width=\textwidth]{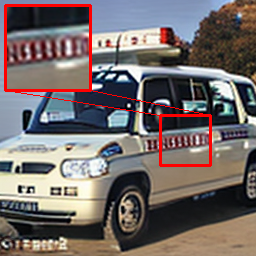}
        \caption{DPM++}
    \end{subfigure}
    \hspace{0.008\textwidth}
    \begin{subfigure}[b]{0.235\textwidth}
        \centering
        \includegraphics[width=\textwidth]{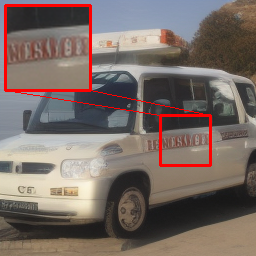}
        \caption{STORK}
    \end{subfigure}
    \hspace{0.008\textwidth}
    \begin{subfigure}[b]{0.235\textwidth}
        \centering
        \includegraphics[width=\textwidth]{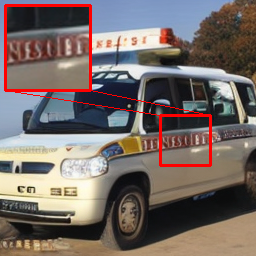}
        \caption{CAB-3}
    \end{subfigure}
\caption*{\textbf{10 NFEs.} DDIM oversmooths and loses detail, DPM++ introduces noise and distorted text, STORK remains blurry, while CAB-3 preserves sharp structure, clean edges, and readable text.}

\caption{Comparison of samplers on class-conditional ImageNet ($256\times256$) generation using DiT.}
\label{fig:dit_comparison2}

\end{figure}

\begin{figure}[!htp]
    \centering

    \begin{subfigure}[b]{0.24\textwidth}
        \centering
        \includegraphics[width=\textwidth]{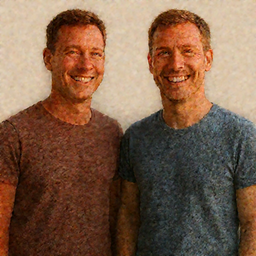}
        \caption{Euler}
    \end{subfigure}
    \hfill
    \begin{subfigure}[b]{0.24\textwidth}
        \centering
        \includegraphics[width=\textwidth]{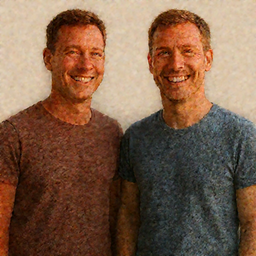}
        \caption{DPM++}
    \end{subfigure}
    \hfill
    \begin{subfigure}[b]{0.24\textwidth}
        \centering
        \includegraphics[width=\textwidth]{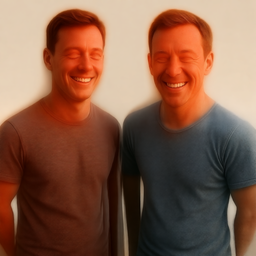}
        \caption{STORK}
    \end{subfigure}
    \hfill
    \begin{subfigure}[b]{0.24\textwidth}
        \centering
        \includegraphics[width=\textwidth]{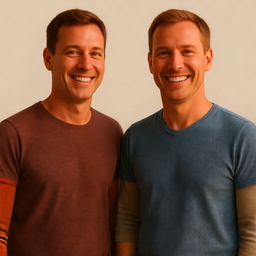}
        \caption{CAB-2}
    \end{subfigure}
\caption*{\textbf{6 NFEs.} \textit{``Two men standing side by side and smiling.''}
CAB-2 better preserves facial detail and natural color, while the baselines show noise, oversmoothing, or loss of detail.}

\vspace{0.2em}
    \begin{subfigure}[b]{0.24\textwidth}
        \centering
        \includegraphics[width=\textwidth]{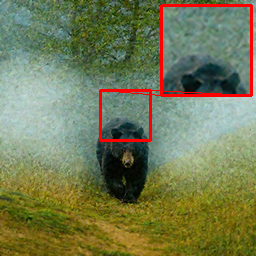}
        \caption{Euler}
    \end{subfigure}
    \hfill
    \begin{subfigure}[b]{0.24\textwidth}
        \centering
        \includegraphics[width=\textwidth]{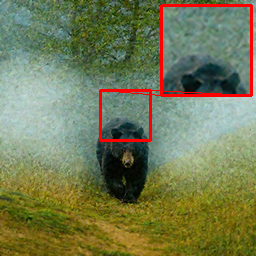}
        \caption{DPM++}
    \end{subfigure}
    \hfill
    \begin{subfigure}[b]{0.24\textwidth}
        \centering
        \includegraphics[width=\textwidth]{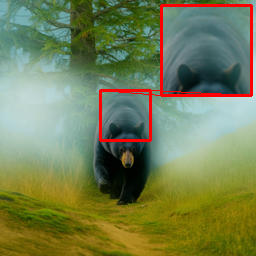}
        \caption{STORK}
    \end{subfigure}
    \hfill
    \begin{subfigure}[b]{0.24\textwidth}
        \centering
        \includegraphics[width=\textwidth]{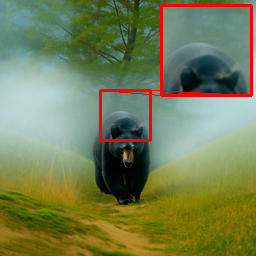}
        \caption{CAB-2}
    \end{subfigure}
\caption*{\textbf{6 NFEs.} \textit{``A black bear walks down a hill away from a tree.''} CAB-2 gives clearer bear structure and surrounding scene content, while competing samplers show haze, noise, or oversmoothing.}

\vspace{0.2em}
    \begin{subfigure}[b]{0.24\textwidth}
        \centering
        \includegraphics[width=\textwidth]{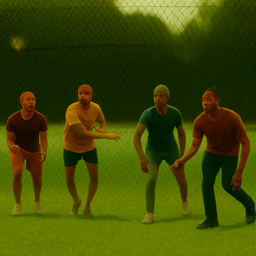}
        \caption{Euler}
    \end{subfigure}
    \hfill
    \begin{subfigure}[b]{0.24\textwidth}
        \centering
        \includegraphics[width=\textwidth]{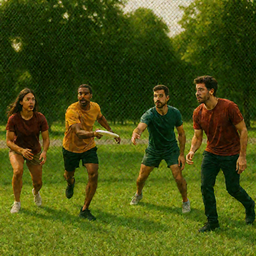}
        \caption{DPM++}
    \end{subfigure}
    \hfill
    \begin{subfigure}[b]{0.24\textwidth}
        \centering
        \includegraphics[width=\textwidth]{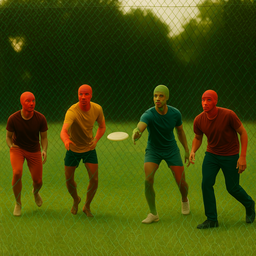}
        \caption{STORK}
    \end{subfigure}
    \hfill
    \begin{subfigure}[b]{0.24\textwidth}
        \centering
        \includegraphics[width=\textwidth]{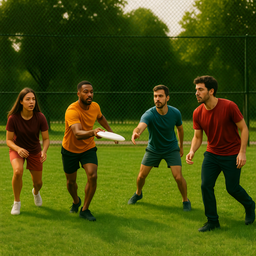}
        \caption{CAB-2}
    \end{subfigure}
\caption*{\textbf{8 NFEs.} \textit{``Four men playing frisbee in a fenced park.''}
CAB-2 better preserves human structure, color, and scene coherence, while baselines exhibit color shifts, artifacts, or oversmoothing.}
\vspace{0.2em}
    \begin{subfigure}[b]{0.24\textwidth}
        \centering
        \includegraphics[width=\textwidth]{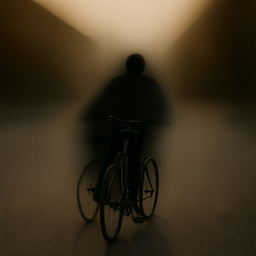}
        \caption{Euler}
    \end{subfigure}
    \hfill
    \begin{subfigure}[b]{0.24\textwidth}
        \centering
        \includegraphics[width=\textwidth]{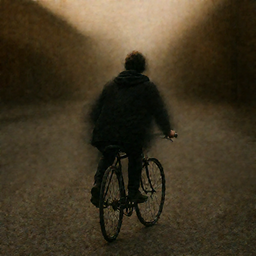}
        \caption{DPM++}
    \end{subfigure}
    \hfill
    \begin{subfigure}[b]{0.24\textwidth}
        \centering
        \includegraphics[width=\textwidth]{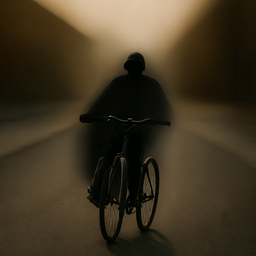}
        \caption{STORK}
    \end{subfigure}
    \hfill
    \begin{subfigure}[b]{0.24\textwidth}
        \centering
        \includegraphics[width=\textwidth]{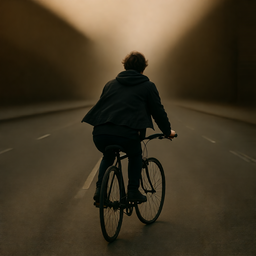}
        \caption{CAB-2}
    \end{subfigure}

\caption*{\textbf{8 NFEs.} \textit{``A person riding a bicycle on a deserted street.''} CAB-2 recovers sharper subject structure and road geometry, while baselines remain noisy, blurred, or overly smooth.}

\caption{Comparison of training-free samplers on QWEN-Image \(1024\times1024\) text-to-image generation.}   
    \label{fig:qwen_comparison2}

\end{figure}

\begin{figure*}[!htp]
\centering
\setlength{\tabcolsep}{2pt}
\renewcommand{\arraystretch}{0.98}

\resizebox{\textwidth}{!}{%
\begin{tabular}{@{}c@{\hspace{3pt}}c@{\hspace{3pt}}c@{\hspace{3pt}}c@{\hspace{3pt}}c@{}}
\toprule
& \textbf{Frame 1} & \textbf{Frame 2} & \textbf{Frame 3} & \textbf{Frame 4} \\
\midrule

\rotatebox{90}{\textbf{DPM++}} &
\includegraphics[width=0.23\textwidth]{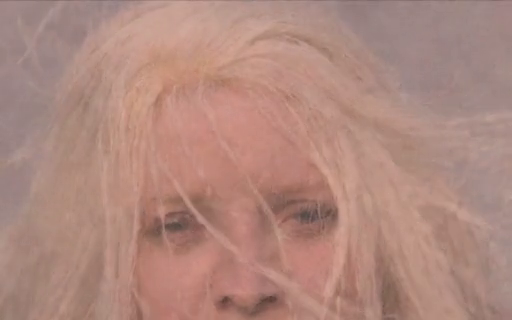} &
\includegraphics[width=0.23\textwidth]{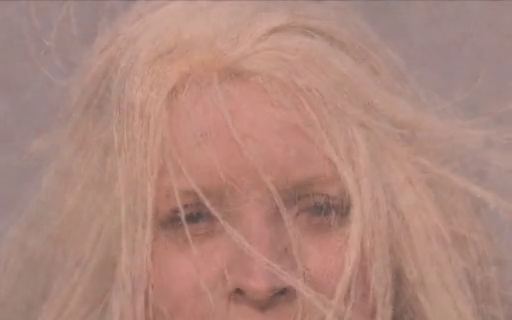} &
\includegraphics[width=0.23\textwidth]{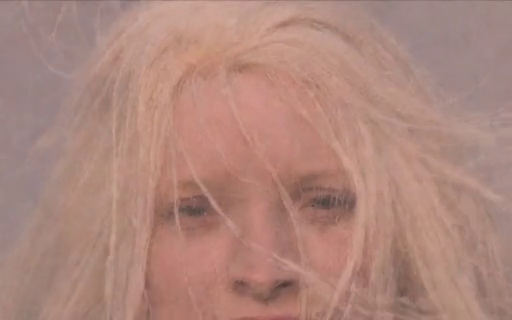} &
\includegraphics[width=0.23\textwidth]{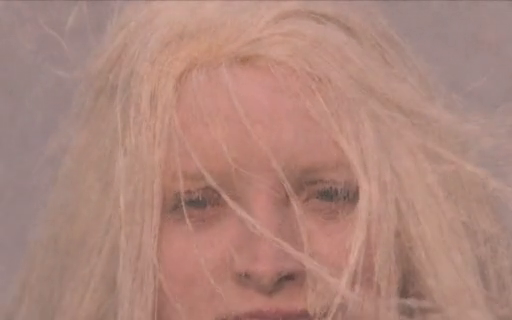} \\[0.35em]

\rotatebox{90}{\textbf{STORK}} &
\includegraphics[width=0.23\textwidth]{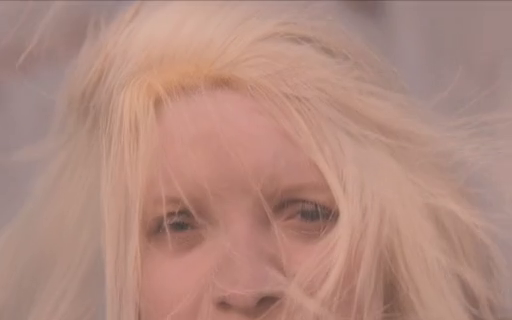} &
\includegraphics[width=0.23\textwidth]{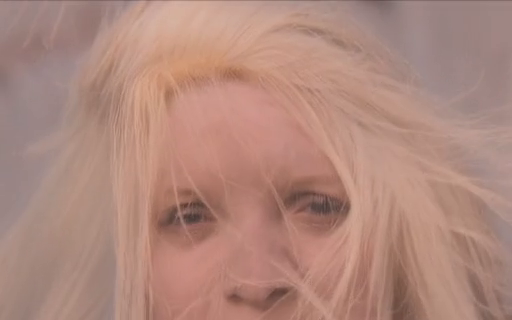} &
\includegraphics[width=0.23\textwidth]{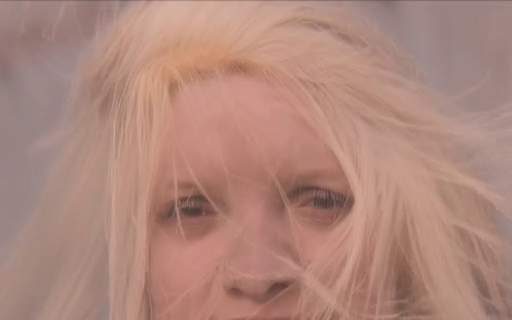} &
\includegraphics[width=0.23\textwidth]{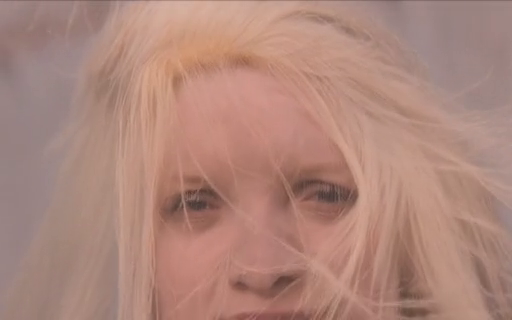} \\[0.35em]

\rotatebox{90}{\textbf{CAB-2}} &
\includegraphics[width=0.23\textwidth]{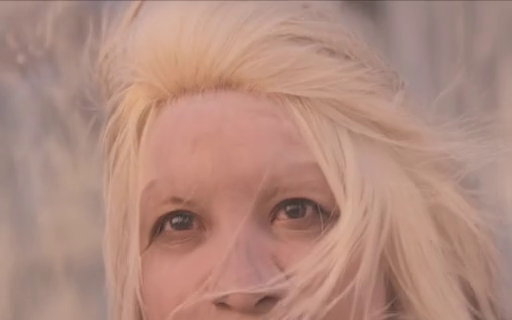} &
\includegraphics[width=0.23\textwidth]{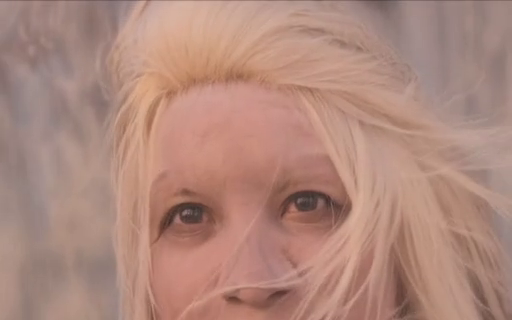} &
\includegraphics[width=0.23\textwidth]{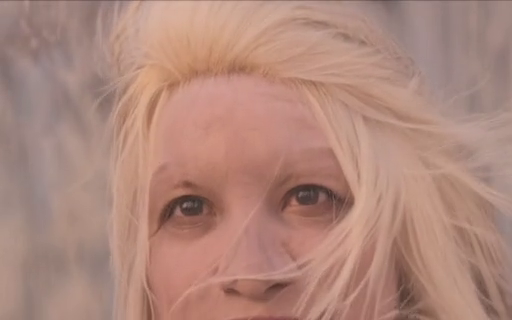} &
\includegraphics[width=0.23\textwidth]{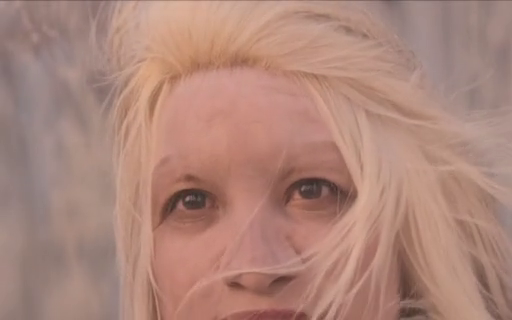} \\

\bottomrule
\end{tabular}%
}

\vspace{0.35em}
\caption{\textbf{Prompt:} \textit{``light wind, feathers moving, she moves her gaze, 4k.''} 
Temporal comparison of training-free samplers on HunyuanVideo-1.5 over four frames from the same generated video. CAB-2 better preserves appearance consistency and smoother motion progression, while DPM++ and STORK show stronger temporal drift and frame-to-frame variation.}
\label{fig:video_compare}
\end{figure*}

\begin{figure*}[!htp]
\centering
\setlength{\tabcolsep}{2pt}
\renewcommand{\arraystretch}{0.98}

\resizebox{\textwidth}{!}{%
\begin{tabular}{@{}c@{\hspace{3pt}}c@{\hspace{3pt}}c@{\hspace{3pt}}c@{\hspace{3pt}}c@{}}
\toprule
& \textbf{Frame 1} & \textbf{Frame 2} & \textbf{Frame 3} & \textbf{Frame 4} \\
\midrule

\rotatebox{90}{\textbf{DPM++}} &
\includegraphics[width=0.23\textwidth]{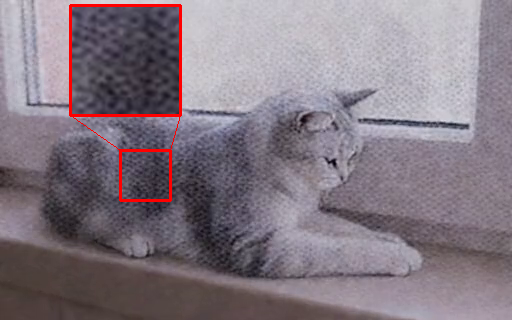} &
\includegraphics[width=0.23\textwidth]{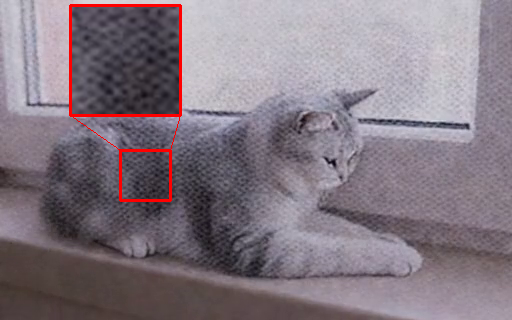} &
\includegraphics[width=0.23\textwidth]{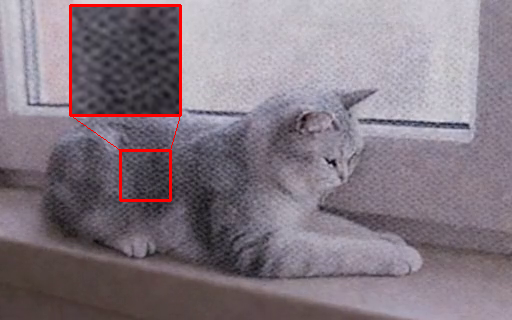} &
\includegraphics[width=0.23\textwidth]{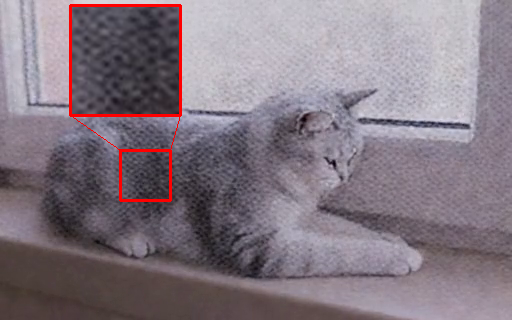} \\[0.35em]

\rotatebox{90}{\textbf{STORK}} &
\includegraphics[width=0.23\textwidth]{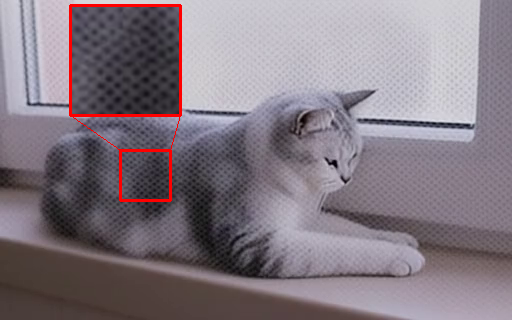} &
\includegraphics[width=0.23\textwidth]{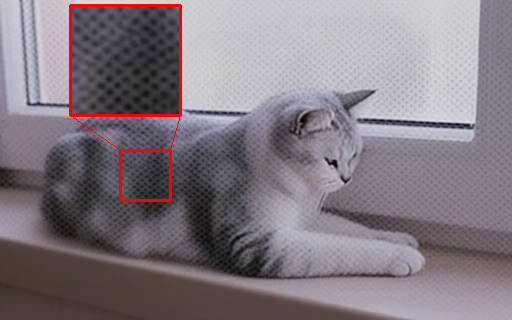} &
\includegraphics[width=0.23\textwidth]{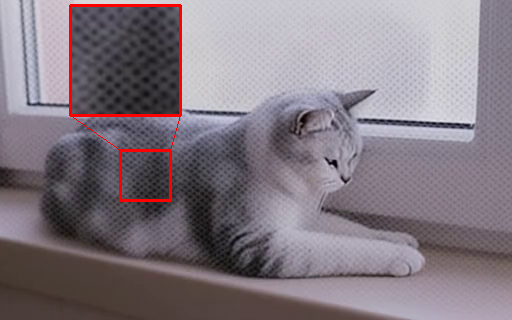} &
\includegraphics[width=0.23\textwidth]{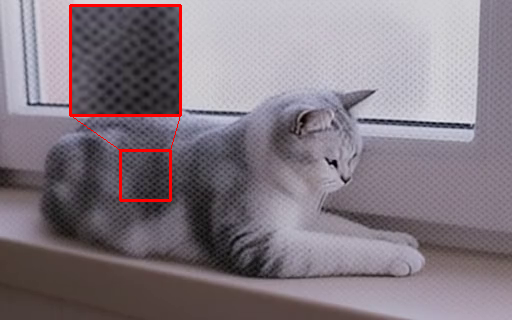} \\[0.35em]

\rotatebox{90}{\textbf{CAB-2}} &
\includegraphics[width=0.23\textwidth]{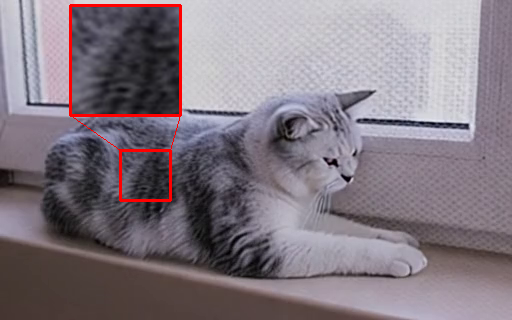} &
\includegraphics[width=0.23\textwidth]{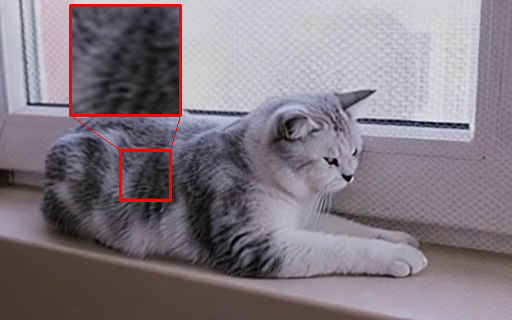} &
\includegraphics[width=0.23\textwidth]{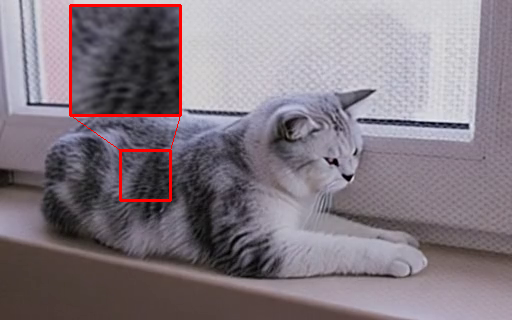} &
\includegraphics[width=0.23\textwidth]{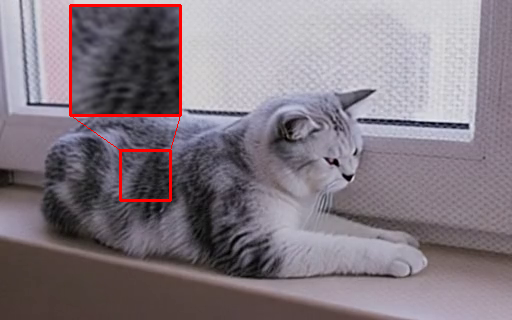} \\

\bottomrule
\end{tabular}%
}

\vspace{0.35em}
\caption{\textbf{Prompt:} \textit{``A fluffy grey and white cat is lazily stretched out on a sunny window sill, enjoying a nap after a long day of lounging.''}
Temporal comparison of training-free samplers on HunyuanVideo-1.5 over four frames from the same generated video. CAB-2 preserves sharper fur texture, clearer facial structure, and more consistent window-side lighting, while DPM++ and STORK appear blurrier and show weaker appearance consistency across frames.}
\label{fig:video_compare_cat}
\end{figure*}

\begin{figure*}[t]
\centering
\setlength{\tabcolsep}{2pt}
\renewcommand{\arraystretch}{0.98}

\resizebox{\textwidth}{!}{%
\begin{tabular}{@{}c@{\hspace{3pt}}c@{\hspace{3pt}}c@{\hspace{3pt}}c@{\hspace{3pt}}c@{}}
\toprule
& \textbf{Frame 1} & \textbf{Frame 2} & \textbf{Frame 3} & \textbf{Frame 4} \\
\midrule

\rotatebox{90}{\textbf{DPM++}} &
\includegraphics[width=0.23\textwidth]{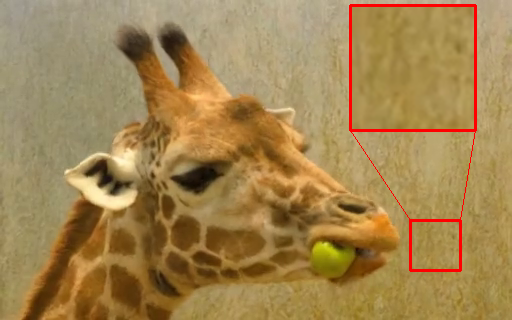} &
\includegraphics[width=0.23\textwidth]{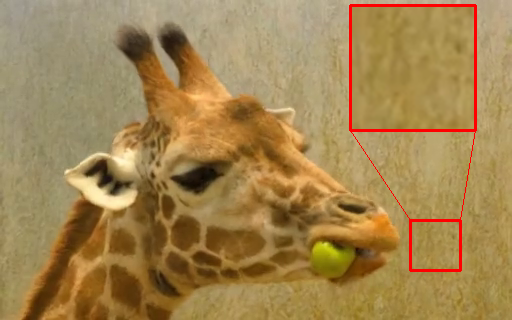} &
\includegraphics[width=0.23\textwidth]{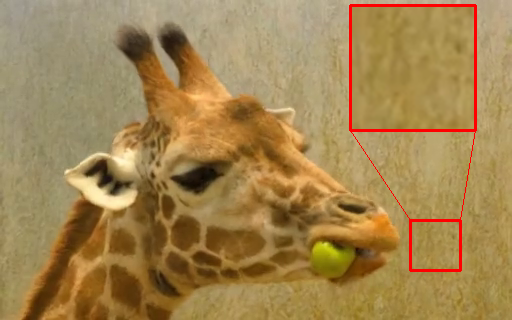} &
\includegraphics[width=0.23\textwidth]{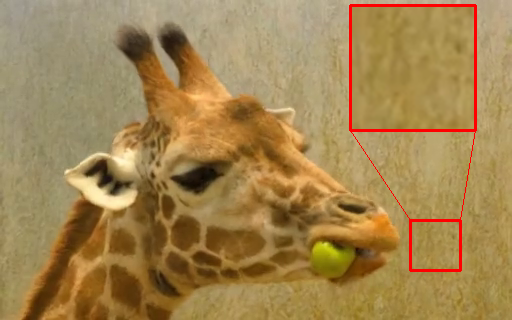} \\[0.35em]

\rotatebox{90}{\textbf{STORK}} &
\includegraphics[width=0.23\textwidth]{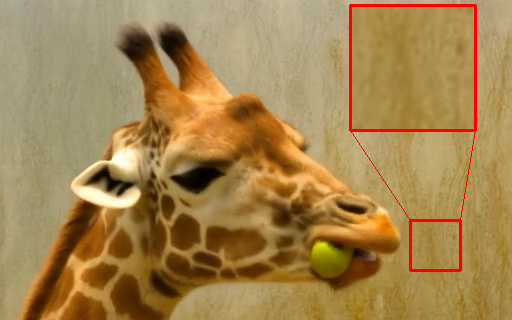} &
\includegraphics[width=0.23\textwidth]{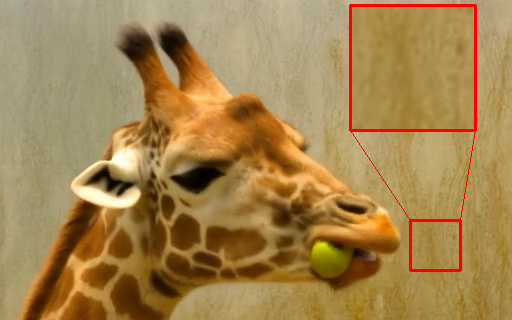} &
\includegraphics[width=0.23\textwidth]{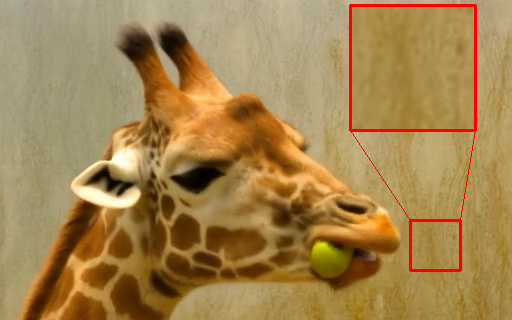} &
\includegraphics[width=0.23\textwidth]{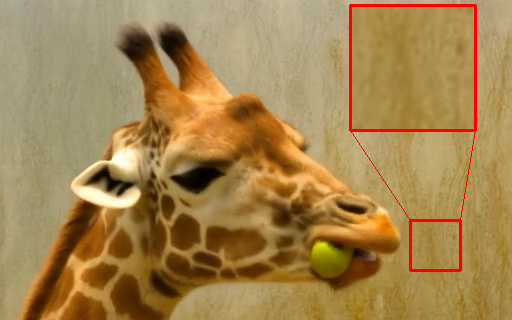} \\[0.35em]

\rotatebox{90}{\textbf{CAB-2}} &
\includegraphics[width=0.23\textwidth]{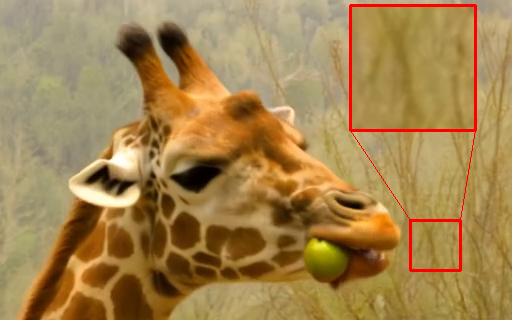} &
\includegraphics[width=0.23\textwidth]{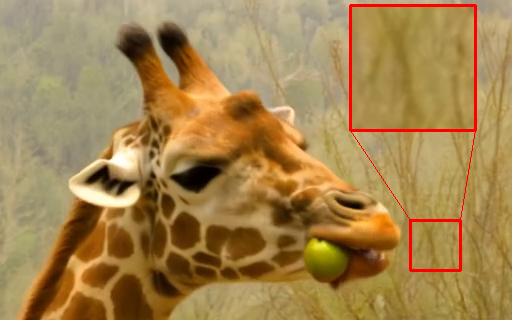} &
\includegraphics[width=0.23\textwidth]{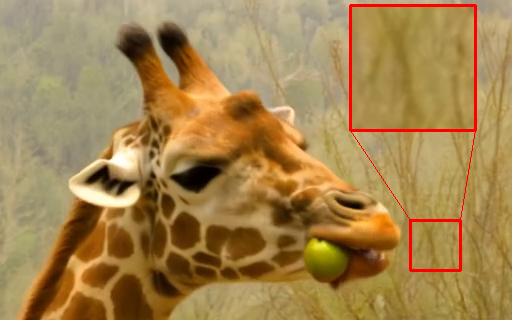} &
\includegraphics[width=0.23\textwidth]{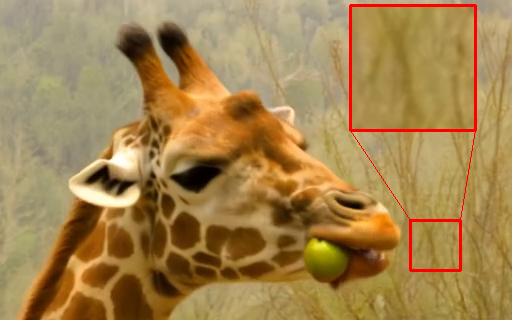} \\

\bottomrule
\end{tabular}%
}

\vspace{0.35em}
\caption{\textbf{Prompt:} \textit{``a giraffe eating an apple.''}
Temporal comparison of training-free samplers on HunyuanVideo-1.5 over four frames from the same generated video. CAB-2 preserves cleaner appearance and more coherent scene context across frames, while DPM++ and STORK show weaker texture consistency and less stable background structure.}
\label{fig:video_compare_giraffe}
\end{figure*}

\begin{table*}[!htp]
\centering
\caption{Extension of Table.~\ref{tab:hunyuan_video_eval_ext} including AB2 and AB3}
\label{tab:hunyuan_video_eval}
\vspace{0.2em}
\scriptsize
\setlength{\tabcolsep}{4.2pt}
\renewcommand{\arraystretch}{1.05}

\begin{tabular}{c l ccccc | c l ccccc}
\toprule
\textbf{Metric} & \textbf{Method}
& \multicolumn{5}{c|}{\textbf{NFE}}
& \textbf{Metric} & \textbf{Method}
& \multicolumn{5}{c}{\textbf{NFE}} \\
\cmidrule(lr){3-7}
\cmidrule(lr){10-14}
& & \textbf{4} & \textbf{6} & \textbf{8} & \textbf{10} & \textbf{20}
& & & \textbf{4} & \textbf{6} & \textbf{8} & \textbf{10} & \textbf{20} \\
\midrule

\multirow{6}{*}{\textbf{Visual}}
& DPM++ & 45.07 & 44.88 & 45.78 & 47.35 & 51.74
&
\multirow{6}{*}{\textbf{Align.}}
& DPM++ & 30.87 & 36.88 & 40.13 & 41.04 & 42.50 \\

& STORK & 45.18 & 47.44 & \textbf{49.35} & 50.47 & \textbf{53.09}
&
& STORK & 35.30 & 37.61 & 40.00 & 40.53 & 42.06 \\

& AB2 & 45.32 & 47.43 & \textbf{49.44} & 50.52 & \textbf{53.15}
&
& AB2 & 34.93 & 37.65 & 39.61 & 40.40 & 42.48 \\

& AB3 & 45.27 & 47.45 & 49.20 & 50.30 & 53.01
&
& AB3 & 34.78 & 37.37 & 40.11 & 40.27 & 42.19 \\

& CAB-2 & \textbf{46.08} & \textbf{47.70} & 49.03 & \textbf{50.54} & 52.71
&
& CAB-2 & \textbf{37.92} & \textbf{39.80} & 40.97 & 42.12 & \textbf{43.17} \\

& CAB-3 & 46.03 & 47.05 & 47.88 & 49.33 & 51.53
&
& CAB-3 & 37.72 & 39.67 & \textbf{41.41} & \textbf{42.34} & 42.73 \\

\midrule

\multirow{6}{*}{\textbf{Temp.}}
& DPM++ & 62.22 & 62.88 & 63.04 & 63.13 & \textbf{63.13}
&
\multirow{6}{*}{\textbf{Final}}
& DPM++ & 138 & 146 & 149 & 152 & 157 \\

& STORK & \textbf{63.71} & \textbf{63.47} & \textbf{63.22} & \textbf{63.30} & 63.03
&
& STORK & 144 & 149 & 152 & 154 & 158 \\

& AB2 & 63.09 & 62.91 & 62.89 & 62.87 & 62.67
&
& AB2 & 143 & 148 & 151 & 154 & 158 \\

& AB3 & 63.14 & 62.95 & 62.92 & 62.91 & 62.61
&
& AB3 & 143 & 148 & 152 & 153 & 158 \\

& CAB-2 & 62.84 & 62.72 & 62.78 & 63.04 & 62.86
&
& CAB-2 & \textbf{147} & \textbf{150} & \textbf{153} & \textbf{156} & \textbf{159} \\

& CAB-3 & 62.73 & 62.42 & 62.45 & 62.73 & 62.72
&
& CAB-3 & 146 & 149 & 151 & 154 & 157 \\

\bottomrule
\end{tabular}

\vspace{-0.4em}
\end{table*}

\FloatBarrier

\subsubsection{Comparison with Distillation-Based Fast Sampling}
\label{app:distillation_comparison}

To contextualize training-free acceleration against distillation-based sampling, we compare CAB-2 applied to the original Qwen-Image model with Qwen-Image-Lightning~\cite{fan2025qwenimagelightning}, a recent distribution-matching-distilled model optimized for few-step inference using an Euler sampler. The developers report using approximately 420,000 text prompts for the model-specific distillation process. The released distilled models target 4- and 8-step generation and require substantial additional training, whereas CAB is applied directly to the pretrained model without retraining.

Table~\ref{tab:qwen_distillation_comparison} presents the quality--latency comparison between Qwen-Image-Lightning and training-free CAB-2, evaluated on the same prompt set. Qwen-Image-Lightning achieves higher quality and approximately 17--19\% lower latency in the very-low-step regime, as expected for a model explicitly distilled for few-step generation. Nevertheless, CAB-2 becomes increasingly competitive as the NFE increases: the ImageReward gap decreases from 1.1981 at 4 NFEs to 0.2591 at 8 NFEs, where CAB-2 reaches an ImageReward of 1.0085 without additional training data or distillation cost.

\begin{table}[!htbp]
\centering
\caption{Quality--latency comparison of distilled Qwen-Image-Lightning and training-free CAB-2 on the same prompts.}
\label{tab:qwen_distillation_comparison}
\small
\setlength{\tabcolsep}{6pt}
\renewcommand{\arraystretch}{1.12}
\begin{tabular}{@{}c cc cc@{}}
\toprule
& \multicolumn{2}{c}{\textbf{Qwen-Image-Lightning}}
& \multicolumn{2}{c}{\textbf{Qwen-Image + CAB-2}} \\
\cmidrule(lr){2-3}\cmidrule(l){4-5}
\textbf{NFE} & \textbf{ImageReward $\uparrow$} & \textbf{Time (s) $\downarrow$}
& \textbf{ImageReward $\uparrow$} & \textbf{Time (s) $\downarrow$} \\
\midrule
4 & \textbf{1.2514} & \textbf{3.497} & 0.0533 & 4.208 \\
6 & \textbf{1.2637} & \textbf{5.138} & 0.7536 & 6.169 \\
8 & \textbf{1.2676} & \textbf{6.590} & 1.0085 & 8.142 \\
\bottomrule
\end{tabular}
\vspace{-2mm}
\end{table}

\FloatBarrier

\subsubsection{Compatibility with Cache-Based Acceleration}
\label{app:cache_compatibility}

We further investigate whether CAB can be combined with
cache-based acceleration methods such as TaylorSeer~\cite{liu2025taylorseer}, which
reduces computational cost by approximating intermediate
transformer features instead of recomputing them at every step. Such approximations may introduce errors in the velocity evaluations stored in CAB's multistep history, potentially affecting its extrapolation-defect correction.

To assess compatibility, we integrate CAB-2 with TaylorSeer on Qwen-Image and compare it against TaylorSeer with the default Euler sampler. Both configurations use identical denoising steps, timestep grids, caching settings, prompts, and random seeds. As shown in Table~\ref{tab:cab_taylorseer},
CAB-2 consistently improves ImageReward from $-0.3115$
to $-0.1443$ at $6$ steps, $0.1991$ to $0.3618$ at
$8$ steps, and $0.7079$ to $0.7785$ at $10$ steps.
These results demonstrate that CAB's correction remains
effective under the tested feature-caching approximations,
supporting its compatibility with cache-based acceleration.

\subsubsection{Applicability to Distilled Models}
\label{app:distilled_models}

Although CAB is designed as a training-free acceleration method, its correction mechanism can also be applied to models already optimized for few-step generation. To examine this compatibility, we compare Euler and CAB-2 on Qwen-Image-Lightning across $2$--$8$ NFEs. As shown in Table~\ref{tab:cab_distilled_models}, CAB-2 consistently improves ImageReward at every evaluated
budget, demonstrating that its multistep formulation can
complement distillation-based acceleration.

The gains are modest at $4$--$8$ NFEs, where the distilled model is already optimized for Euler sampling and only a few correction steps are available. At $2$ NFEs, the larger improvement arises from the modified startup update rather than the full extrapolation-defect correction. Overall, these findings establish compatibility with distilled models and motivate integrating CAB into the student sampling trajectory during distribution-matching distillation, potentially combining solver-aware training with improved
low-NFE integration.


\begin{table}[!htbp]
\centering

\begin{minipage}[t]{0.32\linewidth}
\vspace{0pt}
\centering
\captionsetup{justification=raggedright,
singlelinecheck=false}

\caption{CAB-2 with TaylorSeer on Qwen-Image.}
\label{tab:cab_taylorseer}
\vspace{2pt}

{\scriptsize
\setlength{\tabcolsep}{2pt}
\renewcommand{\arraystretch}{0.8}
\begin{tabular}{@{}c r r r@{}}
\toprule
\textbf{Steps}
& \textbf{Euler}
& \textbf{CAB-2}
& \textbf{Gain} \\
\midrule
$6$  & $-0.3115$ & $\mathbf{-0.1443}$ & $+0.1672$ \\
$8$  & $0.1991$  & $\mathbf{0.3618}$  & $+0.1627$ \\
$10$ & $0.7079$  & $\mathbf{0.7785}$  & $+0.0706$ \\
\bottomrule
\end{tabular}
}
\end{minipage}%
\hfill
\begin{minipage}[t]{0.34\linewidth}
\vspace{0pt}
\centering
\captionsetup{justification=raggedright,
singlelinecheck=false}

\caption{CAB-2 on Qwen-Image-Lightning.}
\label{tab:cab_distilled_models}
\vspace{2pt}

{\scriptsize
\setlength{\tabcolsep}{3pt}
\renewcommand{\arraystretch}{1.15}
\begin{tabular}{@{}l r r r r@{}}
\toprule
\textbf{Method}
& \textbf{$2$}
& \textbf{$4$}
& \textbf{$6$}
& \textbf{$8$} \\
\midrule
Euler
& $1.0917$ & $1.2531$ & $1.2699$ & $1.2722$ \\
CAB-2
& $\mathbf{1.2479}$
& $\mathbf{1.2684}$
& $\mathbf{1.2756}$
& $\mathbf{1.2769}$ \\
\bottomrule
\end{tabular}
}
\end{minipage}%
\hfill
\begin{minipage}[t]{0.32\linewidth}
\vspace{0pt}
\centering
\captionsetup{justification=raggedright,
singlelinecheck=false}

\caption{CAB-2 rectification on DiT/ImageNet-256.
FID $\downarrow$.}
\label{tab:rectification_comparison_main}
\vspace{2pt}

{\scriptsize
\setlength{\tabcolsep}{3pt}
\renewcommand{\arraystretch}{1.15}
\begin{tabular}{@{}l r r r@{}}
\toprule
\textbf{Method}
& \textbf{$6$}
& \textbf{$8$}
& \textbf{$10$} \\
\midrule
$\theta$-space
& $73.08$ & $14.19$ & $8.51$ \\
$\rho$-space
& $\mathbf{15.13}$
& $\mathbf{9.21}$
& $\mathbf{7.99}$ \\
\bottomrule
\end{tabular}
}
\end{minipage}

\vspace{-2mm}
\end{table}

\FloatBarrier

\subsubsection[Native x0- and v-Prediction Rectifications]
{Native $x_0$- and $v$-Prediction Rectifications}
\label{app:native_rectifications}

The $\epsilon$--$\rho$ formulation is not unique.
For the affine-Gaussian path
$x_t=s_t x_0+\sigma_t\epsilon$, an $x_0$-prediction
model admits the rectification
\begin{equation}
z_\kappa=\frac{x_t}{\sigma_t},\qquad
\kappa_t=\frac{s_t}{\sigma_t}=\rho_t^{-1},
\qquad
\frac{dz_\kappa}{d\kappa}
=\hat{x}_{0,\theta}(x_t,t).
\label{eq:native_x0_rectification}
\end{equation}
Thus, the predicted clean sample directly defines
the rectified velocity field in $\kappa$-space.

For $v$-prediction, defining
$r_t=\sqrt{s_t^2+\sigma_t^2}$,
$\theta_t=\operatorname{atan2}(\sigma_t,s_t)$,
and $z_\theta=x_t/r_t$ yields
\begin{equation}
\frac{dz_\theta}{d\theta}
=-\sin\theta_t\,x_0+\cos\theta_t\,\epsilon
=\frac{v_\theta(x_t,t)-\dot r_t z_\theta}
{r_t\dot\theta_t}.
\label{eq:native_v_rectification}
\end{equation}
Here, $v_\theta$ denotes the predicted physical-time
velocity. CAB's Adams--Bashforth predictor and
extrapolation-defect correction apply to these
rectified dynamics without additional NFEs.

Table~\ref{tab:rectification_comparison_main}
compares the $\rho$-space ($\epsilon$-prediction)
and $\theta$-space ($v$-prediction) formulations
on DiT/ImageNet-256. The $\rho$-space formulation
consistently achieves lower FID, particularly
in the low-NFE regime.


\subsubsection{Third-Order Corrector for AB3}
\label{app:third_order_corrector}

Following the Lagrange-extrapolation derivation, quadratic extrapolation from the previous three network evaluations gives, with $h_j=\rho_{j+1}-\rho_j$,
\begin{equation}
\begin{aligned}
\epsilon_i^{\mathrm{ext},3}
={}&
\frac{
(h_{i-1}+h_{i-2})
(h_{i-1}+h_{i-2}+h_{i-3})
}{
h_{i-2}(h_{i-2}+h_{i-3})
}
\epsilon_{i-1}
\\
&-
\frac{
h_{i-1}
(h_{i-1}+h_{i-2}+h_{i-3})
}{
h_{i-2}h_{i-3}
}
\epsilon_{i-2}
\\
&+
\frac{
h_{i-1}(h_{i-1}+h_{i-2})
}{
h_{i-3}(h_{i-2}+h_{i-3})
}
\epsilon_{i-3}.
\end{aligned}
\label{eq:third_order_extrapolation}
\end{equation}

The corresponding extrapolation defect is
\begin{equation}
d_i^{(3)}
=
\epsilon_i-\epsilon_i^{\mathrm{ext},3}.
\label{eq:third_order_defect}
\end{equation}

As shown in Table~\ref{tab:third_order_corrector}, the third-order corrector improves FID at 4--20 NFEs, as expected from higher-order extrapolation, while requiring one additional stored model evaluation but no additional NFEs. At larger sampling budgets, the second-order corrector performs slightly better.

\begin{table}[!htbp]
\centering
\caption{Second- vs. third-order correction for CAB-3 on CIFAR-10(VP).}
\label{tab:third_order_corrector}
\scriptsize
\setlength{\tabcolsep}{5pt}
\renewcommand{\arraystretch}{1.10}
\begin{tabular}{@{}l rrrrrrrr@{}}
\toprule
\textbf{Corrector} & \multicolumn{8}{c}{\textbf{NFE}} \\
\cmidrule(l){2-9}
& \textbf{4} & \textbf{6} & \textbf{8} & \textbf{10} & \textbf{20} & \textbf{30} & \textbf{40} & \textbf{50} \\
\midrule
CAB-3 + 2nd-order & 22.8001 & 7.6202 & 4.1510 & 2.9953 & 1.9157 & \textbf{1.7906} & \textbf{1.7656} & \textbf{1.7708} \\
CAB-3 + 3rd-order & \textbf{22.7895} & \textbf{7.5886} & \textbf{4.1110} & \textbf{2.9353} & \textbf{1.8721} & 1.8128 & 1.8282 & 1.8512 \\
\bottomrule
\end{tabular}
\vspace{-2mm}
\end{table}

\FloatBarrier

\subsubsection{Robustness Across CFG Strengths}
\label{app:cfg_robustness}

The optimal correction strength $\gamma$ may depend on classifier-free guidance (CFG), since guidance changes the magnitude and variation of the learned ODE field. Nevertheless, a fixed $\gamma=0.9$ remains effective across CFG values from $1.25$ to $1.50$, indicating robustness to moderate guidance changes. As shown in Table~\ref{tab:cfg_robustness}, the sampling performance remains stable across the evaluated NFEs, with CFG $=1.50$ achieving the lowest FID throughout. This does not imply universal optimality, and tuning $\gamma$ for individual CFG settings may further improve performance.

\begin{table}[!htbp]
\centering
\caption{CAB-2 with fixed $\gamma=0.9$ across CFG strengths (FID $\downarrow$).}
\label{tab:cfg_robustness}
\small
\setlength{\tabcolsep}{6pt}
\renewcommand{\arraystretch}{1.10}
\begin{tabular}{@{}c rrrrrrrr@{}}
\toprule
\textbf{CFG} & \multicolumn{8}{c}{\textbf{NFE}} \\
\cmidrule(l){2-9}
& \textbf{4} & \textbf{6} & \textbf{8} & \textbf{10} & \textbf{20} & \textbf{30} & \textbf{40} & \textbf{50} \\
\midrule
1.25 & 88.04 & 15.13 & 9.21 & 7.99 & 6.86 & 6.86 & 6.91 & 6.95 \\
1.35 & 81.00 & 12.88 & 7.87 & 6.85 & 5.81 & 5.72 & 5.74 & 5.78 \\
1.50 & \textbf{71.68} & \textbf{10.56} & \textbf{6.77} & \textbf{6.08} & \textbf{5.31} & \textbf{5.19} & \textbf{5.19} & \textbf{5.20} \\
\bottomrule
\end{tabular}
\vspace{-2mm}
\end{table}

\FloatBarrier

\subsubsection{Effect of Discretization Time Grid}
\label{app:time_grid}

The original experiments use uniform spacing in $t$; after rectification, the corresponding noise-to-signal coordinate $\rho$ generally follows a non-uniform grid. Since the discretization grid affects finite-step sampling performance, we further investigate the effect of non-uniform timestep selection.

As shown in Table~\ref{tab:time_grid}, a non-uniform time grid improves ImageReward at every evaluated NFE, particularly in the low-NFE regime. For instance, ImageReward increases from $0.3393$ to $0.7536$ at 6 NFEs and from $0.6561$ to $1.0085$ at 8 NFEs. These results demonstrate that CAB is compatible with different discretization grids and can benefit substantially from improved timestep design.

\begin{table}[!htbp]
\centering
\caption{Uniform vs. non-uniform timestep grids for CAB-2 on Qwen-Image (ImageReward $\uparrow$).}
\label{tab:time_grid}
\scriptsize
\setlength{\tabcolsep}{5pt}
\renewcommand{\arraystretch}{1.10}
\begin{tabular}{@{}l rrrrrrrr@{}}
\toprule
\textbf{Grid} & \multicolumn{8}{c}{\textbf{NFE}} \\
\cmidrule(l){2-9}
& \textbf{4} & \textbf{6} & \textbf{8} & \textbf{10} & \textbf{20} & \textbf{30} & \textbf{40} & \textbf{50} \\
\midrule
Uniform & $-0.3544$ & $0.3393$ & $0.6561$ & $0.8689$ & $1.2026$ & $1.2655$ & $1.2707$ & $1.2866$ \\
Non-uniform & $\mathbf{0.0533}$ & $\mathbf{0.7536}$ & $\mathbf{1.0085}$ & $\mathbf{1.1240}$ & $\mathbf{1.2777}$ & $\mathbf{1.2867}$ & $\mathbf{1.2937}$ & $\mathbf{1.3104}$ \\
\bottomrule
\end{tabular}
\vspace{-2mm}
\end{table}

\FloatBarrier

\subsection{Limitations}
\label{sec:limitations}
\begin{figure}[!htp]
    \centering
    \begin{subfigure}[t]{0.32\textwidth}
        \centering
        \includegraphics[width=\textwidth]{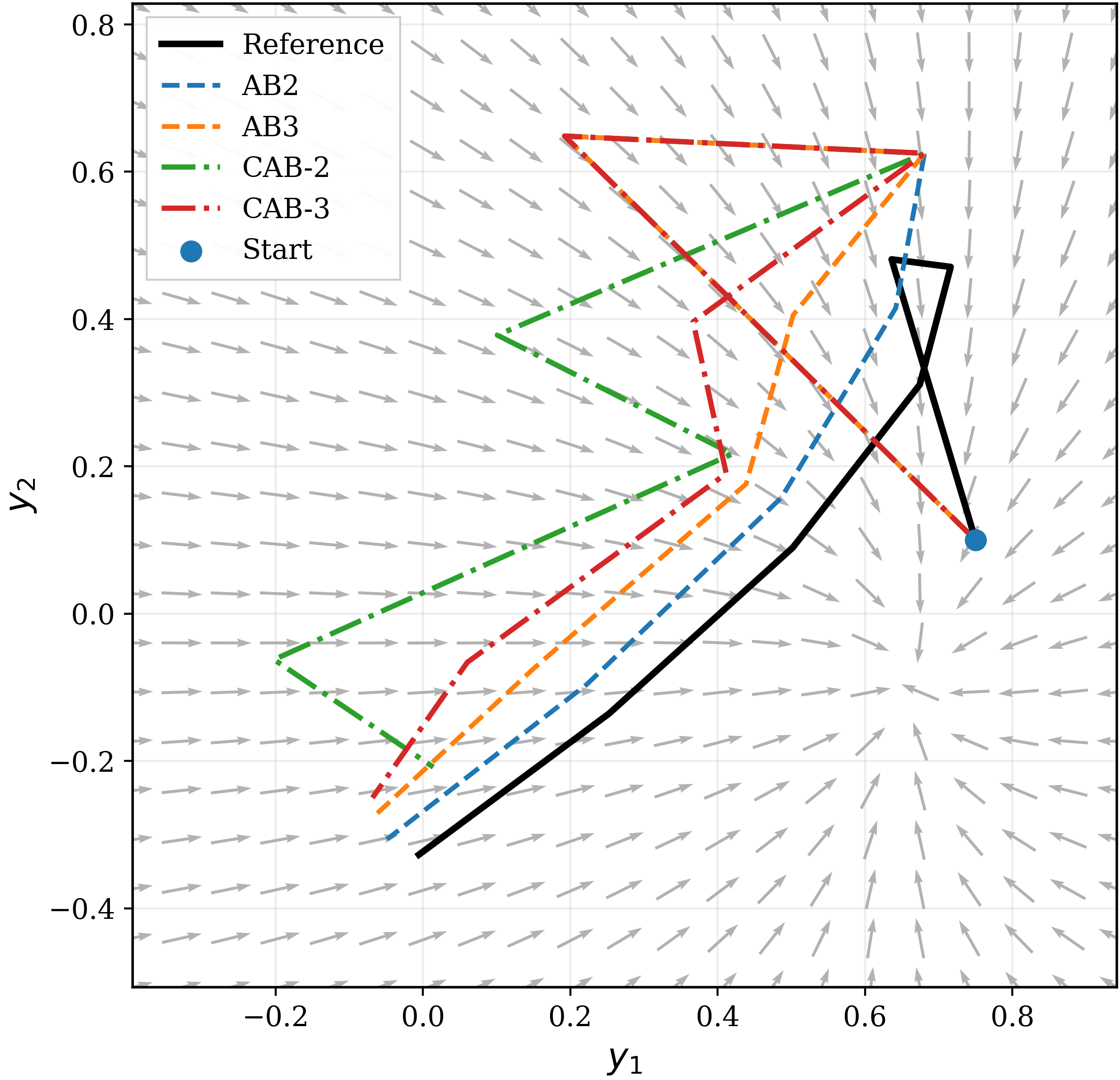}
        \caption{\(N=6\)}
        \label{fig:traj_n6}
    \end{subfigure}
    \hfill
    \begin{subfigure}[t]{0.32\textwidth}
        \centering
        \includegraphics[width=\textwidth]{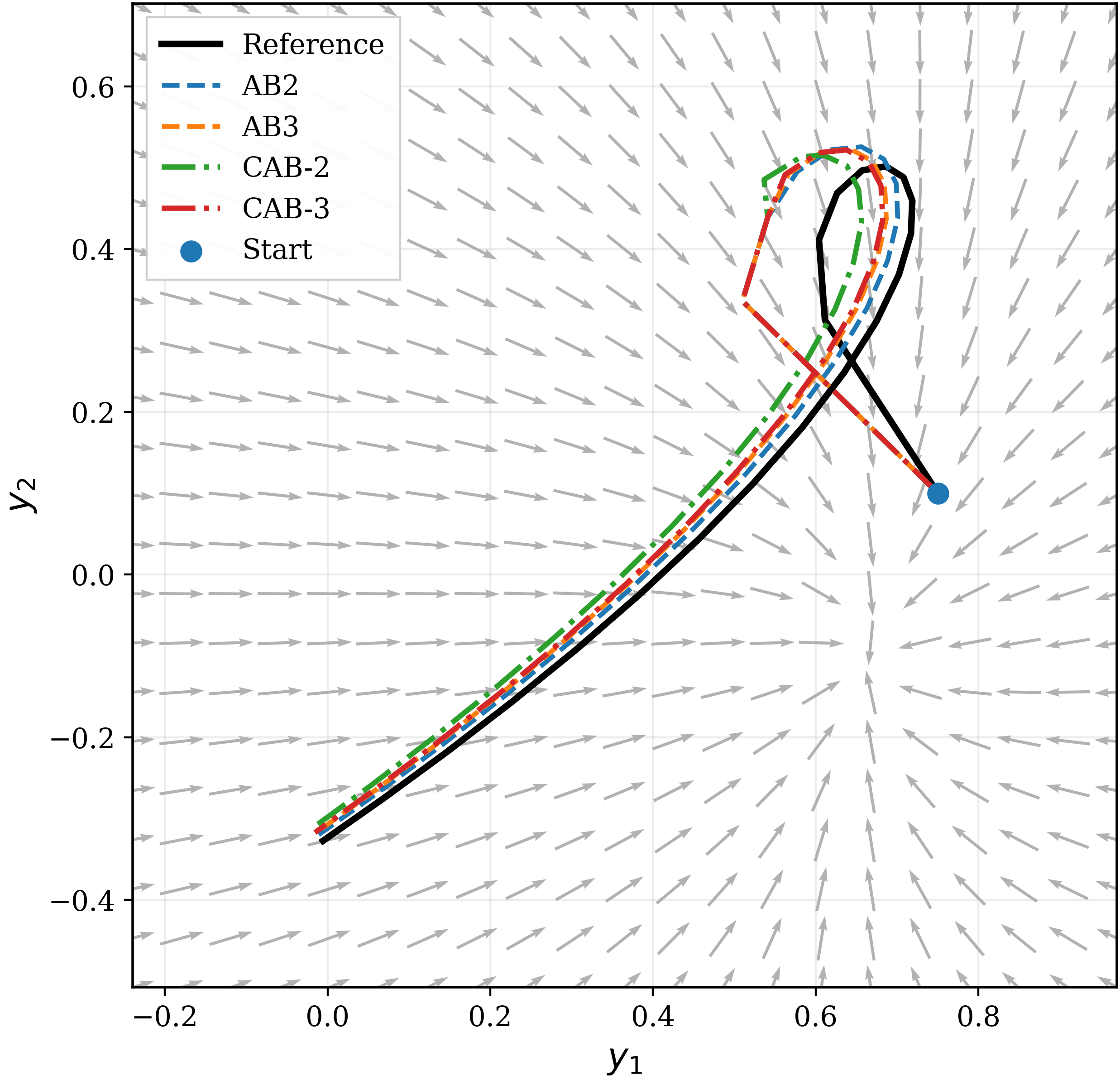}
        \caption{\(N=20\)}
        \label{fig:traj_n20}
    \end{subfigure}
    \hfill
    \begin{subfigure}[t]{0.32\textwidth}
        \centering
        \includegraphics[width=\textwidth]{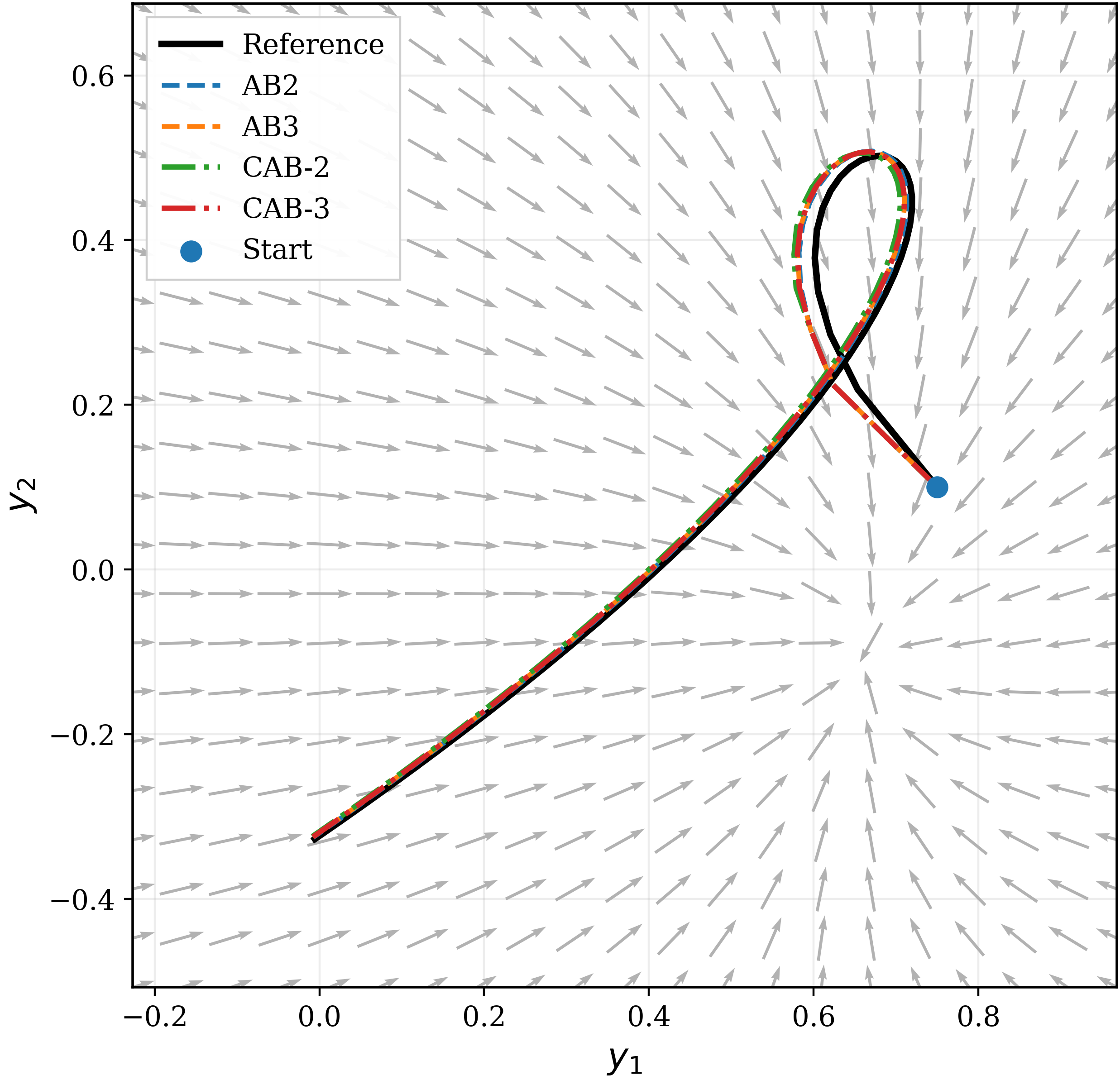}
        \caption{\(N=50\)}
        \label{fig:traj_n50}
    \end{subfigure}
    \caption{Trajectories produced by AB2, AB3, CAB-2, and CAB-3, compared with the reference solution, for different sampling budgets. At very low step counts, particularly \(N=6\), the numerical trajectories can deviate substantially from the reference path, revealing geometric mis-tracking in challenging regions of the velocity field. As the number of sampling steps increases, the trajectories become better aligned with the reference.}
    \label{fig:trajectory_comparison_steps}
\end{figure}

As illustrated in Figure~\ref{fig:trajectory_comparison_steps}, although CAB-2 and CAB-3 perform well in flow and diffusion sampling settings after appropriate rectification, their behavior can degrade in strongly curved or poorly conditioned dynamical regimes, especially at very small step budgets. In particular, similar to AB2 and AB3, when the number of sampling steps is extremely limited (e.g., \(N=6\)), both methods may produce wayward trajectories that deviate substantially from the reference path, indicating that the correction term does not fully prevent geometric mis-tracking caused by aggressive multistep extrapolation. These observations suggest that the proposed corrections improve robustness but do not completely eliminate failure modes associated with rapidly varying dynamics, large curvature, or stiffness-like effects. Developing more reliable adaptive corrections for such challenging low-step regimes remains an important direction for future work.

\end{document}